\documentclass{article} 
\usepackage{iclr2025_conference,times}

\usepackage{booktabs}
\usepackage{amsmath}
\usepackage{makecell}
\usepackage{hyperref}
\usepackage{url}
\usepackage{graphicx}
\usepackage{multirow}
\usepackage{booktabs}
\usepackage{textcomp}
\usepackage{algorithm}
\usepackage{algpseudocode}
\usepackage{tikz}
\usepackage{wrapfig}
\newcommand{\hut}[1]{\textcolor{black}{#1}}
\newcommand{\zjn}[1]{\textcolor{black}{#1}} 
\newcommand{\yr}[1]{\textcolor{black}{#1}}
\newcommand{\reb}[1]{\textcolor{black}{#1}}

\definecolor{green_code}{RGB}{55, 126, 34}

\def \pzo {\phantom{0}} 

\begin{wrapfigure}{l}{0.15\textwidth}
    \vspace{-20pt}
    \includegraphics[width=\linewidth]{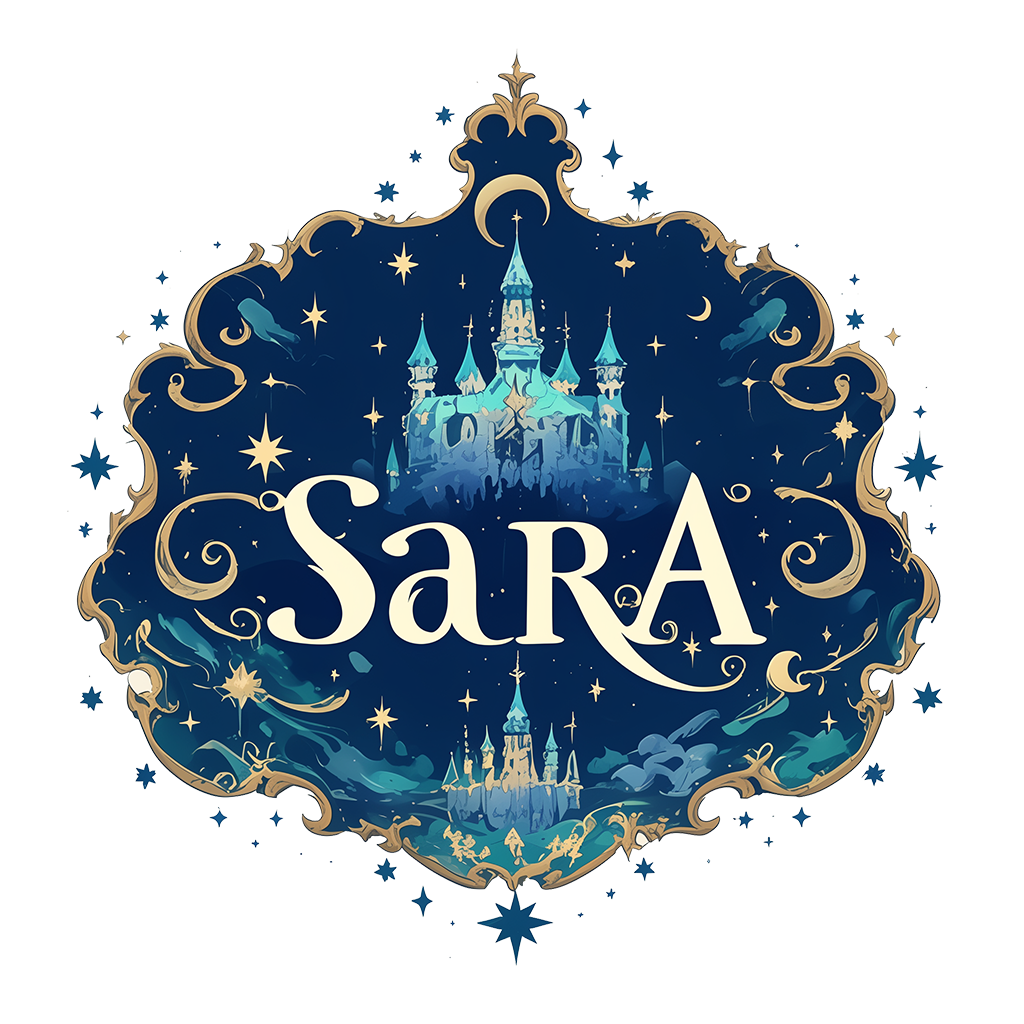} 
    \vspace{-20pt}
\end{wrapfigure}

\title{High-Efficient Diffusion Model \\Fine-tuning with Progressive \\ \underline{S}p\underline{a}rse Low-\underline{R}ank \underline{A}daptation}

\author{Teng Hu$^{1\ast}$, Jiangning Zhang$^{2}$\thanks{Equal Contribution} , Ran Yi$^{1}$\thanks{Corresponding Author} , Hongrui Huang$^{1}$, Yabiao Wang$^2$, Lizhuang Ma$^1$ \\
$^1$ Shanghai Jiao Tong University ~~~~~~
$^2$ Youtu Lab, Tencent\\
\texttt{\{hu-teng,ranyi,lzma\}@sjtu.edu.cn} ~~~
\texttt{2022212016@stu.hit.edu.cn}\\
\texttt{\{vtzhang,caseywang\}@tencent.com}\\
}

\iclrfinalcopy 
\begin{document}

\maketitle

\begin{abstract}
The development of diffusion models has led to significant progress in image and video generation tasks, with pre-trained models like the Stable Diffusion series playing a crucial role.
However, a key challenge remains in downstream task applications: how to effectively and efficiently adapt pre-trained diffusion models to new tasks.
Inspired by model pruning which lightens large pre-trained models by removing unimportant parameters, we propose \textbf{SaRA}, a novel model fine-tuning method with progressive \textbf{\underline{S}}p\textbf{\underline{a}}rse low-\textbf{\underline{R}}ank \textbf{\underline{A}}daptation to 
make full use of these ineffective parameters and enable the pre-trained model with new task-specified capabilities.
In this work, we first investigate the importance of parameters in pre-trained diffusion models and discover that parameters with the smallest absolute values do not contribute to the generation process due to training instabilities.
Based on this observation, we propose a fine-tuning method termed SaRA that re-utilizes these temporarily ineffective parameters, equating to optimizing a sparse weight matrix to learn the task-specific knowledge.
To mitigate potential overfitting, we propose a nuclear-norm-based low-rank sparse training scheme for efficient fine-tuning.
Furthermore, we design a new progressive parameter adjustment strategy to make full use of the finetuned parameters.
Finally, we propose a novel unstructural backpropagation strategy, which significantly reduces memory costs during fine-tuning.
Our method enhances the generative capabilities of pre-trained models in downstream applications and outperforms existing fine-tuning methods in maintaining model's generalization ability. Source code is available at \url{https://sjtuplayer.github.io/projects/SaRA}.

\end{abstract}

\section{Introduction}

In recent years, with the development of diffusion models~\citep{ddpm,stablediffusion}, tasks such as image generation~\citep{dreambooth,zhang2023controlnet}, video generation~\citep{animatediff,blattmann2023alignyoulatents}, and 3D generation~\citep{dreamfusion,sun2023dreamcraft3d} have made significant advancements. Pre-trained diffusion models, particularly the Stable Diffusion series~\citep{stablediffusion}, have played a crucial role in these developments
, including image customization~\citep{dreambooth}, image editing~\citep{kawar2023imagic}, and controllable generation~\citep{zhang2023controlnet,t2i-adapter}. Additionally, by leveraging prior information from the image domain, diffusion models have been extended to tasks such as video~\citep{animatediff,blattmann2023alignyoulatents} and 3D generation~\citep{dreamfusion,sun2023dreamcraft3d}. 
\hut{As these applications continue to evolve, a core issue emerges: how to effectively} \zjn{and efficiently fine-tune} the foundational pre-trained diffusion models and apply them to new \yr{tasks}.

\begin{figure}[t]
\centering
\includegraphics[width=0.73\textwidth]{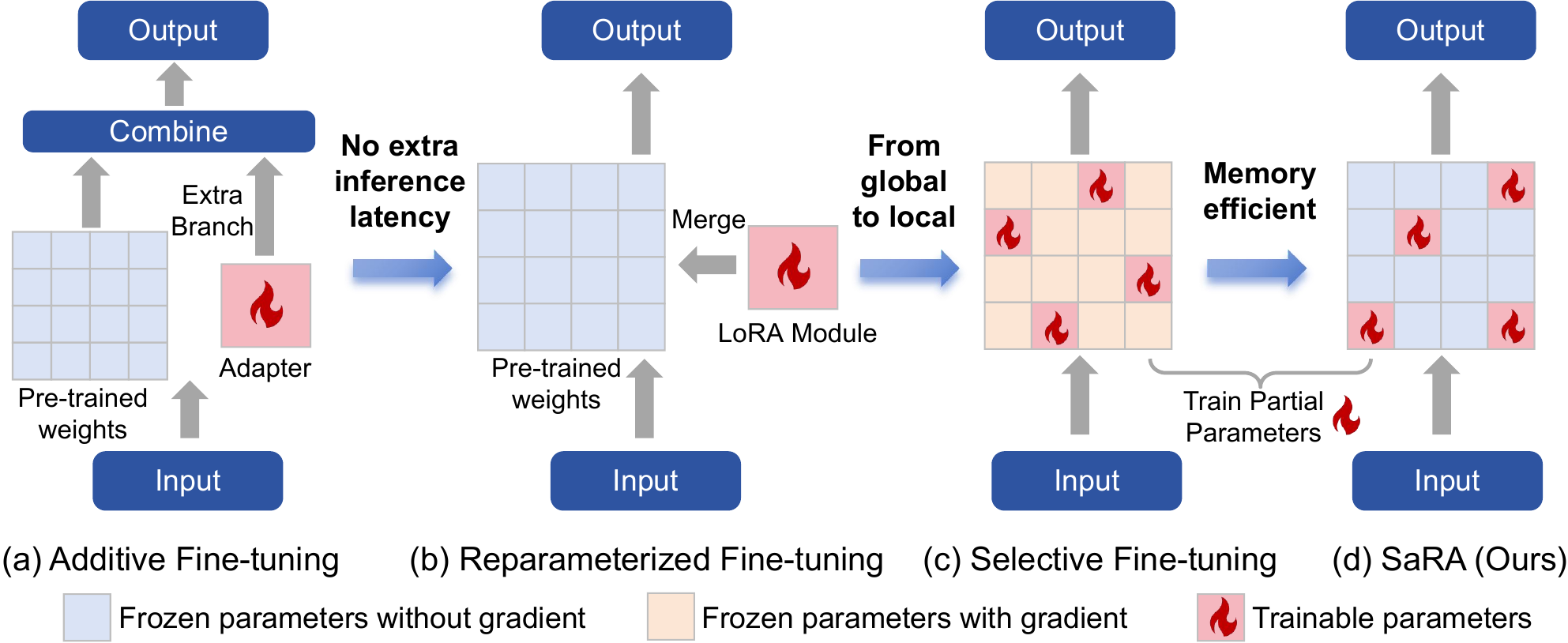}
\vspace{-0.15in}
\caption{
The reparameterized fine-tuning methods (b) address the additional inference latency introduced by additive fine-tuning methods (a) through reparameterizing the pre-trained weights from a global view. Selective fine-tuning methods (c) improve upon global parameter updates by employing sparse updates, which better preserve the model prior by freezing most of the pre-trained parameters. Our SaRA (d) further enhances (c) by significantly reducing memory costs and achieving superior performance in both adaptation capability and prior preservation. }
\label{fig:motivation}
\vspace{-0.25in}
\end{figure}

\reb{Existing fine-tuning methods~\citep{han2024PEFTsurvey,pan2024lisa,SpIEL,sung2021fishmask,fang2024dropout}} can be categorized into three \yr{categories} (Fig.~\ref{fig:motivation}): 
\textbf{\textit{1)}} \textbf{Additive fine-tuning (AFT) \yr{methods}}~\citep{chen2022adaptformer}, which introduce additional modules to fine-tune the model, such as adapter-based tuning~\citep{ipadapter,t2i-adapter}. \yr{However, these methods require additional} modules \yr{and parameters}, \yr{which has changed} the source model, \hut{\yr{and} also \yr{introduced} additional latency during the inference stage}. 
\textbf{\textit{2)}} \textbf{Reparameterized fine-tuning (RFT) \yr{methods}}~\citep{hu2021lora,zhang2023adalora}, which primarily utilize low-rank matrices to learn new information \hut{and merge the learned parameters with the pre-trained one}, but it still suffers \yr{from} the risk of overfitting\yr{,} since all parameters are adjusted by the low-rank matrices \hut{globally}. 
Moreover, the choice of rank and the specific layers to which LoRA is applied \yr{requires} a tailored design for each model. 
\textbf{\textit{3)}} \textbf{Selective-based fine-tuning (SFT) \yr{methods}}~\citep{guo2020diffpruninig,ansell2021Lt-sft}, which \yr{select} a subset of the model's existing parameters for fine-tuning. However, the complex parameter selection process and \hut{high memory cost} restrict their application in diffusion models. 
\hut{Overall, both AFT and RFT methods \yr{require} model-specific designs, \yr{\textit{e.g.,}} exploration of which layers to \yr{apply Adapters or LoRAs} within the model, and the hidden dimension or rank needs to be adjusted according to the specific task\yr{s}. 
The SFT method introduces considerable \yr{latency,} suffers from hyperparameter sensitivity in parameter selection, \yr{and} also \yr{performs poorly} in terms of effectiveness and training efficiency.}
Therefore, a pressing question arises: Can we \yr{design} a universal method that is \textbf{\textit{model-agnostic, does not require \hut{hyperparameter searching}, inherently avoids overfitting, and simultaneously achieves high-efficiency plug-and-play model fine-tuning}}?



\hut{\yr{Inspired by a theory in model p}runing, 
\yr{which} posits that within a trained model, there exist parameters with relatively small \yr{absolute values} that have negligible impact on the model's output, 
an intuitive idea is: whether we can \yr{find a way to} leverage these ineffective parameters to \yr{make them effective again, and} enhance the model's generative capabilities. 
To achieve this \yr{goal}, the target \yr{``}ineffective\yr{"} parameters we seek must possess two properties: 
1) \textbf{temporary ineffectiveness}: the parameters themselves have minimal impact on the current model's output; 
2) \textbf{potential effectiveness}: the parameters \yr{are} not redundant due to the model structure, but have \yr{a certain ability} to learn new knowledge 
\yr{(if handled properly, they can be effective again)}. 
We first 
\yr{conducted an analysis on} the \yr{influence} of small parameters in pre-trained diffusion models \yr{on the model outputs}, and found that the smallest $10\%$ (even $20\%$) of parameters by absolute \yr{values} did not contribute \yr{much} to the generative process (Fig.~\ref{fig:SD-Threshold}\yr{, see Sec.~\ref{sec: useless parameters}}). 
Further\yr{more}, we examine\yr{d} the potential effectiveness of these parameters and discover\yr{ed} that their ineffectiveness is not inherent \zjn{(extrinsic)} to the model's nature, but rather due to the instability of the training process \zjn{(see Sec.~\ref{sec:unstable training})}. Specifically, the randomness in the training process causes \yr{some} parameters to \yr{approach} zero by the end of training. 
This \yr{observation} inspired us to rationally utilize these \yr{temporally ineffective} 
parameters to \yr{make them effective again and} fine-tune pre-trained generative models.}

Therefore, we propose \textbf{SaRA}, a novel fine-tuning method for pre-trained diffusion models \yr{that} train\yr{s} \textbf{the parameters with \yr{relatively small} absolute values}.
\yr{We first identify the ``temporally ineffective, potentially effective'' parameters as paramters} smaller than a threshold \yr{in the pre-trained weights}. 
\yr{We then} efficiently fine-tune \yr{these parameters in the} pre-trained weights by sparse matrices while preserving prior knowledge. 
To mitigate the risk of overfitting due to the 
\yr{potential high rank}
of sparse matrices, we \yr{propose} a \textbf{low-rank sparse training} scheme, which employs a nuclear norm-based low-rank loss to constrain the rank of the learned sparse matrices, achieving efficient fine-tuning of diffusion models. 
\yr{In addition}, recognizing that some parameters may not be fully utilized during the fine-tuning process, we propose a \textbf{progressive parameter adjustment strategy}, \yr{which} introduces a second stage to reselect parameters below the pre-defined threshold and train them, ensuring that almost all parameters contribute effectively.
Finally, 
\yr{different from the typical} selective PEFT method\yr{s that retain the gradient of the entire parameter matrices and} require \yr{high} memory cost (the same as full-parameter fine-tuning), we propose an \textbf{unstructural backpropagation strategy} \yr{with smaller memory cost}. 
In this strategy, we \yr{only} retain the gradients for the parameters to be updated, and automatically discard the gradients for other parameters during the backpropagation process. This results in a memory-efficient selective PEFT method, which also advances the development of future selective PEFT techniques. 
\hut{Compared to previous fine-tuning methods~\citep{hu2021lora,valipour2022dylora,hayou2024lora+}, our SaRA is capable of effectively enhancing the generative capabilities of the pre-trained model itself, and it also demonstrates the best ability for model adaptation and prior preservation in different downstream tasks.}

Contributions \yr{of this paper} can be summarized in the following \yr{four aspects}:
    \textbf{\textit{1)}}\hut{We investigate the importance of the parameters in pre-trained diffusion models, revealing the temporal ineffectiveness and potential effectiveness of the parameters with the smallest absolute weight, which motivates us to make full use of these parameters.}
    \textbf{\textit{2)}} \hut{We propose \zjn{SaRA}, a novel efficient fine-tuning method based on progressive sparse low-rank adaptation, enabling the model to learn new knowledge without influencing the original generalization ability. 
    }
    \textbf{\textit{3)}} \hut{We propose unstructural backpropagation, which resolves the high memory consumption problem of selective PEFT methods and surpasses LoRA in \yr{memory} efficiency (save more than 40\% GPU memory than LoRA and selective PEFT methods)}.
    \textbf{\textit{3)}} \hut{We efficiently encapsulated \yr{and implemented} our method 
    in a single line of code modification, which significantly reduces the coding overhead associated with fine-tuning pre-trained models. 
    }
\section{Related Works}

\subsection{Diffusion Models}

Diffusion models~\citep{ddpm,stablediffusion} have demonstrated significant advantages in image generative tasks. Text-to-image models, represented by Stable Diffusion~\citep{stablediffusion}, have diversified into various applications. \hut{However, their large parameter sizes somewhat limit the feasibility of full fine-tuning to adapt to specific new tasks.}
Methods such as ControlNet~\citep{zhang2023controlnet}, T2I-Adapter~\citep{t2i-adapter}, and IP-Adapter~\citep{ipadapter} achieve controlled generation under different conditions by adding external networks to diffusion models. Additionally, models like LoRA~\citep{hu2021lora} and DreamBooth~\citep{ruiz2023dreambooth} enhance the original diffusion models through fine-tuning, enabling them to generate content in new domains and concepts. Furthermore, some video generation models~\citep{animatediff,blattmann2023alignyoulatents} are built on diffusion models to achieve video generations and employ Lora and adapters to accomplish controllable video generations.

\subsection{Parameter-efficient model fine-tuning}


\hut{\textbf{Addictive Parameter Fine-tuning \yr{(AFT)}.} AFT introduces additional modules to the model while keeping the pre-trained backbone fixed. Serial Adapter~\citep{houlsby2019parameter} enhances the Transformer block by adding new modules after the self-attention layer and FFN layer. AdapterFusion~\citep{pfeiffer2020adapterfusion} streamlines this by inserting adapter layers only after the FFN layers to boost computational efficiency. Parallel adapters, including Adaptformer~\citep{chen2022adaptformer}, CoDA~\citep{lei2023conditional}, and KronA~\citep{edalati2022krona}, reorganize the traditionally sequential adapter layers into a parallel side-network, optimizing both performance and efficiency. To further enhance adapter performance and generalization, multi-task learning strategies like AdaMix~\citep{wang2022adamix}, and Hyperformer~\citep{mahabadi2021parameter} have also been developed. }

\textbf{Reparameterized Parameter Fine-tuning \yr{(RFT)}.} 
\yr{An e}arly work~\citep{aghajanyan2020intrinsic} has verified the presence of low intrinsic dimensionality in pre-trained models. LoRA~\citep{hu2021lora} proposes \yr{to use} a low-rank matrix to learn new feature representations. 
To address the issue of selecting the appropriate rank, DyLoRA~\citep{valipour2022dylora} employs a dynamic and search-free approach to obtain the optimal rank. AdaLoRA~\citep{zhang2023adalora} decomposes the trainable low-rank matrix using singular value decomposition (SVD) and implements dynamic rank adjustment by pruning singular values. Furthermore, numerous subsequent methods~\citep{yang2023bayesian,ding2023sparse,hayou2024lora+} have aimed to enhance the performance of LoRA. 

\textbf{Selective Parameter Fine-tuning \yr{(SFT)}.} Selective parameter finetuning~\citep{han2024PEFTsurvey} \yr{methods} finetune \yr{a selected subset} of the parameters in the pre-trained model. 
Diffpruning~\citep{guo2020diffpruninig} fine-tunes specific parameters by learning a mask matrix, constraining its size through a differentiable L0 norm. PaFi~\citep{liao2023pafi} selects the parameters with the smallest absolute values for learning, 
while LTSFT~\citep{ansell2021Lt-sft}, grounded in the Lottery Ticket Hypothesis~\citep{frankle2018lottery}, selects the parameters that change the most during fine-tuning. \reb{SHiRA~\citep{bhardwaj2024shira} proposes a sparse high-rank adaption method to improve the adaptation ability.}
Essentially, these methods all learn a sparse mask matrix to fine-tune the pre-trained models.  


Overall, both AFT and RFT methods require model-specific designs, e.g., determining which layers to apply Adapter or LoRA, and adjusting the hidden dimension or rank according to specific tasks. Additionally, the SFT method introduces significant latency, exhibits hyperparameter sensitivity in parameter selection, and underperforms in terms of effectiveness and training efficiency. In contrast, our SaRA is model-agnostic, which eliminates the need for layer selection, and can effectively fine-tune the pre-trained model while reducing training costs in both time and memory.

 

 \section{The potential effectiveness of the ineffective parameters  }
\label{sec:potential effectiveness}

\begin{figure}[t]
\centering
\includegraphics[width=0.9\textwidth]{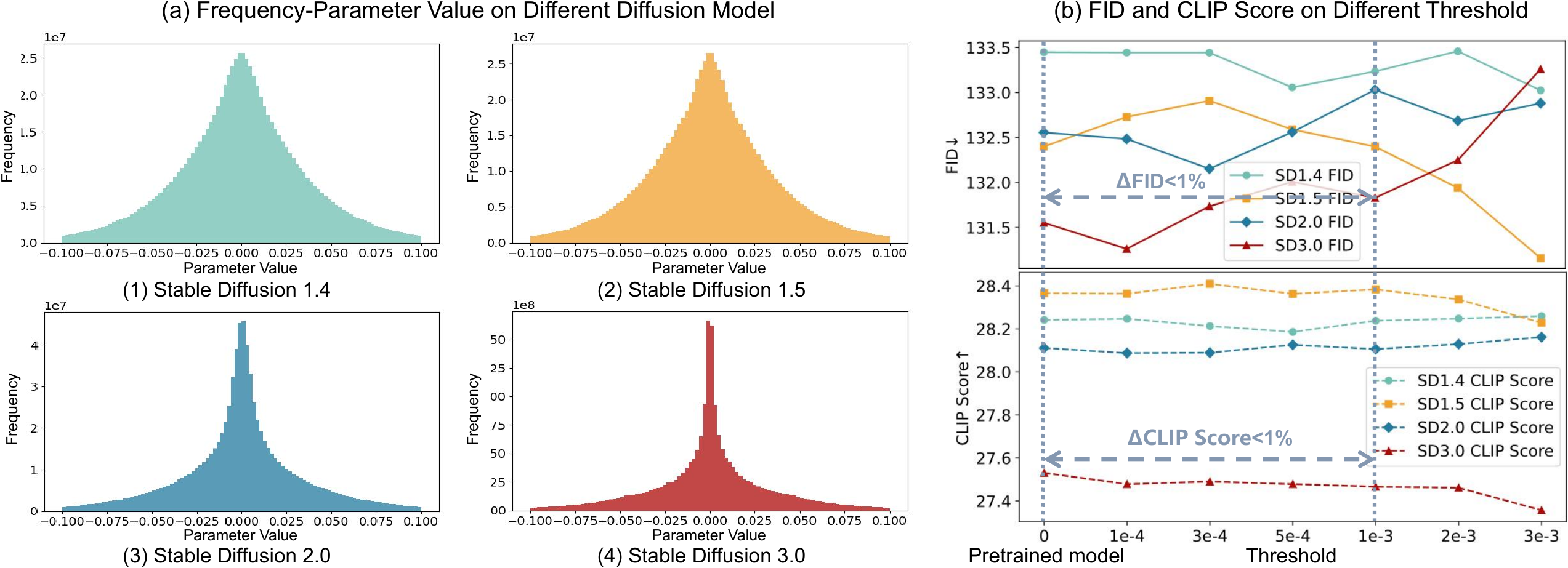}
\vspace{-0.15in}
\caption{\hut{(a) Weight distributions of the pre-trained parameters in Stable Diffusion \yr{(SD) 1.4}, 1.5, 2.0, and 3.0, which \reb{are all similar} to a Gaussian distribution, therefore a large number of parameters \yr{are} around $0$. 
\reb{(b) The performance (FID and CLIP Score ) of SD Models when the parameters in the pre-trained models with absolute values smaller than a certain threshold are set to $0$.}}
}
\label{fig:SD-Threshold}
\vspace{-0.2in}
\end{figure}

\subsection{Ineffective parameters in Stable Diffusion models}
\label{sec: useless parameters}
Based on the theorem of model pruning \yr{proposed in}~\citep{liang2021pruningsurvey}\yr{,} which regards the parameters with the smallest absolute \yr{values} as \yr{``}ineffective\yr{"} parameters, we investigate\yr{d} the effectiveness of these parameters in pre-trained stable diffusion models (version 1.4, 1.5, 2.0, and 3.0). 
\yr{W}e set parameters with absolute values below a certain threshold $\theta_t$ (from $10^{-3}$ to $10^{-5}$) to zero, and evaluated the performance of the regularized models on generative tasks with \yr{CLIP} Score~\citep{radford2021clip} and Fréchet Inception Distance (FID)~\citep{heusel2017fid}.

The results are shown in Fig.~\ref{fig:SD-Threshold}\yr{(b)}. We observe\yr{d} that within a certain threshold range $\theta_t\in(0,10^{-3}]$, \yr{with the small parameters set to $0$,} the generative \yr{ability} of the SD models is minimally affected. 
\yr{A}nd in some cases, the regularized model \yr{with ``ineffective" parameters set to $0$} even outperforms the original model (\yr{\textit{i.e.,} no parameters are set to $0$, with} $\theta_t=0$). 
Specifically, SD1.4 and SD1.5 show better FID scores than the original model when thresholds are in the range of $\theta_t\in[5\times 10^{-4},10^{-3}]$, and SD2.0 and SD3.0 exhibit superior FID scores at a threshold of $\theta_t=10^{-4}$. 
\yr{These results show} that parameters with the smallest absolute values have a limited impact on the generative process, and in some cases, they may even slightly impair the model's generative ability. 

\subsection{Unstable Training Process Contributes to Useless Parameters}
\label{sec:unstable training}

Sec.~\ref{sec: useless parameters} demonstrated that parameters with smaller absolute values have minimal impact on the generative capability of diffusion models. 
\yr{A natural} question arises: \yr{\textbf{are these currently ineffective parameters caused by the model structure and inherently redundant, or are they caused by the training process and can become effective again?}
If it is the former case, }
the structural design of the model prevents these parameters from learning effective information, 
then these parameters are \yr{redundant and} unlikely to be useful in subsequent training processes. 
\yr{While if it is the latter case, these parameters are potentially effective when leveraged rationally in the subsequent training.}
Therefore, we further investigate\yr{d} the reasons behind the ineffectiveness of these parameters, and \yr{found} that the ineffectiveness is due to the \yr{randomness} of the optimization process, rather than an inherent inability \yr{caused by model structure}.

\begin{wrapfigure}{r}{0.45\textwidth}
\centering
\vspace{-0.2in}
\includegraphics[width=0.45\textwidth]{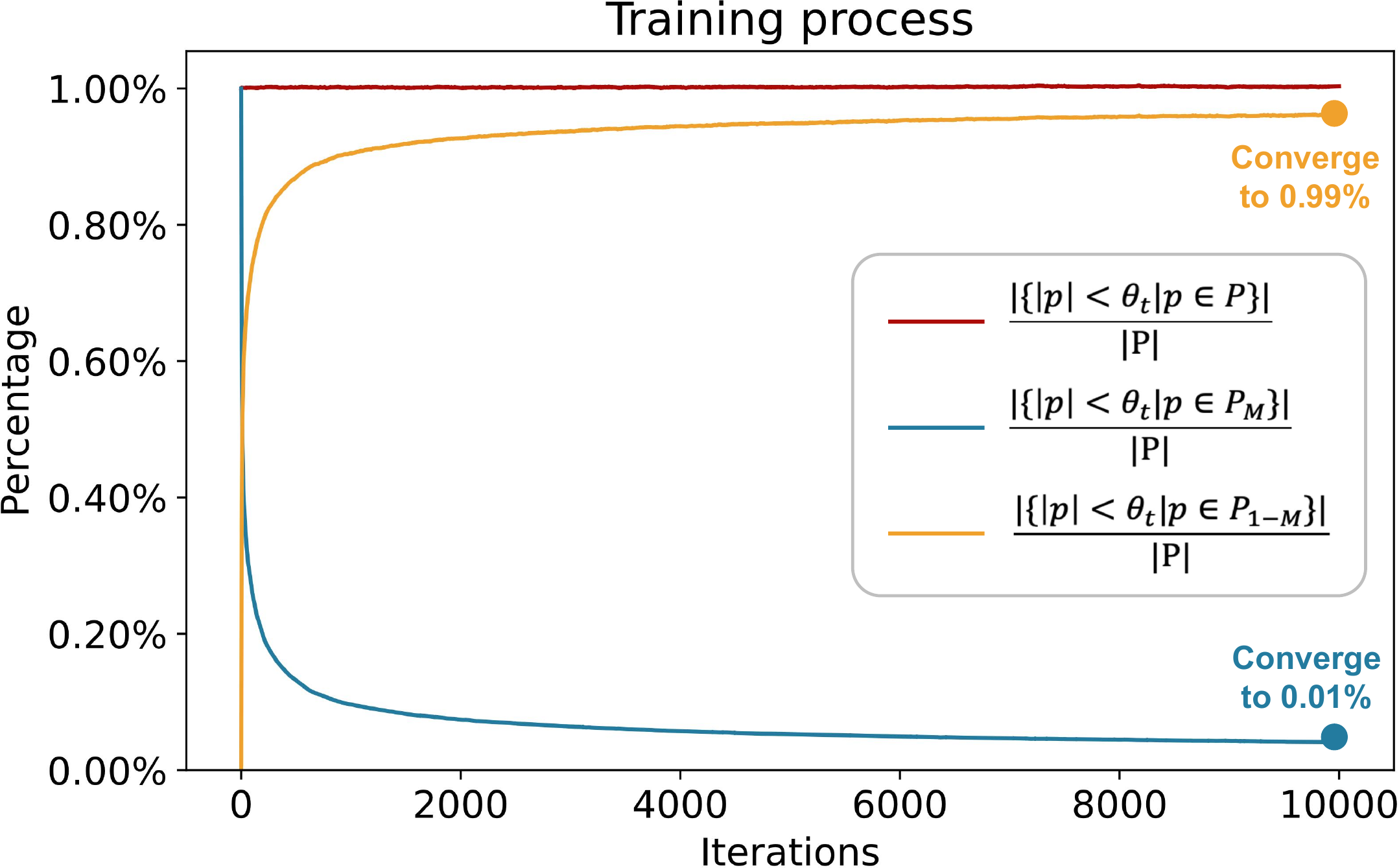}
\vspace{-0.3in}
\caption{
\yr{The changes of parameters whose absolute values are bewlow the $1\%$ theshold $\theta_t$ during full-parameter fine-tuning. The blue and yellow curves show the proportions of parameters originated from both the initially below-threshold $P_{M}$ and the initially above-threshold $P_{1-M}$.}
}
\label{fig:training process}
\vspace{-0.25in}
\end{wrapfigure}

Specifically, we employ\yr{ed} a Stable Diffusion model pre-trained on the FFHQ dataset~\citep{karras2019ffhq}, \yr{whose} parameter matrices \yr{are denoted} as $P_0$. 
We record\yr{ed the} parameters \yr{in the pre-trained model} with absolute values below the $1\%$ threshold $\theta_t$ by \yr{a} parameter mask $M$, 
where 
\yr{$P_{M} = P_0\odot M$ denotes the initially below-threshold parameters ($1\%$ of all parameters)}, 
and 
\yr{$P_{1-M} = P_0\odot (1-M)$ denotes the initially above-threshold parameters ($99\%$ of all parameters), \hut{satisfying}:}
\yr{
\begin{equation}
\begin{aligned}
\left| p \right|<\theta_t, &\forall p\in P_M, \\
\left| p \right|\ge \theta_t, &\forall p\in P_{1-M}.
\end{aligned}
\end{equation}
}
\noindent Then, we continue training this pre-trained model on the FFHQ dataset\yr{,} and observe the changes \yr{of} its parameters $P$ during the training process. 
\yr{During this fine-tuning stage}, we recorded the 
\yr{``source''} of parameters \yr{whose absolute values are} below the threshold $\theta_t$\yr{ ($\{ \left| p \right|<\theta_t \}$), \textit{i.e.,} whether they are initially below-threshold or initially above-threshold.}
\yr{And we found that t}hese parameters originated from both the initially below-threshold $P_M$ and the initially above-threshold $P_{1-M}$. 

The proportions of these two groups \yr{and how they change during the finetuning} are shown in Fig.~\ref{fig:training process}. 
As \yr{the} training progressed, the proportion of $P_M$ remaining below $\theta_t$ gradually \yr{decreased from $100\%$} to 
\yr{$1\%$ (blue curve decreases from $1.00\%$ to $0.01\%$);}
while 
\yr{$1\%$} of the initially above-threshold $P_{1-M}$ eventually fell below $\theta_t$ \yr{(yellow curve raises from $0.00\%$ to $0.99\%$)}. 
\yr{The results} indicate that initially ineffective parameters $P_M$ \yr{caused by} the randomness of the training process, mostly become effective over time \yr{(only $1\%$ remaining below threshold)}. 
Conversely, some initially effective parameters become ineffective as training continues. 
This \yr{pheonomenon} demonstrates that the ineffectiveness of parameters is not inherent \yr{to model structure,} but rather a result of the stochastic nature of the training process, which causes some parameters to fall below the threshold $\theta_t$ at the last training step coincidentally, \yr{making} them \textbf{temporarily ineffective}. 
As \yr{the} training continues, most of these parameters regain effectiveness, proving their \textbf{potential\yr{ly} effectiveness,} 
\yr{which motivates us to leverage these temporarily ineffective parameters}
to fine-tune the pre-trained model.

\section{Progressive sparse low-rank model adaptation}

\hut{Inspired by the potential effectiveness of parameters with the smallest absolute values, as discussed in Sec.~\ref{sec:potential effectiveness}, we propose SaRA, a novel parameter-efficient fine-tuning method designed to fully utilize these temporarily ineffective parameters. Specifically, we first identify the ineffective parameters in the pre-trained parameters $P_0$ by computing \yr{a} sparse mask $M=P_0<\theta_t$\yr{, where $\theta_t$ is a threshold and the sparse mask only selects a small portion from all parameters}. 
We then use this sparse mask to update the 
\yr{initially ineffective} parameters $P\odot M$, while keeping the initially effective parameters $P\odot(1-M)$ \yr{frozen}. 
This approach enables the pre-trained model to acquire new capabilities for downstream tasks (through the \yr{learnable} $P\odot M$) while preserving prior information (through the fixed $P\odot(1-M)$).
To \yr{avoid the problem of overfitting caused by strong representation ability due to the potential high rank of} the \yr{learnable} sparse matrix $P\odot M$, 
we \yr{propose} a nuclear \yr{norm}-based low-rank loss to mitigate overfitting \yr{(Sec.~\ref{ssec:low-rank})}. 
\yr{In addition}, we propose a progressive parameter adjustment strategy to further \yr{make full} use of \yr{the} ineffective parameters by progressively reselecting them \yr{(Sec.~\ref{ssec:progressive})}. 
Finally, we \yr{propose} an unstructured backpropagation strategy, which significantly reduces memory costs and can be applied to enhance all selective PEFT methods. }

\subsection{Fine-tuning on the potential effective parameters} 
\yr{In} Sec.~\ref{sec:potential effectiveness}, we have demonstrated that parameters with small absolute values are ineffective in the generative process of diffusion models, and this ineffectiveness is not due to the model's architecture but rather the stochastic nature of the optimization process.
\hut{Therefore, we propose SaRA, which fine-tunes these \yr{temporarily ineffective} parameters to adapt the pre-trained diffusion model to downstream tasks}, enabling it to learn new knowledge while preserving its original generative capability. 
Specifically, we first obtain a mask $M$
for the initial parameter set $P_0$, \yr{which satisfies:} 
\yr{
\begin{equation}
\begin{aligned}
\left| p \right|<\theta_t, \forall p\in P_0\odot M,
\end{aligned}
\end{equation}
}
where $M$ is a \textbf{sparse matrix}, since \yr{the theshold $\theta_t$ is set low and} 
only selects a small portion from \yr{all} the parameters.
\yr{We then use this sparse mask to update the initially ineffective parameters $P_M=P\odot M$, while keeping the initially effective parameters $P\odot(1-M)$ frozen.}
During training, for the gradient $\nabla P$ of the parameters, we use the pre-defined \yr{sparse} mask $M$ to retain the gradients we need and update the corresponding parameters $P_M\yr{=P\odot M}$ by:
\begin{equation}
\begin{aligned}
\nabla P_M&=\nabla P\odot M+\mathbf{0}\odot (1-M),\quad P_{new}&=P-\lambda \cdot \nabla P_M.
\end{aligned}
\end{equation}
In this way, we can focus on training the ineffective parameters while keeping the other parameters unchanged, ensuring the original generation ability of the pre-trained model \yr{is preserved,} while learning new knowledge by the parameters $P_M$.





\subsection{Nuclear \yr{Norm}-based Low-Rank Constraint} 
\label{ssec:low-rank}
The sparse parameter matrices $P_M$ can sometimes have a high rank, resulting in strong representational capabilities that may lead to overfitting during \yr{the training process of} downstream tasks. 
To mitigate this issue, we introduce a \yr{nuclear norm-based low-}rank constraint on the sparse matrix to prevent the rank from becoming excessively high during the training process.

\yr{A direct way to apply low-rank constraint is to minimize} 
the rank of the sparse parameter matrix $Rank(P)$ as a constraint. 
However, directly minimizing the rank function is computationally intractable due to its non-convex nature.
Therefore, we use \textbf{nuclear norm} to estimate its rank:
\begin{equation}
    \begin{aligned}
        \|P_M\|_* = \sum_{i} \sigma_i(P_M),\,\text{where $\sigma_i$ are the singular values of $P_M$. }
    \end{aligned}
\end{equation}

To compute the nuclear norm $\|P_M\|_*$,  we employ the singular value decomposition (SVD) of the matrix $P_M = U \Sigma V^T$, where \(U\) and \(V\) are orthogonal matrices, and \(\Sigma\) is a diagonal matrix containing the singular values \(\sigma_i(P_M)\). 
The subgradient of the nuclear norm at \(P_M\) has been \yr{derived} by~\citep{watson1992subgradient}.
Based on this derivation of the nuclear norm gradient, we can ensure that gradient descent methods can be employed to incorporate nuclear norm-based low-rank constraints into the training process, thereby achieving \yr{our} \textbf{\yr{nuclear norm-based} low-rank constrained loss}:
\begin{equation}
    \begin{aligned}
        L_{rank}=\|P_M\|_* = \sum_{i} \sigma_i(P_M).
    \end{aligned}
\end{equation}

\subsection{Progressive parameter adjustment}  
\label{ssec:progressive}

\yr{As discussed in Sec.~\ref{sec:unstable training} and Fig.~\ref{fig:training process}, when continuing training the pre-trained model, the initially ineffective parameters gradually become above threshold and effective, with only $1\%$ of initially ineffective parameters remaining below threshold eventually. 
However, the speed at which ineffective parameters become effective (the slope of blue curve in Fig.~\ref{fig:training process}) varies during the finetuning process. 
In the early stage of the finetuning process (\textit{e.g.,} the first $2.5$k iterations), a large portion (over $80\%$) of initially ineffective parameters quickly become effective, with a small part (less than $20\%$) remaining below threshold.
However, the speed slows down in the later stage of finetuning: from $2.5$k to $8$k iterations, the small portion of remaining below-threshold parameters jumps out of the theshold very slowly.
However, the finetuning iterations are typically limited (\textit{e.g.,} a few thousands), in which case the slow speed in the later finetuning stage can cause problems: the remaining below-threshold ineffective parameters may not be trained to be effective and fully utilized.
}

\hut{To address this issue, we propose a \textbf{progressive parameter adjustment} strategy. 
\yr{To alleviate the slow speed of ineffective parameters becoming effective in the later stage, we reselect the ineffective parameters that remain below threshold (about $15\%$-$20\%$ of initial ineffective parameters) after the early finetuning stage and focus on optimizing these remaining below-threshold parameters in the subsequent finetuning stage. 
Compared to finetuning without this reselecting operation, this strategy can quickly make remaining ineffective parameters effective again in the later finetuning stage.}}

\hut{Specifically, we introduce a parameter readjustment phase. After \yr{the early finetuning stage} 
(we \yr{set the first} half of the total iterations \yr{as the early finetuning stage}, \textit{e.g.,} 2,500 iterations when there are 5,000 \yr{finetuning} iterations) on the initially selected 
\yr{below-threshold} parameters $P^0_{learn}$, we reselect parameters from $P^0_{learn}$ that remain below the predefined threshold as new trainable parameters $P_{learn}$ \yr{(which is a subset of $P^0_{learn}$ and typically has $15\%$-$20\%$ of $P^0_{learn}$'s parameters)}.
\yr{Then in the subsequent finetuning stage, we only optimize this subset of initial ineffective parameters, and keep other parameters of $P^0_{learn}$ frozen.}
\yr{By focusing on optimizing the small subset of remaining below-threshold parameters, this strategy greatly improves the speed of ineffective parameters jumping out of the threshold in the later finetuning stage,}
thereby enhancing the model's adaptation capability. 
In our experiments, we found that under the same number of \yr{finetuning} iterations, models without the progressive strategy had $15\%$ of $P^0_{learn}$ remain\yr{ed} ineffective \yr{after finetuning}, \yr{while} models with the \yr{progressive} strategy only \yr{had} $2\%$ of $P^0_{learn}$ \yr{that were still} ineffective. 
\yr{The results indicate this strategy} significantly improves the performance of our \yr{method} during the fine-tuning process.}

\begin{figure}[t]
\centering
\includegraphics[width=0.75\textwidth]{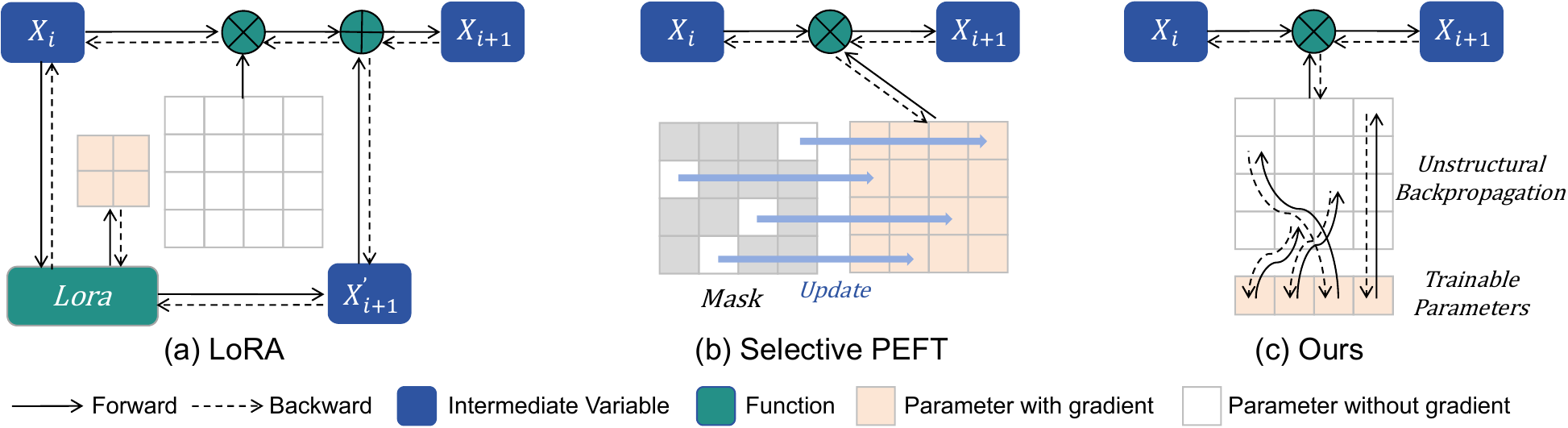}
\vspace{-0.15in}
\caption{Visualization of \yr{our unstructural backpropagation. }
\yr{a)} LoRA stores an additional intermediate variable $X'_{i+1}$ in each LoRA layer, and \yr{b)} selective PEFT methods store the gradients for the \yr{whole} parameters \yr{matrices}, causing a waste of memory and computation resources. 
\yr{c)} In contrast, our Unstructural Backpropagation method \yr{extracts the trainable parameters, sets them as independent leaf nodes, and only retains gradients for them, which largely reduces the memory cost. 
}}
\label{fig:memory improve}
\vspace{-0.25in}
\end{figure}


\begin{wrapfigure}{r}{0.42\textwidth}
\centering
\vspace{-0.15in}
\includegraphics[width=0.42\textwidth]{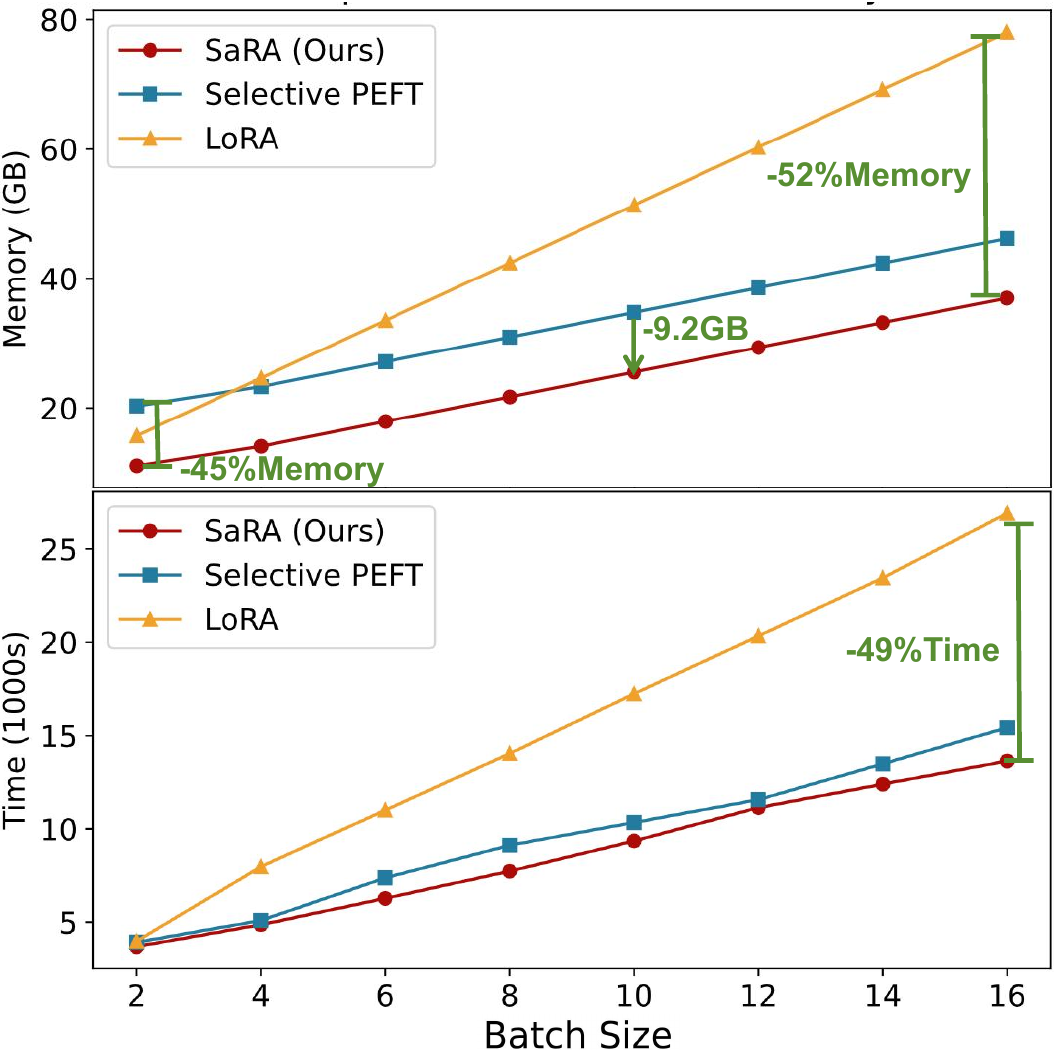}
\vspace{-0.25in}
\caption{\reb{Computation cost on memory and time of different PEFT methods.}
}
\label{fig:comparison on computation cost}
\vspace{-0.15in}
\end{wrapfigure}

\subsection{Unstructural Backpropagation}
\label{sec:Unstructural Backpropagation}
Currently, both the LoRA-based methods (the same for the adapter-based methods) and selective PEFT methods cause a heavy burden on the computation resources: 
\yr{1)} For the LoRA\yr{-based} methods, 
\yr{since the LoRA module is additional to the original model, there is no need to store the gradients of the model parameters,}
but they \yr{still require} additional memory cost\yr{s} to store the intermediate variables in the LoRA module, which is shown in Fig.~\ref{fig:memory improve} (a).
\yr{2)} And for \yr{the} selective PEFT methods, a persistent issue is that they require the same or even more computational resources (especially GPU memory) as full-parameter fine-tuning. 
\yr{Although they only finetune a subset of the model's parameters for fine-tuning, they retain the gradients of the entire parameter matrices $P$,}
because the mainstream deep learning libraries (such as PyTorch and TensorFlow) only support gradient backpropagation and updates for the entire parameter matrices. 
Consequently, previous selective PEFT methods had to perform gradient backpropagation on \yr{the entire} parameter matrices $P$, and then use pre-computed mask matrices $M$ to mask out the gradients of unnecessary parameters by $\nabla P_M=M\odot \nabla P$, and perform an overall parameter update by $P_{new}=P-\lambda \nabla P_M$ (visualized in Fig.~\ref{fig:memory improve} (b)). 
This approach necessitates storing the gradients of all model parameters and the additional mask matrices, leading to greater computational resource demands than full-parameter fine-tuning. This clearly contradicts the \yr{``}efficient" \yr{requirements} of PEFT and limits the practical applications of such methods.

To address this issue, we propose \textbf{Unstructural Backpropagation} (shown in Fig.~\ref{fig:memory improve} (c)), which supports efficient gradient backpropagation and updates for unstructured parameters. 
\yr{Different from previous selective PEFT methods that require retaining the gradient for the whole parameter matrices, our Unstructural Backpropagation only needs to retain gradients for the selected subset of below-threshold parameters $P_M$.}
Specifically, we \yr{first} store the mask matrices $M$ corresponding to each layer's parameters \yr{that} need to be trained\footnote{Since the \yr{mask} $M$ is of boolean type, it does not consume significant GPU memory.}. 
In the computational graph, we deviate from the traditional approach of setting model parameters as leaf nodes. 
Instead, we extract the trainable parameters $P_{learn}=P[M]\in \mathcal{R}^{\|M\|_0}$ and set them as independent leaf nodes, where $[\cdot]$ denotes element-wise indexing of the matrix. 
\yr{I.e., as shown in Fig.~\ref{fig:memory improve} (c), we extract the subset of trainable parameters and combine them into a separate parameter vector, and only retain gradients for this vector.}
\yr{Then, d}uring the forward pass, we define an Unstructural Mapping function $UM(\cdot)$ to update the model parameters $P$ by:
\begin{equation}
    \begin{aligned}
        P=UM(P,P_{learn},M), \yr{where}
\begin{cases} 
P[M]&=P_{learn},\\
P[1-M]&=P[1-M].
\end{cases}
    \end{aligned}
\end{equation}
\yr{And the updated model paramaters $P$ will} then participate in the training process. 
During backpropagation, we define the Unstructural Backpropagation  function $UB(\cdot)$ to propagate the gradients from the model parameters to the trainable parameters by:
\begin{equation}
    \begin{aligned}
        \nabla P_{learn}=UB(\nabla P, M)=\nabla P[M].
    \end{aligned}
\end{equation}
In this way, during backpropagation, the gradients on the model parameters $\nabla P$ \yr{will} be automatically cleared, since it is no longer a leaf node, and \yr{only} the gradients on the \yr{learnable} parameters $\nabla P_{learn}$ are stored, 
which significantly reduces the GPU memory during the training process. \reb{Notably, unstructual backpropagation is not limited to our method but can be employed in other SFT methods like LT-SFT~\cite{ansell2021Lt-sft}, which can advance the development of future SFT fields. }


\section{Experiments}



\begin{figure}[t]
\centering
\includegraphics[width=0.85\textwidth]{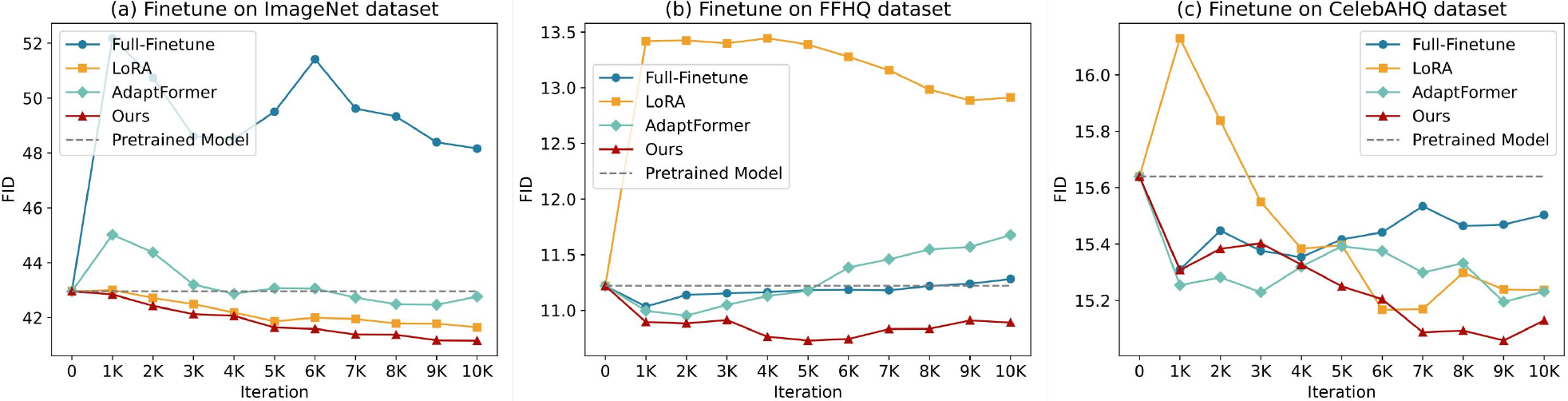}
\vspace{-0.15in}
\caption{
Quantitative comparison among different PEFT methods on Backbone Fine-tuning on ImageNet, FFHQ, and CelebA-HQ datasets. Our method achieves the best FID scores, indicating our method effectively improves the performance of the pre-trained models on the main task.
}
\label{fig:finetune-backbone}
\vspace{-0.25in}
\end{figure}

To validate the effectiveness of our method, we conduct experiments on various tasks, including backbone fine-tuning, downstream dataset fine-tuning, image customization, and controllable video generation (appendix). 
We compare our method with \yr{three} state-of-the-art parameter efficient fine-tunining methods: LoRA~\citep{hu2021lora}, Adaptformer~\citep{chen2022adaptformer}, and LT-SFT~\citep{ansell2021Lt-sft}\yr{;} along with the full-parameter fine-tuning method.
We evaluate the generation models by three metrics: 
\textit{\yr{1})} Fréchet Inception Distance (FID)~\citep{heusel2017fid}, \textit{2)} CLIP Score, and \textit{3)} Visual-Linguistic Harmony Index (VLHI), 
\reb{which balances FID and CLIP Score by:}
$$
       \reb{VLHI_i=\frac{\max(\{FID_i\}_{i=1}^{n})-FID_i}{\max(\{FID_i\}_{i=1}^{n})-\min(\{FID_i\}_{i=1}^{n})}+\frac{CLIP_i-\min(\{CLIP_i\}_{i=1}^{n})}{\max(\{CLIP_i\}_{i=1}^{n})-\min(\{CLIP_i\}_{i=1}^{n})}}
$$


\subsection{Backbone Fine-tuning}
\label{sec:improving backbone}
Different from the previous parameter-efficient fine-tuning methods \yr{that} mainly aim to fine-tune the pre-trained model to downstream tasks\yr{,}
our model enables the pre-trained model to make full use of the parameters. 
\yr{I}n other words, \yr{our finetuning method} can improve the performance of pre-trained models on the main task (the original task it is trained on), by \yr{optimizing the initially ineffective parameters to be effective and thus} increasing the number of effective parameters. 
Therefore, \yr{apart from} experimenting on downstream tasks like traditional PEFT methods, we \yr{first} apply our method to the main task of the pre-trained model, continuing to fine-tune the backbone on the original training dataset, \yr{in order} to explore whether our method can enhance the base model's performance. 
Specifically, we employ the pre-trained Stable Diffusion models on ImageNet~\citep{deng2009imagenet}, FFHQ~\citep{karras2019ffhq}, and CelebA-HQ~\citep{karras2017celebahq} datasets, and fine-tune them on \yr{these} pre-trained datasets for 10K iterations. 
We compare our method with full-parameter finetuning, LoRA, AdaptFormer, and LT-SFT by computing the FID metric between 5K generated data and 5K randomly sampled data from the source dataset. 
The results are shown in Fig.~\ref{fig:finetune-backbone}\yr{, which demonstrates that our method achieves the best FID scores, indicating our method effectively improves the performance of the pre-trained models on the main task}.

\begin{table}[t]
\renewcommand{\arraystretch}{1.3}
\resizebox{1.0\linewidth}{!}{
\begin{tabular}{c|c|c|ccc|ccc|ccc|ccc|ccc|ccc}
\toprule
\multirow{2}{*}{Backbone} & \multirow{2}{*}{Params} & \multirow{2}{*}{Model} & \multicolumn{3}{c}{BarbieCore} & \multicolumn{3}{c}{Cyberpunk} & \multicolumn{3}{c}{ElementFire} & \multicolumn{3}{c}{Expedition} & \multicolumn{3}{c}{Hornify} & \multicolumn{3}{c}{Mean}     \\ 
                          &                        &                         & FID $\downarrow$    & CLIP $\uparrow$   & VLHI $\uparrow$ & FID $\downarrow$ & CLIP $\uparrow$    & VLHI $\uparrow$  & FID $\downarrow$  & CLIP $\uparrow$     & VLHI $\uparrow$  & FID $\downarrow$  & CLIP $\uparrow$    & VLHI $\uparrow$  & FID $\downarrow$ & CLIP $\uparrow$   & VLHI $\uparrow$ & FID $\downarrow$   & CLIP $\uparrow$     & VLHI $\uparrow$ \\ \midrule
\multirow{13}{*}{SD 1.5}  & \multirow{4}{*}{50M}              & LoRA                                        & 161.88     & \textbf{29.93}      & \underline{1.34} & \textbf{117.49}    & \textbf{28.22}      & \textbf{1.85} & 181.66      & \textbf{27.47}      & 1.20 & \underline{136.31}     & \underline{27.39}      & 1.32 & 156.36  & \textbf{26.80}      &\textbf{1.28} & 150.74 & \textbf{27.96}      & \underline{1.45} \\
                          &                                   & Adaptformer                                 & 166.09     & \underline{29.00}      & 1.00 & 126.21    & 27.13      & 0.66 & \underline{151.27}      & 26.57      & \underline{1.29} & 138.01     & 26.41      & 0.63 & \underline{151.53}  & 26.20      & \underline{1.18} & \underline{146.62} & 27.06      & 1.18 \\
                          &                                   & LT-SFT                                      & \underline{157.80}     & 23.80      & 0.54 & 123.59    & 25.71     & 0.45 & 171.67      & 25.11      & 0.44 & 139.29     & \textbf{27.81}      & \underline{1.46} & 158.52  & \underline{26.35}      & 1.06 & 150.18 & 25.76      & 0.49\\
                          &                                   & \textbf{SaRA (Ours)}                                        & \textbf{148.54}     & 28.60      & \textbf{1.75} & \underline{121.67}    & \underline{27.30}      & \underline{1.15} & \textbf{132.67}      & \underline{26.77}      & \textbf{1.63} & \textbf{131.56}     & 27.34      & \textbf{1.48} & \textbf{140.36}  & 25.40      & 1.15 & \textbf{134.96} & \underline{27.08}      & \textbf{1.55} \\
                          \cline{2-21}
                          & \multirow{4}{*}{20M}              & LoRA                                        & 159.64     & \textbf{29.65}      & \underline{1.40} & \underline{117.21}    & \textbf{28.43}     & \textbf{1.95} & 174.79      & \textbf{27.61}      & \underline{1.35} & \underline{136.38}     & 27.00      & 1.07 & \underline{155.85}  & \textbf{27.16}      & \textbf{1.43} & \underline{148.77} & \textbf{27.97}      & \underline{1.52} \\
                          &                                   & Adaptformer                                 & 159.02     & 29.08      & 1.34 & 123.88    & 28.07      & 1.19 & \underline{174.17}      & 26.53      & 0.95 & 137.03     & 26.67      & 0.83 & 157.09  & 26.63      & 1.20 & 150.24 & 27.39      & 1.21 \\
                          &                                   & LT-SFT                                     & \underline{156.60}     & 23.76      & 0.59 & 119.75    & 25.33      & 0.70 & 191.01      & 25.96      & 0.49 & 144.57     & \textbf{28.01}      & \textbf{1.37} & 165.47  & \underline{26.89}      & 1.10 & 155.48 & 25.99      & 0.42 \\
                          &                                   & \textbf{SaRA (Ours)}                                        & \textbf{153.68}     & \underline{29.33}      & \textbf{1.63} & \textbf{116.69}    & \underline{28.24}      & \underline{1.94} & \textbf{138.64}      & \underline{26.63}     &\textbf{1.50} & \textbf{129.98}     & \underline{27.04}      & \underline{1.36} & \textbf{145.62}  & 26.40      & \underline{1.39} & \textbf{136.92} & \underline{27.53}      & \textbf{1.69} \\
                          \cline{2-21}
                          & \multirow{4}{*}{5M}               & LoRA                                        & \underline{163.80}     & \textbf{29.93}      & \underline{1.25} & \textbf{117.58}    & \textbf{28.32}      & \textbf{1.88} & 184.99      & \textbf{27.74}      & \textbf{1.25} & \underline{137.96}     & 27.10      & \underline{1.07} & \textbf{153.57}  & 26.93      & \textbf{1.40} & \underline{151.58} & \textbf{28.00}      & \textbf{1.44} \\
                          &                                   & Adaptformer                                 & 164.22     & 29.37      & 1.14 & 120.98    & 28.11      & 1.48 & \underline{184.84}      & 26.66      & 0.84 & 143.01     & \underline{27.35}      & 1.01 & 171.34  & 26.85      & 0.94 & 156.88 & 27.67      & 1.13 \\
                          &                                   & LT-SFT                                      & 169.24     & 24.23      & 0.08 & 127.01    & 25.43      & 0.03 & 202.47      & 26.90      & 0.68 & 153.49    & \textbf{27.96}      & 0.97 & 176.41  & \textbf{27.34}      & 1.00 & 165.72 & 26.37      & 0.27 \\
                          &                                   & \textbf{SaRA (Ours)}                                        & \textbf{153.69}     & \underline{29.39}      & \textbf{1.64} & \underline{118.74}    & \underline{28.17}      & \underline{1.72} & \textbf{174.86}      & \textbf{27.04}      & \underline{1.13} & \textbf{134.45}     & 27.06      & \textbf{1.18} & \underline{157.24}  & \underline{26.97}      & \underline{1.33} & \textbf{147.80} & \underline{27.73}      & \textbf{1.44} \\
                          \cline{2-21}
                           & 860M     & Full-finetune                                   & 147.81     & 27.77      & 1.65 & 120.22    & 27.84      & 1.47 & 136.49      & 25.10      & 0.95 & 129.07     & 26.75      & 1.21 & 134.86  & 24.64      & 1.00 & 133.69 & 26.42      & 1.30 \\
                           \midrule
\multirow{13}{*}{SD 2.0}  & \multirow{4}{*}{50M}              & LoRA                                        & \textbf{157.41}     & \underline{29.81}      & \underline{1.64} & \textbf{133.22}    & \underline{28.00}      & \underline{1.52}& 187.32      & \textbf{27.70}      & \underline{1.29} & \underline{148.18}     & \underline{27.58}      & \textbf{1.38} & 169.92  & \textbf{26.99}      & \textbf{1.09} & \underline{159.21} & \textbf{28.02}      & \underline{1.51} \\
                          &                                   & Adaptformer                                 & \underline{161.87}     & \textbf{30.78}      & \textbf{1.75} & 138.02    & 27.85      & 1.12 & \underline{179.44}      & \underline{27.35}      & 1.26 & 162.45     & 27.06      & 0.47 & 175.39  & 26.59      & 0.76 & 163.43 & \underline{27.93}      & 1.25 \\
                          &                                   & LT-SFT                                      & 164.80     & 28.13      & 0.59 & \underline{134.97}    & 26.40      & 0.59 & 183.23      & 25.90      & 0.50 & 153.94     & \textbf{27.88}      & \underline{1.33} & \underline{167.19}  & \underline{26.83}      & \underline{1.08} & 160.83 & 27.03      & 0.57 \\
                          &                                   & \textbf{SaRA (Ours)}                                        & 162.72    & 29.72      & 1.31 & 135.05    & \textbf{28.30}      & \textbf{1.55} & \textbf{151.82}      & 27.24      & \textbf{1.68} & \textbf{138.77}     & 26.30      & 0.96 & \textbf{165.62}  & 26.71      & 1.05 & \textbf{150.80} & 27.65      & \textbf{1.55} \\
                           \cline{2-21}
                          & \multirow{4}{*}{20M}              & LoRA                                        & \underline{161.92}     & 30.18      & \underline{1.52} & \textbf{129.01}    & \textbf{28.36}      & \textbf{2.00} & 190.90      & \underline{27.72}      & \underline{1.24} & \textbf{147.05}     & \underline{27.60}      & \textbf{1.44} & \underline{168.03}  & \underline{26.97}      & \textbf{1.13} & \underline{159.38} & \textbf{28.16}     & \underline{1.63} \\
                          &                                   & Adaptformer                                & \textbf{160.29}    & \textbf{30.42}      & \textbf{1.70} & 141.80    & 27.92     & 0.89 & \underline{190.57}      & 27.33      & 1.05 & 157.31     & 27.07      & 0.69 & 175.39  & 26.59      & 0.76 & 165.07 & 27.86      & 1.13 \\
                          &                                   & LT-SFT                                       & 168.09     & 28.29      & 0.47 & 135.03    & 26.47      & 0.62 & 194.17      & 26.64      & 0.66 & 155.51     & \textbf{27.88}      & \underline{1.27} & 174.64  & \textbf{27.12}      & 1.04 & 165.48 & 27.28      & 0.59 \\
                          &                                   & \textbf{SaRA (Ours)}                                        & 164.57     & \underline{30.22}     & 1.39 & \underline{134.28}    & \underline{28.29}      & \underline{1.60} & \textbf{163.67}      & \textbf{27.90}      & \textbf{1.79} & \underline{149.29}     & 27.01      & 0.98 & \textbf{165.62}  & 26.71      & \underline{1.05} & \textbf{155.49} & \underline{28.03}      & \textbf{1.68} \\
                          \cline{2-21}
                          & \multirow{4}{*}{5M}               & LoRA                                       & \underline{162.47}     & 29.91      & 1.39 & \textbf{132.35}    & \textbf{28.13}      & \textbf{1.65} & \underline{183.55}      & \textbf{27.68}      & \underline{1.34} & \underline{152.69}     & \underline{27.41}      & \textbf{1.09} & \textbf{164.00}  & 26.81      & \textbf{1.15} & \underline{159.01} & \underline{27.99}      & \underline{1.49} \\
                          &                                   & Adaptformer                                 & \textbf{162.25}     & \underline{30.52}      & \textbf{1.63} & 143.41    & 27.69      & 0.66 & 188.42      & 27.45      & 1.15 & 160.23     & 27.37      & 0.76 & 180.07  & 26.72      & 0.71 & 166.88 & 27.95      & 1.12 \\
                          &                                   & LT-SFT                                      & 175.45     & 28.74      & 0.23 & 137.84    & 26.55      & 0.46 & 209.51      & 27.29      & 0.70 & 161.67     & \textbf{27.90}      & 1.03 & 186.69  & \textbf{27.62}     & 1.00 & 174.23 & 27.62      & 0.52 \\
                          &                                   & \textbf{SaRA (Ours)}                                        & 165.57     & \textbf{30.58}      & \underline{1.47} & \underline{136.89}   & \underline{27.77}     & \underline{1.15} & \textbf{174.73}     & \underline{27.60}      & \textbf{1.45} & \textbf{150.89}     & 27.29      & \underline{1.09} & \underline{166.40}  & \underline{26.90}      & \underline{1.13} & \textbf{158.90} & \textbf{28.03}      & \textbf{1.53} \\
                          \cline{2-21}
                          &866M       & Full-finetune                                 & 160.87     & 29.30      & 1.25 & 133.19    & 28.33      & 1.70 & 198.45      & 25.81      & 0.19 & 137.84     & 26.74      & 1.27 & 145.99  & 25.64      & 1.00 & 155.27 & 27.16      & 0.93 \\
                         \midrule
\multirow{13}{*}{SD 3.0}  & \multirow{4}{*}{50M}              & LoRA                                        & \underline{165.22}     & 29.57      & 1.28 & \textbf{123.59}    & \underline{28.38}      & \textbf{1.59} & 187.26      & \underline{27.54}      & \underline{1.36} & \underline{148.38}     & \underline{26.83}      & \underline{1.50} & \underline{169.00}  & \underline{26.96}      & \underline{1.35} & \underline{158.69} & \underline{27.85}      & \underline{1.52} \\
                          &                                   & Adaptformer                                 & \textbf{164.09}     & \underline{29.81}      & \underline{1.43} & \underline{126.73}    & 28.23      & 1.38 & \underline{186.05}      & 27.14      & 0.93 & 156.77     & \textbf{27.12}      & \textbf{1.62} & 180.11  & 26.91      & 1.09 & 162.75 & 27.84      & 1.41 \\
                          &                                   & LT-SFT                                     & 209.04     & 29.45      & 0.39 & 158.59    & 27.72      & 0.10 & 208.94      & 27.03      & 0.14 & 204.16     & 26.02      & 0.00 & 189.26  & 26.06      & 0.38 & 194.00 & 27.26      & 0.18 \\
                         
                          &                                   & \textbf{SaRA (Ours)}                                        & 170.73     & \textbf{30.63}      & \textbf{1.73} & 128.35    & \textbf{28.45}      & \underline{1.50} & \textbf{179.87}      & \textbf{27.63}      & \textbf{1.68} & \textbf{137.92}    & 26.47      & 1.35 & \textbf{158.86}  & \textbf{27.10}     & \textbf{1.65} & \textbf{155.15} & \textbf{28.06}      & \textbf{1.77} \\
                           \cline{2-21}
                          & \multirow{4}{*}{20M}              & LoRA                                        & \textbf{156.23}     & 30.18      & \textbf{1.77} & \textbf{123.12}    & \underline{28.22}      & \textbf{1.48} & 187.14      & \underline{27.76}      & \underline{1.62} & \textbf{143.59}     & \textbf{26.96}      & \textbf{1.68} & 174.62  & \underline{26.63}      & \underline{1.03} & \textbf{156.94} & \underline{27.95}      & \textbf{1.64} \\
                          &                                   & Adaptformer                                 & 174.32     & \underline{30.30}      & 1.49 & 128.73    & \textbf{28.31}      & \underline{1.39} & \textbf{175.60}      & \textbf{27.77}      & \textbf{1.96} & 150.69     & 26.50     & 1.19 & \underline{174.61}  & 26.56      & 0.99 & 160.79 & 27.89      & 1.50 \\
                          &                                   & LT-SFT                                      & 167.02     & 29.19      & 1.05 & 154.04    & 27.67      & 0.19 & 203.23      & 26.91      & 0.16 & 155.68     & 26.48      & 1.10 & 177.42  & 26.21      & 0.71 & 171.48 & 27.29      & 0.77 \\
                          
                          &                                   & \textbf{SaRA (Ours)}                                        & \underline{166.21}     & \textbf{30.41}      & \underline{1.70} & \underline{126.69}    & 28.19      & 1.35 & \underline{180.74}      & 27.31      & 1.28 & \underline{150.15}     & \underline{26.88}      & \underline{1.52} & \textbf{163.78}  & \textbf{27.01}      & \textbf{1.48} & \underline{157.52} & \textbf{27.96}      & \textbf{1.64} \\
                           \cline{2-21}
                          & \multirow{4}{*}{5M}               & LoRA                                        & \underline{161.80}     & 30.14      & \underline{1.64} & \textbf{124.17}    & \underline{28.06}      & \underline{1.33} & \textbf{174.66}      & 27.27      & \textbf{1.40} & \textbf{149.85}     & 27.01      & 1.63 & \textbf{172.56}  & 26.88      & \underline{1.23} & \textbf{156.61} & 27.87      & \underline{1.59} \\
                          &                                   & Adaptformer                                 & 168.98     & \underline{30.50}      & \textbf{1.69} & 127.35    & 27.89      & 1.11 & 204.69      & \textbf{27.71}      & 1.05 & 158.60     & 27.03      & 1.52 & 182.22  & \underline{26.88}      & 1.03 & 168.37 & \underline{28.00}      & 1.40 \\
                          &                                   & LT-SFT                                      & \textbf{158.26}     & 29.29      & 1.27 & 134.81    & 27.69      & 0.75 & \underline{181.68}      & 27.27      & 1.20 & \underline{153.52}     & \textbf{27.20}      & \textbf{1.74} & 193.25  & 26.61      & 0.63 & \underline{164.30} & 27.61      & 1.20 \\
                          &                                   & \textbf{SaRA (Ours)}                                        & 174.42     & \textbf{30.60}      & 1.64 & \underline{125.14}    & \textbf{28.91}      & \textbf{1.94} & 194.79      & \underline{27.63}      & \underline{1.24} & 157.20     & \underline{27.17}      & \underline{1.65} & \underline{181.39}  & \textbf{27.20}      & \textbf{1.24} & 166.59 & \textbf{28.30}      & \textbf{1.68} \\
                          \cline{2-21}
& 2085M &Full-finetune                                        & 162.33     & 28.69      & 0.88 & 151.57    & 27.59      & 0.20 & 174.12      & 27.16      & 1.29 & 135.28     & 26.09      & 1.06 & 144.56  & 25.58      & 1.00 & 153.57 & 27.02      & 1.00\\
                                \bottomrule
\end{tabular}}
\vspace{-0.1in}
\caption{\hut{Comparison with different parameter-efficient fine-tuning methods on Stable Diffusion 1.5, 2.0, and 3.0. For most of the conditions, our model achieves the best FID and VLHI score, indicating that our model learns domain-specific knowledge successfully while keeping the prior information well. \textbf{Bold} and \underline{underline} represent \reb{the best and second best} results, respectively.}}
\vspace{-0.10in}
\label{tab:quantitative comparison on finetuning}
\end{table}


\begin{figure}[t]
\centering
\includegraphics[width=0.7\textwidth]{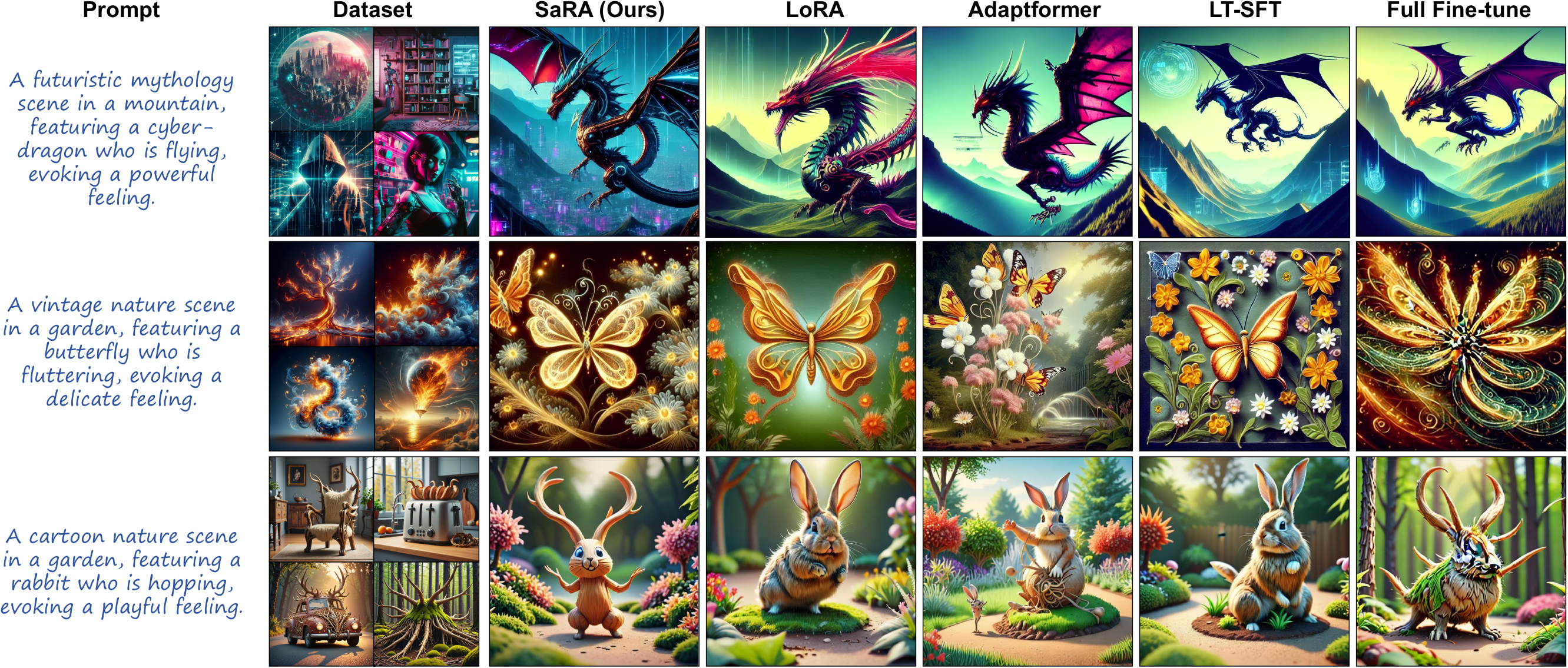}
\vspace{-0.15in}
\caption{\reb{Comparison of the generated images between different PEFT methods.}}
\vspace{-0.2in}
\label{fig:main paper fig}
\end{figure}

\subsection{Model Fine-tuning \yr{on Downstream Tasks}}
\label{sec:model finetuning}
{\bf Downstream Dataset Fine-tuning.} 
In this experiment, we choose 5 widely-used datasets from CIVITAI\footnote{https://civitai.com/articles/2138/lora-datasets-training-data-list-civitai-dataset-guide} with 5 different styles to conduct the fine-tuning experiments, which are Barbie Style, Cyberpunk Style, Elementfire Style, Expedition Style and Hornify Style. To comprehensively compare PEFT methods, we conduct three sets of experiments for each PEFT method on Stable Diffusion 1.5, 2.0, and 3.0, \yr{with selected} trainable parameter sizes of 50M, 20M, and 5M. 
We compute the FID score, CLIP score for the generated data, along with VLHI, which measures both the style (FID) and generalization (CLIP score). 
The quantitative results are shown in 
Tab.~\ref{tab:quantitative comparison on finetuning}, \yr{from which} we can draw the following conclusions:
\textbf{\textit{1)}} Our model can always achieve the best VLHI \yr{on average across five datasets}, indicating that our model can preserve the prior information in the pre-trained model well (a good CLIP score), while learning as much task-\yr{specific} knowledge as possible (a good FID), outperforming all the other PEFT methods and full-finetune method;
\textbf{\textit{2)}} As the \yr{number of} learnable parameter \yr{increases}, our model can learn more task-\yr{specific} knowledge (\yr{better} FID), 
\yr{but may lose} part of the prior information (lower CLIP score);
\textbf{\textit{3)}} For Stable Diffusions 1.5 and 2.0, our model achieves the best FID and usually the \reb{second best} CLIP score \yr{on average across five datasets, and} under different parameter numbers; 
while for Stable Diffusion 3.0, which has much more parameters than SD 1.5 and 2.0, our model achieves the best CLIP score and usually the \reb{second best} FID \yr{on average across five datasets}. 
\yr{The results} indicate that for a larger pre-trained model, more learnable parameters are needed to learn the task-\yr{specific} knowledge well.  \reb{Moreover, we provide some qualitative comparisons in Fig.~\ref{fig:main paper fig}, which shows the superior generation quality of our method (See appendix for more details.).}





\label{sec:downstream tasks}
\begin{wraptable}{r}{0.5\linewidth}
\centering
\vspace{-0.15in}
\renewcommand{\arraystretch}{1.1}
\resizebox{1.0\linewidth}{!}{
\begin{tabular}{l|ccc}
\toprule
\pzo \pzo \pzo \pzo Method                            & FID $\downarrow$ & CLIP Score $\uparrow$   & VLHI $\uparrow$ \\
\midrule
w/o. $PPA$ \&$\mathcal{L}_{rank}$ & 134.75   & 27.01     & 1.16  \\
w. $PPA$\yr{, w/o. $\mathcal{L}_{rank}$}                         & \underline{130.95} & 26.66 & \underline{1.56} \\
w. $\mathcal{L}_{rank}$\yr{, w/o. $PPA$}           & 135.31 & \underline{27.12}     & 0.89 \\
\textbf{w. $\mathbf{PPA}$ \&$\mathbf{\mathcal{L}_{rank}}$ (Ours)}  & 131.56 & \textbf{27.34} & \textbf{1.79} \\
Tuning Largest Parameters         & \textbf{130.55}   & 25.42     & 1.00        \\
Tuning Random Parameters        & 133.57   & 26.58     & 0.97 \\
\bottomrule
\end{tabular}}
\vspace{-0.15in}
\caption{Ablation studies on six ablated models.}
\vspace{-0.15in}
\label{tab:quantitative ablation studies}
\end{wraptable}


\subsection{Ablation Studies}
\hut{We conduct ablation studies to validate the effectiveness of our proposed modules: the progressive parameter adjustment (PPA), and the low-rank constrained loss ($\mathcal{L}_{rank}$). 
\yr{Then}, we \yr{further} assess the effectiveness of training parameters with the smallest absolute values, by comparing different parameter-selection strategies, including selecting the largest parameters and random parameters. 
We \yr{conduct downstream dataset finetuning} experiments 
using the Expedition dataset 
\yr{comparing} six \yr{ablated} models: 
\textit{1)} model without PPA and $\mathcal{L}_{rank}$, 
\textit{2)} model with PPA \yr{but without $\mathcal{L}_{rank}$}, 
\textit{3)} model with $\mathcal{L}_{rank}$ \yr{but without PPA}, 
\textit{4)} model with both PPA and $\mathcal{L}_{rank}$ (Ours), 
\textit{5)} model \yr{fine-}tuned with the largest \yr{absolute values} parameters, 
and \textit{6)} model \yr{fine-}tuned with randomly selected parameters.}
The \yr{quantitative metric} results are presented in Tab.~\ref{tab:quantitative ablation studies}:
\yr{\textit{1)}} The model without both the PPA and $\mathcal{L}_{rank}$ results in a \yr{poor} FID and low CLIP score. 
\yr{\textit{2)}} Introducing PPA improves the FID but decreases the CLIP score, indicating its effectiveness in learning task-specific knowledge. 
\yr{\textit{3)}} Incorporating $\mathcal{L}_{rank}$ helps \yr{achieve} a better CLIP score, but results in a \yr{worse} FID, indicating \yr{its effectivenes in } better preserving the model prior knowledge, \yr{but with} a loss of task-specific information. 
\yr{\textit{4)}} Regarding parameter-selection strategies, \yr{fine-}tuning the largest \yr{absolute values} parameters yields a relatively good FID but the worst CLIP score, suggesting that \yr{fine-}tuning the most effective parameters severely disrupts the model's prior knowledge \yr{and leads to worse content-text consistency}. 
\yr{\textit{5)}} Moreover, \yr{fine-}tuning \yr{r}andomly select\yr{ed} parameters results in both poor FID and CLIP scores, indicating \yr{randomly selecting parameters to finetune is unable} to learn task-specific knowledge and preserve the model's prior.
\yr{\textit{6)}} In contrast, our model 
achieves the best VLHI, validating its effectiveness in both fitting capability and prior preservation. More analysis of the hyperparameters is presented in the appendix.

\section{Conclusion}

In this paper, we propose SaRA, a novel parameter-efficient fine-tuning method, which makes full use of the ineffective parameters \yr{with the smallest absolute values} in the pre-trained model. 
We \yr{propose} a nuclear norm-based low-rank loss to constrain the rank of the learned sparse matrices, thereby avoiding model overfitting. 
Moreover, we design a progressive parameter adjustment strategy, which can further improve the effectiveness of the fine-tuned parameters. 
Finally, we propose a novel unstructural backpropagation method, largely saving the memory cost during parameter fine-tuning, which can 
\yr{also reduce the memory costs for other selective PEFT methods.}
Extensive experiments demonstrate the effectiveness of our method, which achieves the best fitting capability while keeping the prior information of the pre-trained model well. 

%
\section*{Acknowledgments}
This work was supported by National Natural Science Foundation of China (No. 62302297,72192821,62272447,62472282,62472285), Shanghai Sailing Program (22YF1420300), Young Elite Scientists Sponsorship Program by CAST (2022QNRC001), the Fundamental Research Funds for the Central Universities (YG2023QNB17, YG2024QNA44), Beijing Natural Science Foundation (L222117).

\bibliography{iclr2025_conference}

\begin{thebibliography}{49}
\providecommand{\natexlab}[1]{#1}
\providecommand{\url}[1]{\texttt{#1}}
\expandafter\ifx\csname urlstyle\endcsname\relax
  \providecommand{\doi}[1]{doi: #1}\else
  \providecommand{\doi}{doi: \begingroup \urlstyle{rm}\Url}\fi

\bibitem[Aghajanyan et~al.(2020)Aghajanyan, Zettlemoyer, and Gupta]{aghajanyan2020intrinsic}
Armen Aghajanyan, Luke Zettlemoyer, and Sonal Gupta.
\newblock Intrinsic dimensionality explains the effectiveness of language model fine-tuning.
\newblock \emph{arXiv preprint arXiv:2012.13255}, 2020.

\bibitem[Ansell et~al.(2021)Ansell, Ponti, Korhonen, and Vuli{\'c}]{ansell2021Lt-sft}
Alan Ansell, Edoardo~Maria Ponti, Anna Korhonen, and Ivan Vuli{\'c}.
\newblock Composable sparse fine-tuning for cross-lingual transfer.
\newblock \emph{arXiv preprint arXiv:2110.07560}, 2021.

\bibitem[Ansell et~al.(2024)Ansell, Vuli{\'c}, Sterz, Korhonen, and Ponti]{SpIEL}
Alan Ansell, Ivan Vuli{\'c}, Hannah Sterz, Anna Korhonen, and Edoardo~M Ponti.
\newblock Scaling sparse fine-tuning to large language models.
\newblock \emph{arXiv preprint arXiv:2401.16405}, 2024.

\bibitem[Bhardwaj et~al.(2024)Bhardwaj, Pandey, Priyadarshi, Ganapathy, Esteves, Kadambi, Borse, Whatmough, Garrepalli, Van~Baalen, et~al.]{bhardwaj2024shira}
Kartikeya Bhardwaj, Nilesh~Prasad Pandey, Sweta Priyadarshi, Viswanath Ganapathy, Rafael Esteves, Shreya Kadambi, Shubhankar Borse, Paul Whatmough, Risheek Garrepalli, Mart Van~Baalen, et~al.
\newblock Sparse high rank adapters.
\newblock In \emph{NIPS}, 2024.

\bibitem[Blattmann et~al.(2023)Blattmann, Rombach, Ling, Dockhorn, Kim, Fidler, and Kreis]{blattmann2023alignyoulatents}
Andreas Blattmann, Robin Rombach, Huan Ling, Tim Dockhorn, Seung~Wook Kim, Sanja Fidler, and Karsten Kreis.
\newblock Align your latents: High-resolution video synthesis with latent diffusion models.
\newblock In \emph{Proceedings of the IEEE/CVF Conference on Computer Vision and Pattern Recognition}, pp.\  22563--22575, 2023.

\bibitem[Chen et~al.(2022)Chen, Ge, Tong, Wang, Song, Wang, and Luo]{chen2022adaptformer}
Shoufa Chen, Chongjian Ge, Zhan Tong, Jiangliu Wang, Yibing Song, Jue Wang, and Ping Luo.
\newblock Adaptformer: Adapting vision transformers for scalable visual recognition.
\newblock \emph{Advances in Neural Information Processing Systems}, 35:\penalty0 16664--16678, 2022.

\bibitem[Deng et~al.(2009)Deng, Dong, Socher, Li, Li, and Fei-Fei]{deng2009imagenet}
Jia Deng, Wei Dong, Richard Socher, Li-Jia Li, Kai Li, and Li~Fei-Fei.
\newblock Imagenet: A large-scale hierarchical image database.
\newblock In \emph{2009 IEEE conference on computer vision and pattern recognition}, pp.\  248--255. Ieee, 2009.

\bibitem[Ding et~al.(2023)Ding, Lv, Wang, Chen, Zhou, Liu, and Sun]{ding2023sparse}
Ning Ding, Xingtai Lv, Qiaosen Wang, Yulin Chen, Bowen Zhou, Zhiyuan Liu, and Maosong Sun.
\newblock Sparse low-rank adaptation of pre-trained language models.
\newblock \emph{arXiv preprint arXiv:2311.11696}, 2023.

\bibitem[Edalati et~al.(2022)Edalati, Tahaei, Kobyzev, Nia, Clark, and Rezagholizadeh]{edalati2022krona}
Ali Edalati, Marzieh Tahaei, Ivan Kobyzev, Vahid~Partovi Nia, James~J Clark, and Mehdi Rezagholizadeh.
\newblock Krona: Parameter efficient tuning with kronecker adapter.
\newblock \emph{arXiv preprint arXiv:2212.10650}, 2022.

\bibitem[Fang et~al.(2024)Fang, Wang, Yi, and Ma]{fang2024dropout}
Zhengyi Fang, Yue Wang, Ran Yi, and Lizhuang Ma.
\newblock Dropout mixture low-rank adaptation for visual parameters-efficient fine-tuning.
\newblock In \emph{European Conference on Computer Vision}. Springer, 2024.

\bibitem[Frankle \& Carbin(2018)Frankle and Carbin]{frankle2018lottery}
Jonathan Frankle and Michael Carbin.
\newblock The lottery ticket hypothesis: Finding sparse, trainable neural networks.
\newblock \emph{arXiv preprint arXiv:1803.03635}, 2018.

\bibitem[Gal et~al.(2022)Gal, Alaluf, Atzmon, Patashnik, Bermano, Chechik, and Cohen-Or]{textualinversion}
Rinon Gal, Yuval Alaluf, Yuval Atzmon, Or~Patashnik, Amit~H Bermano, Gal Chechik, and Daniel Cohen-Or.
\newblock An image is worth one word: Personalizing text-to-image generation using textual inversion.
\newblock \emph{arXiv preprint arXiv:2208.01618}, 2022.

\bibitem[Guo et~al.(2020)Guo, Rush, and Kim]{guo2020diffpruninig}
Demi Guo, Alexander~M Rush, and Yoon Kim.
\newblock Parameter-efficient transfer learning with diff pruning.
\newblock \emph{arXiv preprint arXiv:2012.07463}, 2020.

\bibitem[Guo et~al.(2023)Guo, Yang, Rao, Wang, Qiao, Lin, and Dai]{animatediff}
Yuwei Guo, Ceyuan Yang, Anyi Rao, Yaohui Wang, Yu~Qiao, Dahua Lin, and Bo~Dai.
\newblock Animatediff: Animate your personalized text-to-image diffusion models without specific tuning.
\newblock \emph{arXiv preprint arXiv:2307.04725}, 2023.

\bibitem[Han et~al.(2024)Han, Gao, Liu, Zhang, et~al.]{han2024PEFTsurvey}
Zeyu Han, Chao Gao, Jinyang Liu, Sai~Qian Zhang, et~al.
\newblock Parameter-efficient fine-tuning for large models: A comprehensive survey.
\newblock \emph{arXiv preprint arXiv:2403.14608}, 2024.

\bibitem[Hayou et~al.(2024)Hayou, Ghosh, and Yu]{hayou2024lora+}
Soufiane Hayou, Nikhil Ghosh, and Bin Yu.
\newblock Lora+: Efficient low rank adaptation of large models.
\newblock \emph{arXiv preprint arXiv:2402.12354}, 2024.

\bibitem[Heusel et~al.(2017)Heusel, Ramsauer, Unterthiner, Nessler, and Hochreiter]{heusel2017fid}
Martin Heusel, Hubert Ramsauer, Thomas Unterthiner, Bernhard Nessler, and Sepp Hochreiter.
\newblock Gans trained by a two time-scale update rule converge to a local nash equilibrium.
\newblock \emph{Advances in neural information processing systems}, 30, 2017.

\bibitem[Ho et~al.(2020)Ho, Jain, and Abbeel]{ddpm}
Jonathan Ho, Ajay Jain, and Pieter Abbeel.
\newblock Denoising diffusion probabilistic models.
\newblock \emph{Advances in neural information processing systems}, 33:\penalty0 6840--6851, 2020.

\bibitem[Houlsby et~al.(2019)Houlsby, Giurgiu, Jastrzebski, Morrone, De~Laroussilhe, Gesmundo, Attariyan, and Gelly]{houlsby2019parameter}
Neil Houlsby, Andrei Giurgiu, Stanislaw Jastrzebski, Bruna Morrone, Quentin De~Laroussilhe, Andrea Gesmundo, Mona Attariyan, and Sylvain Gelly.
\newblock Parameter-efficient transfer learning for nlp.
\newblock In \emph{International conference on machine learning}, pp.\  2790--2799. PMLR, 2019.

\bibitem[Hu et~al.(2021)Hu, Shen, Wallis, Allen-Zhu, Li, Wang, Wang, and Chen]{hu2021lora}
Edward~J Hu, Yelong Shen, Phillip Wallis, Zeyuan Allen-Zhu, Yuanzhi Li, Shean Wang, Lu~Wang, and Weizhu Chen.
\newblock Lora: Low-rank adaptation of large language models.
\newblock \emph{arXiv preprint arXiv:2106.09685}, 2021.

\bibitem[Huang et~al.(2023)Huang, Sun, Xie, Li, and Liu]{huang2023t2ibench}
Kaiyi Huang, Kaiyue Sun, Enze Xie, Zhenguo Li, and Xihui Liu.
\newblock T2i-compbench: A comprehensive benchmark for open-world compositional text-to-image generation.
\newblock \emph{Advances in Neural Information Processing Systems}, 36:\penalty0 78723--78747, 2023.

\bibitem[Karras et~al.(2017)Karras, Aila, Laine, and Lehtinen]{karras2017celebahq}
Tero Karras, Timo Aila, Samuli Laine, and Jaakko Lehtinen.
\newblock Progressive growing of gans for improved quality, stability, and variation.
\newblock \emph{arXiv preprint arXiv:1710.10196}, 2017.

\bibitem[Karras et~al.(2019)Karras, Laine, and Aila]{karras2019ffhq}
Tero Karras, Samuli Laine, and Timo Aila.
\newblock A style-based generator architecture for generative adversarial networks.
\newblock In \emph{Proceedings of the IEEE/CVF conference on computer vision and pattern recognition}, pp.\  4401--4410, 2019.

\bibitem[Kawar et~al.(2023)Kawar, Zada, Lang, Tov, Chang, Dekel, Mosseri, and Irani]{kawar2023imagic}
Bahjat Kawar, Shiran Zada, Oran Lang, Omer Tov, Huiwen Chang, Tali Dekel, Inbar Mosseri, and Michal Irani.
\newblock Imagic: Text-based real image editing with diffusion models.
\newblock In \emph{Proceedings of the IEEE/CVF Conference on Computer Vision and Pattern Recognition}, pp.\  6007--6017, 2023.

\bibitem[Lei et~al.(2023)Lei, Bai, Brahma, Ainslie, Lee, Zhou, Du, Zhao, Wu, Li, et~al.]{lei2023conditional}
Tao Lei, Junwen Bai, Siddhartha Brahma, Joshua Ainslie, Kenton Lee, Yanqi Zhou, Nan Du, Vincent Zhao, Yuexin Wu, Bo~Li, et~al.
\newblock Conditional adapters: Parameter-efficient transfer learning with fast inference.
\newblock \emph{Advances in Neural Information Processing Systems}, 36:\penalty0 8152--8172, 2023.

\bibitem[Li et~al.(2022)Li, Li, Xiong, and Hoi]{li2022blip}
Junnan Li, Dongxu Li, Caiming Xiong, and Steven Hoi.
\newblock Blip: Bootstrapping language-image pre-training for unified vision-language understanding and generation.
\newblock In \emph{International conference on machine learning}, pp.\  12888--12900. PMLR, 2022.

\bibitem[Liang et~al.(2021)Liang, Glossner, Wang, Shi, and Zhang]{liang2021pruningsurvey}
Tailin Liang, John Glossner, Lei Wang, Shaobo Shi, and Xiaotong Zhang.
\newblock Pruning and quantization for deep neural network acceleration: A survey.
\newblock \emph{Neurocomputing}, 461:\penalty0 370--403, 2021.

\bibitem[Liao et~al.(2023)Liao, Meng, and Monz]{liao2023pafi}
Baohao Liao, Yan Meng, and Christof Monz.
\newblock Parameter-efficient fine-tuning without introducing new latency.
\newblock \emph{arXiv preprint arXiv:2305.16742}, 2023.

\bibitem[Liu et~al.(2024)Liu, Wang, Yin, Molchanov, Wang, Cheng, and Chen]{liu2024dora}
Shih-Yang Liu, Chien-Yi Wang, Hongxu Yin, Pavlo Molchanov, Yu-Chiang~Frank Wang, Kwang-Ting Cheng, and Min-Hung Chen.
\newblock Dora: Weight-decomposed low-rank adaptation.
\newblock \emph{arXiv preprint arXiv:2402.09353}, 2024.

\bibitem[Loshchilov et~al.(2017)Loshchilov, Hutter, et~al.]{loshchilov2017adamw}
Ilya Loshchilov, Frank Hutter, et~al.
\newblock Fixing weight decay regularization in adam.
\newblock \emph{arXiv preprint arXiv:1711.05101}, 5, 2017.

\bibitem[Mahabadi et~al.(2021)Mahabadi, Ruder, Dehghani, and Henderson]{mahabadi2021parameter}
Rabeeh~Karimi Mahabadi, Sebastian Ruder, Mostafa Dehghani, and James Henderson.
\newblock Parameter-efficient multi-task fine-tuning for transformers via shared hypernetworks.
\newblock \emph{arXiv preprint arXiv:2106.04489}, 2021.

\bibitem[Mou et~al.(2024)Mou, Wang, Xie, Wu, Zhang, Qi, and Shan]{t2i-adapter}
Chong Mou, Xintao Wang, Liangbin Xie, Yanze Wu, Jian Zhang, Zhongang Qi, and Ying Shan.
\newblock T2i-adapter: Learning adapters to dig out more controllable ability for text-to-image diffusion models.
\newblock In \emph{Proceedings of the AAAI Conference on Artificial Intelligence}, volume~38, pp.\  4296--4304, 2024.

\bibitem[Pan et~al.(2024)Pan, Liu, Diao, Pi, Zhang, Han, and Zhang]{pan2024lisa}
Rui Pan, Xiang Liu, Shizhe Diao, Renjie Pi, Jipeng Zhang, Chi Han, and Tong Zhang.
\newblock Lisa: Layerwise importance sampling for memory-efficient large language model fine-tuning.
\newblock \emph{Advances in Neural Information Processing Systems}, 2024.

\bibitem[Pfeiffer et~al.(2020)Pfeiffer, Kamath, R{\"u}ckl{\'e}, Cho, and Gurevych]{pfeiffer2020adapterfusion}
Jonas Pfeiffer, Aishwarya Kamath, Andreas R{\"u}ckl{\'e}, Kyunghyun Cho, and Iryna Gurevych.
\newblock Adapterfusion: Non-destructive task composition for transfer learning.
\newblock \emph{arXiv preprint arXiv:2005.00247}, 2020.

\bibitem[Podell et~al.(2023)Podell, English, Lacey, Blattmann, Dockhorn, M{\"u}ller, Penna, and Rombach]{podell2023sdxl}
Dustin Podell, Zion English, Kyle Lacey, Andreas Blattmann, Tim Dockhorn, Jonas M{\"u}ller, Joe Penna, and Robin Rombach.
\newblock Sdxl: Improving latent diffusion models for high-resolution image synthesis.
\newblock \emph{arXiv preprint arXiv:2307.01952}, 2023.

\bibitem[Poole et~al.(2022)Poole, Jain, Barron, and Mildenhall]{dreamfusion}
Ben Poole, Ajay Jain, Jonathan~T Barron, and Ben Mildenhall.
\newblock Dreamfusion: Text-to-3d using 2d diffusion.
\newblock \emph{arXiv preprint arXiv:2209.14988}, 2022.

\bibitem[Radford et~al.(2021)Radford, Kim, Hallacy, Ramesh, Goh, Agarwal, Sastry, Askell, Mishkin, Clark, et~al.]{radford2021clip}
Alec Radford, Jong~Wook Kim, Chris Hallacy, Aditya Ramesh, Gabriel Goh, Sandhini Agarwal, Girish Sastry, Amanda Askell, Pamela Mishkin, Jack Clark, et~al.
\newblock Learning transferable visual models from natural language supervision.
\newblock In \emph{International conference on machine learning}, pp.\  8748--8763. PMLR, 2021.

\bibitem[Rombach et~al.(2022)Rombach, Blattmann, Lorenz, Esser, and Ommer]{stablediffusion}
Robin Rombach, Andreas Blattmann, Dominik Lorenz, Patrick Esser, and Bj{\"o}rn Ommer.
\newblock High-resolution image synthesis with latent diffusion models.
\newblock In \emph{Proceedings of the IEEE/CVF conference on computer vision and pattern recognition}, pp.\  10684--10695, 2022.

\bibitem[Ruiz et~al.(2023)Ruiz, Li, Jampani, Pritch, Rubinstein, and Aberman]{ruiz2023dreambooth}
Nataniel Ruiz, Yuanzhen Li, Varun Jampani, Yael Pritch, Michael Rubinstein, and Kfir Aberman.
\newblock Dreambooth: Fine tuning text-to-image diffusion models for subject-driven generation.
\newblock In \emph{Proceedings of the IEEE/CVF Conference on Computer Vision and Pattern Recognition}, pp.\  22500--22510, 2023.

\bibitem[Sun et~al.(2023)Sun, Zhang, Shao, Wang, Liu, Xie, and Liu]{sun2023dreamcraft3d}
Jingxiang Sun, Bo~Zhang, Ruizhi Shao, Lizhen Wang, Wen Liu, Zhenda Xie, and Yebin Liu.
\newblock Dreamcraft3d: Hierarchical 3d generation with bootstrapped diffusion prior.
\newblock \emph{arXiv preprint arXiv:2310.16818}, 2023.

\bibitem[Sung et~al.(2021)Sung, Nair, and Raffel]{sung2021fishmask}
Yi-Lin Sung, Varun Nair, and Colin~A Raffel.
\newblock Training neural networks with fixed sparse masks.
\newblock \emph{Advances in Neural Information Processing Systems}, 34:\penalty0 24193--24205, 2021.

\bibitem[Valipour et~al.(2022)Valipour, Rezagholizadeh, Kobyzev, and Ghodsi]{valipour2022dylora}
Mojtaba Valipour, Mehdi Rezagholizadeh, Ivan Kobyzev, and Ali Ghodsi.
\newblock Dylora: Parameter efficient tuning of pre-trained models using dynamic search-free low-rank adaptation.
\newblock \emph{arXiv preprint arXiv:2210.07558}, 2022.

\bibitem[Van~Le et~al.(2023)Van~Le, Phung, Nguyen, Dao, Tran, and Tran]{dreambooth}
Thanh Van~Le, Hao Phung, Thuan~Hoang Nguyen, Quan Dao, Ngoc~N Tran, and Anh Tran.
\newblock Anti-dreambooth: Protecting users from personalized text-to-image synthesis.
\newblock In \emph{Proceedings of the IEEE/CVF International Conference on Computer Vision}, pp.\  2116--2127, 2023.

\bibitem[Wang et~al.(2022)Wang, Agarwal, Mukherjee, Liu, Gao, Awadallah, and Gao]{wang2022adamix}
Yaqing Wang, Sahaj Agarwal, Subhabrata Mukherjee, Xiaodong Liu, Jing Gao, Ahmed~Hassan Awadallah, and Jianfeng Gao.
\newblock Adamix: Mixture-of-adaptations for parameter-efficient model tuning.
\newblock \emph{arXiv preprint arXiv:2205.12410}, 2022.

\bibitem[Watson(1992)]{watson1992subgradient}
G~Alistair Watson.
\newblock Characterization of the subdifferential of some matrix norms.
\newblock \emph{Linear Algebra Appl}, 170\penalty0 (1):\penalty0 33--45, 1992.

\bibitem[Yang et~al.(2023)Yang, Robeyns, Wang, and Aitchison]{yang2023bayesian}
Adam~X Yang, Maxime Robeyns, Xi~Wang, and Laurence Aitchison.
\newblock Bayesian low-rank adaptation for large language models.
\newblock \emph{arXiv preprint arXiv:2308.13111}, 2023.

\bibitem[Ye et~al.(2023)Ye, Zhang, Liu, Han, and Yang]{ipadapter}
Hu~Ye, Jun Zhang, Sibo Liu, Xiao Han, and Wei Yang.
\newblock Ip-adapter: Text compatible image prompt adapter for text-to-image diffusion models.
\newblock \emph{arXiv preprint arXiv:2308.06721}, 2023.

\bibitem[Zhang et~al.(2023{\natexlab{a}})Zhang, Rao, and Agrawala]{zhang2023controlnet}
Lvmin Zhang, Anyi Rao, and Maneesh Agrawala.
\newblock Adding conditional control to text-to-image diffusion models.
\newblock In \emph{Proceedings of the IEEE/CVF International Conference on Computer Vision}, pp.\  3836--3847, 2023{\natexlab{a}}.

\bibitem[Zhang et~al.(2023{\natexlab{b}})Zhang, Chen, Bukharin, He, Cheng, Chen, and Zhao]{zhang2023adalora}
Qingru Zhang, Minshuo Chen, Alexander Bukharin, Pengcheng He, Yu~Cheng, Weizhu Chen, and Tuo Zhao.
\newblock Adaptive budget allocation for parameter-efficient fine-tuning.
\newblock In \emph{International Conference on Learning Representations}. Openreview, 2023{\natexlab{b}}.

\end{thebibliography}
\bibliographystyle{iclr2025_conference}

\appendix

\clearpage

\section{Appendix overview}
Source code is available at \url{https://sjtuplayer.github.io/projects/SaRA}. This appendix provides additional analysis and experiments related to SaRA, including:
\begin{itemize}
    \item More implementation details (Sec.~\ref{sec: more implementation details});
    \item More comparison results on downstream dataset fine-tuning (Sec.~\ref{sec: more comparison results on downstream dataset fine-tuning});
    \item More comparison results on image customization (Sec.~\ref{sec: more comparison results on image customization});
    \item Comparison experiments on controllable video generation (Sec.~\ref{sec: controllable video generation});
    \item \reb{Scaling weight for SaRA parameter (Sec.~\ref{sec: Weights for SaRA Parameters});}
    \item \reb{Merging different SaRA parameters  (Sec.~\ref{sec:merge sara params});}
    \item \reb{More ablation studies  (Sec.~\ref{sec: more ablation studies});}
    \item Analysis on training efficiency (Sec.~\ref{sec: Analysis on Training efficiency});
    \item Further analysis to understand what SaRA have learned (Sec.~\ref{sec: further analysis to understand what we have learned});
    \item Hyperparameter analysis (Sec.~\ref{sec: hyperparameter analysis});
    \item More analysis on the learned matrix $\Delta P$ (Sec.~\ref{sec: more analysis on the learned weight matrix});
    \item  \reb{Limitations (Sec.~\ref{sec:limitation}).}
\end{itemize}

\section{More implementation details}
\label{sec: more implementation details}

\textbf{Metrics.} 
We evaluate the generation models by three metrics: 
\textit{\yr{1})} \textbf{Fréchet Inception Distance (FID)}~\citep{heusel2017fid} to measure the similarity between the generated image distribution and target image distribution\yr{, where a lower score indicates better similarity}; 
\textit{\yr{2})} \textbf{CLIP Score} to measure the matching degree between the given prompts and generated images with a CLIP L/14 backbone~\citep{radford2021clip}\yr{, where a higher score indicates better consistency}; 
\textit{3)} \reb{Additionally, since FID and CLIP scores exhibit a certain degree of mutual exclusivity in finetuning a text-to-image model to downstream tasks (\textit{i.e.,} an overfitted model will result in the best FID but the worst CLIP score),} we introduce a new metric, the \textbf{Visual-Linguistic Harmony Index (VLHI)}, which \yr{is calculated by adding} the normalized FID and CLIP scores, to balance the evaluation of style (FID) and the preservation of model priors (CLIP score)\yr{, where a higher score indicates better performance}. 

\textbf{Visual-Linguistic Harmony Index (VLHI).} \yr{We propose} VLHI \yr{to} evaluate both the style and the generalization of each PEFT method, by balancing FID and CLIP Score. For a group of FIDs $\{FID_i\}_{i=1}^{n}$ and CLIP Scores $\{CLIP_i\}_{i=1}^{n}$, we compute the normalized FID and CLIP Score as VLHI:
\begin{equation}
    \begin{aligned}
        VLHI_i=\frac{\max(\{FID_i\}_{i=1}^{n})-FID_i}{\max(\{FID_i\}_{i=1}^{n})-\min(\{FID_i\}_{i=1}^{n})}+\frac{CLIP_i-\min(\{CLIP_i\}_{i=1}^{n})}{\max(\{CLIP_i\}_{i=1}^{n})-\min(\{CLIP_i\}_{i=1}^{n})}
    \end{aligned}
\end{equation}

For downstream dataset fine-tuning experiments, we regard the methods in one Stable Diffusion version and one dataset as a group.

\reb{{\bf Dataset details.} In the downstream dataset fine-tuning experiments, we choose 5 widely-used datasets from CIVITAI with 5 different styles to conduct the fine-tuning experiments, which are Barbie Style, Cyberpunk Style, Elementfire Style, Expedition Style, and Hornify Style. Each dataset contains about $200\sim400$ images, and for each image, we employ BLIP model~\citep{li2022blip} to generate its text annotations.
The detailed number of images in each dataset is recorded in Tab.~\ref{tab: dataset number}
}

\begin{table}[h]
\centering
\begin{tabular}{c|ccccc}
\toprule
Dataset      & Barbie & Cyberpunk & ElementFile & Expedition & Hornify \\ 
Image Number & 316    & 440       & 156         & 396        & 
236\\
\bottomrule
\end{tabular}
\caption{\reb{The number of images in each dataset.}}
\label{tab: dataset number}
\end{table}

\textbf{Training Details.} We use AdamW~\citep{loshchilov2017adamw} optimizer to train the methods for 5000 iterations with batch size 4, with a cosine learning rate scheduler, where the initial learning rate $lr$ is calculated corresponds to the thresholds $\theta_t$: $lr=10^{-3} \times e^{-350 \theta_t}$ (refer to Sec.~\ref{sec:lr and th}). For the training images and labeled captions, we \yr{recaption} them by adding a prefix `$name\,style,\,$' ($name$ corresponds to the dataset name) before each caption, which is a common trick in fine-tuning Stable Diffusion models to a new domain.

\begin{wrapfigure}{r}{0.5\textwidth}
\vspace{-0.3in}
\begin{minipage}[t]{0.52\textwidth}
    \begin{algorithm}[H]
    \footnotesize
        \caption{SaRA Fine-tuning Pseudocode}
        \begin{algorithmic}[1]
            \State model = Initialize\_model()
                            \State \textcolor{gray}{\# optimizer = AdamW(model.parameters())} 
                \Statex \textbf{optimizer = \textcolor{green_code}{AdamW-SaRA}(model, \textcolor{green_code}{threshold = $\theta_t$})}
            \For{epoch = 1 to $N$}
                \For{each mini-batch $(x, y)$}
                    \State $y_{\text{pred}} = \text{model}(x; \theta)$
                    \State $loss = \text{Loss\_Func}(y_{\text{pred}}, y)$
                    \State $loss.\text{backward}()$
                    \State optimizer.\text{step}()
                \EndFor
            \EndFor
        \end{algorithmic}
        \label{alg:pseudocode}
    \end{algorithm}
    \end{minipage}
\end{wrapfigure}

\textbf{Implementation of SaRA.} To enable easy implementation of SaRA, we have efficiently encapsulated it, allowing users to perform SaRA-based fine-tuning by modifying just a single line of training code. As shown in Algorithm 1, we integrate SaRA into the optimizer class, so users only need to replace the original PyTorch optimizer with the SaRA optimizer to automatically initiate SaRA training (The code that needs to be modified is highlighted in green.). The learning rate will be automatically assigned based on the threshold $\theta_t$ if it is not specified.


\section{More comparison results on downstream dataset fine-tuning}
\label{sec: more comparison results on downstream dataset fine-tuning}

\begin{figure}[h]
\centering
\includegraphics[width=1.0\textwidth]{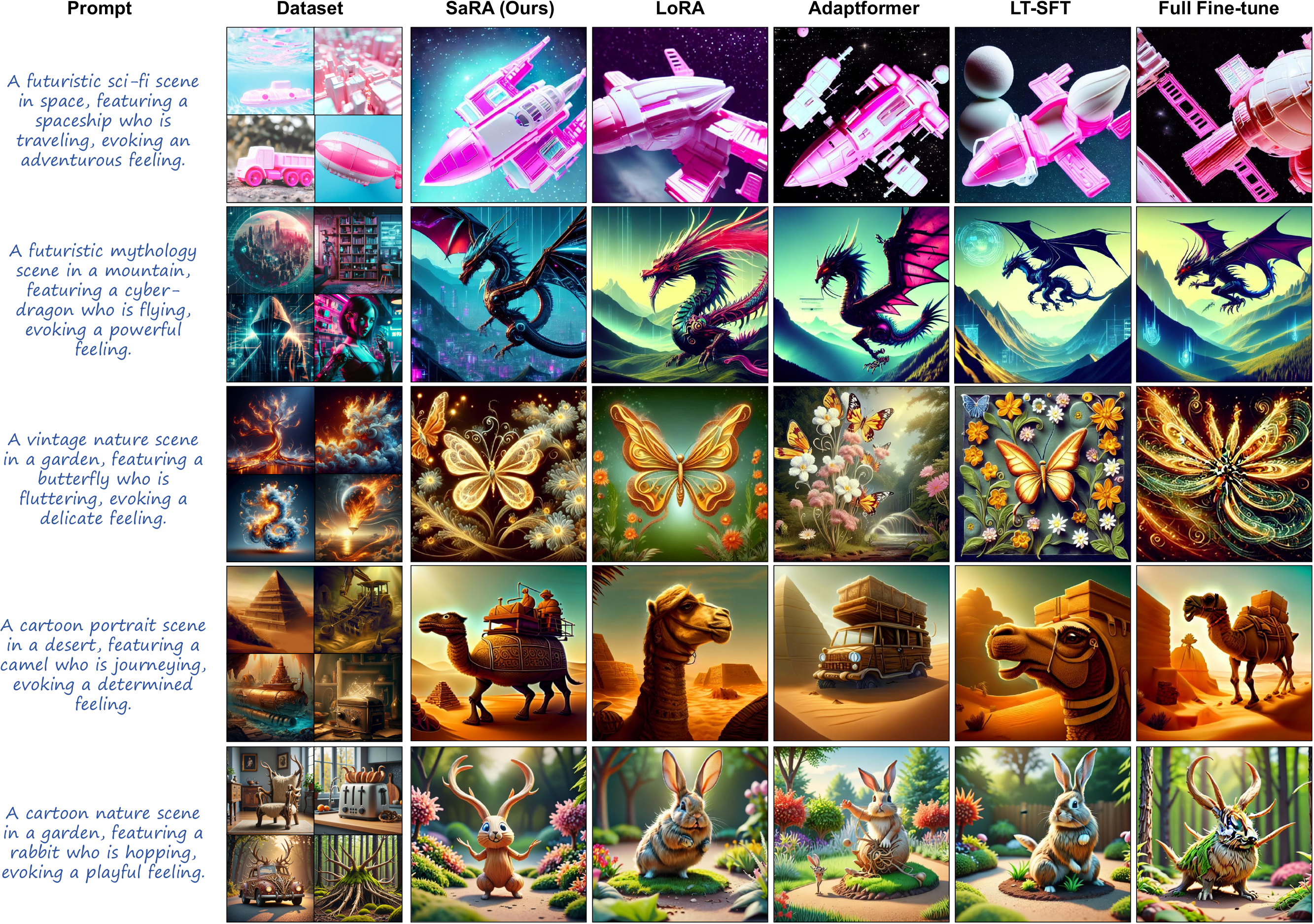}
\vspace{-0.2in}
\caption{Comparison results between different PEFT methods on Stable Diffusion 1.5.}
\label{fig:SD1.5-results}
\vspace{-0.15in}
\end{figure}

\begin{figure}[t]
\centering
\includegraphics[width=1.0\textwidth]{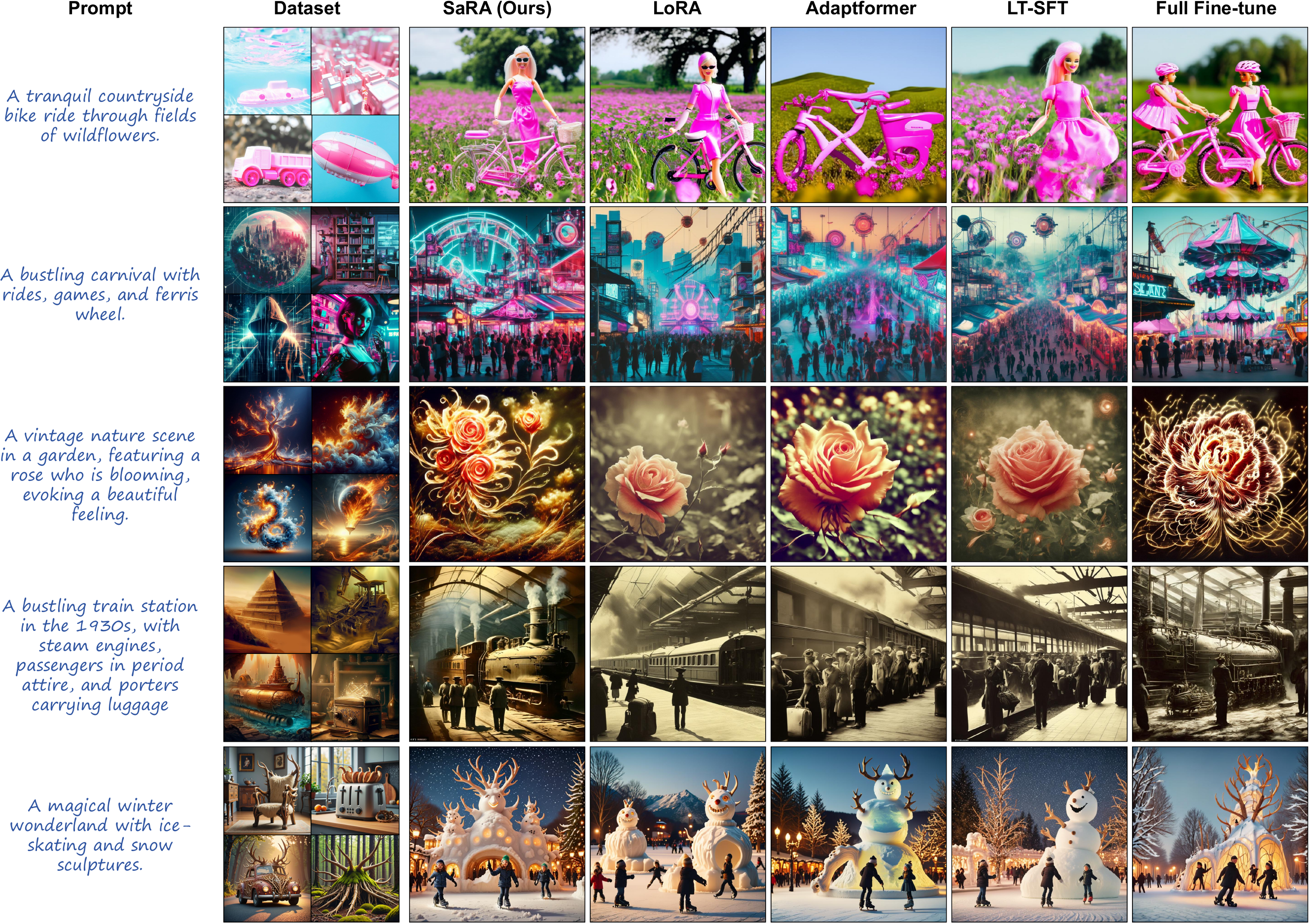}
\vspace{-0.2in}
\caption{Comparison results between different PEFT methods on Stable Diffusion 2.0.}
\label{fig:SD2.0-results}
\vspace{-0.1in}
\end{figure}

\begin{figure}[h]
\centering
\includegraphics[width=1.0\textwidth]{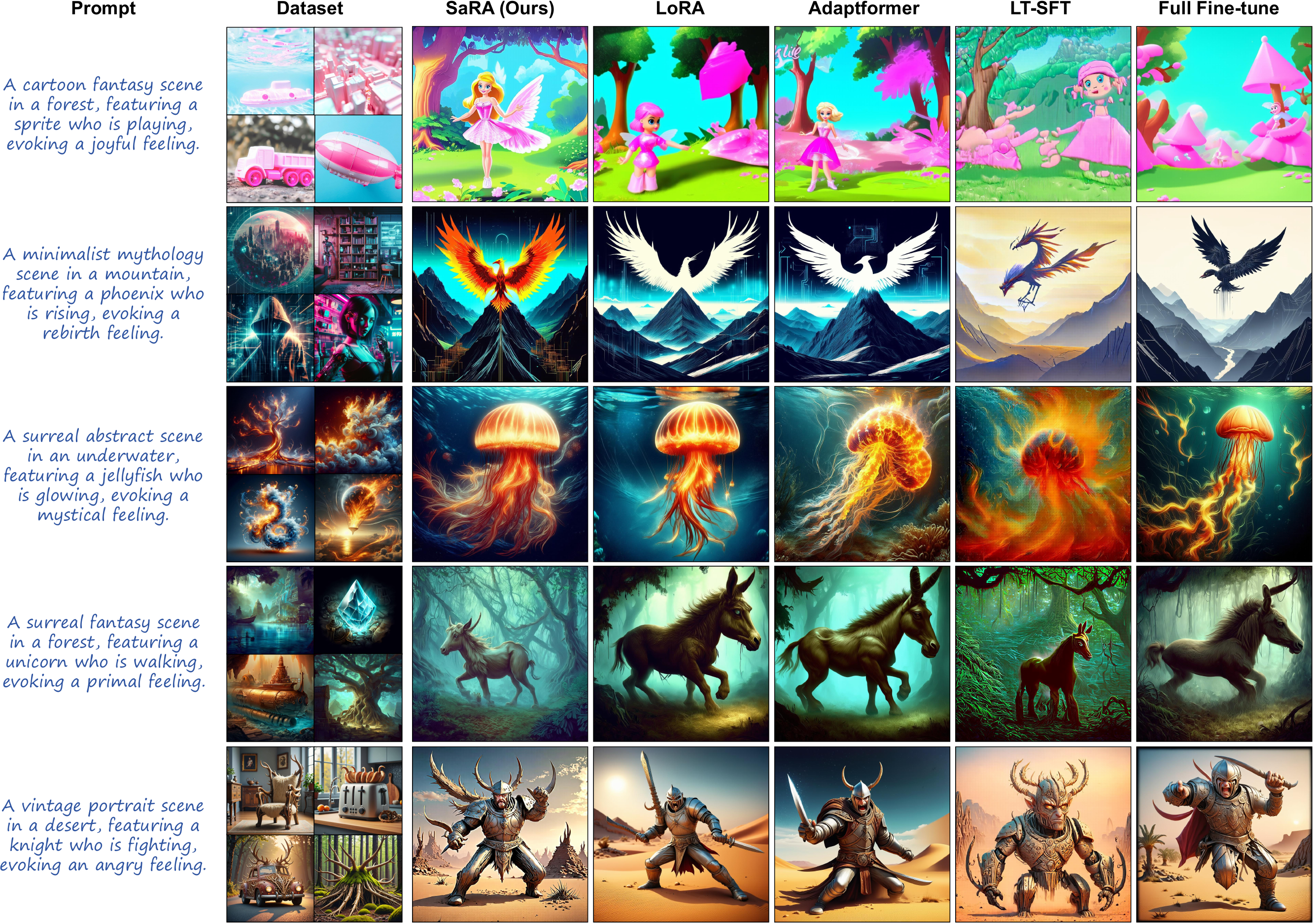}
\vspace{-0.2in}
\caption{Comparison results between different PEFT methods on Stable Diffusion 3.0.}
\label{fig:SD3.0-results}
\vspace{-0.1in}
\end{figure}

\begin{figure}[h]
\centering
\includegraphics[width=1.0\textwidth]{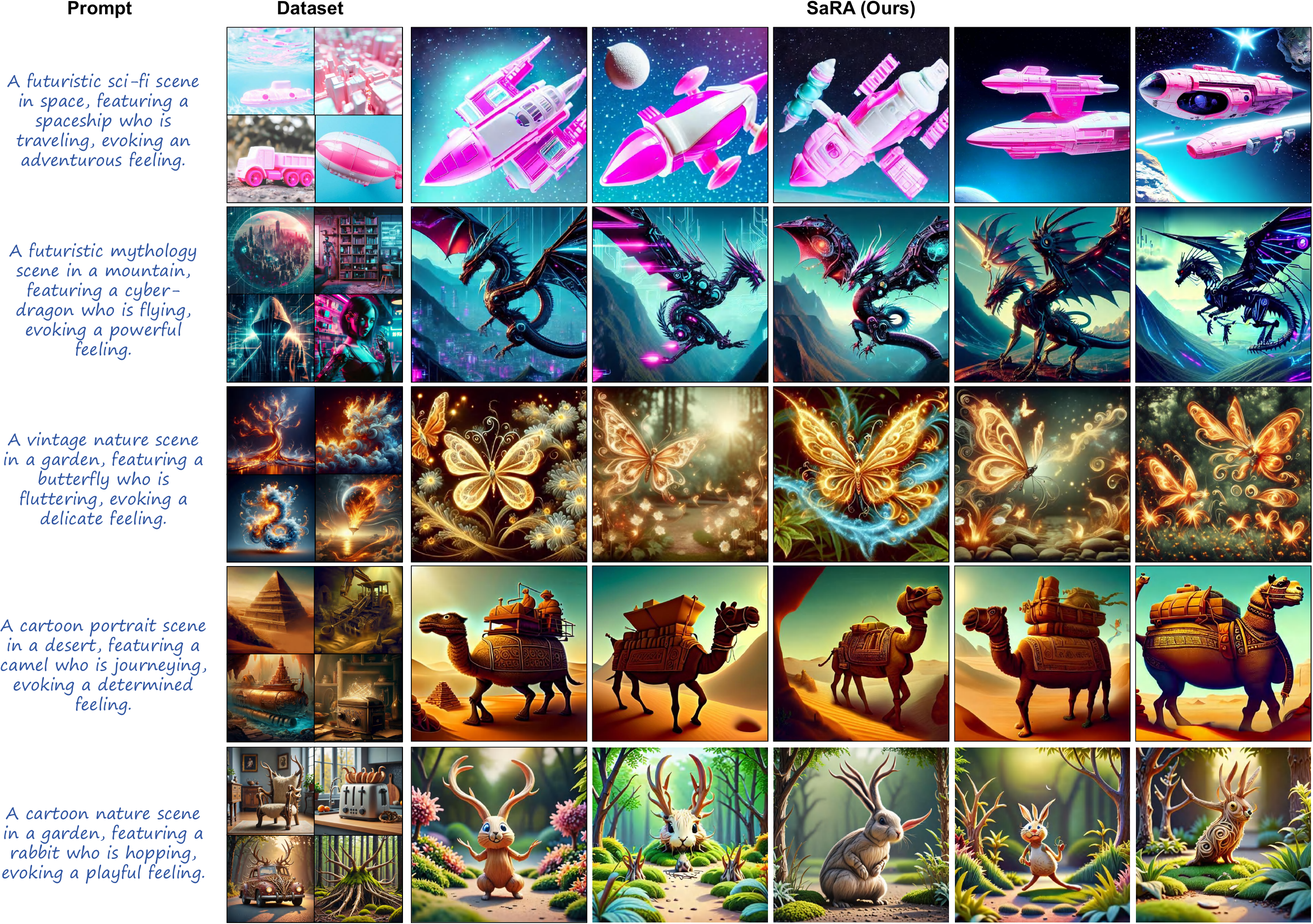}
\vspace{-0.2in}
\caption{\reb{More generation results by SaRA for different downstream datasets on SD 1.5.}}
\label{fig:more SD1.5-results}
\vspace{-0.1in}
\end{figure}

\begin{figure}[h]
\centering
\includegraphics[width=1.0\textwidth]{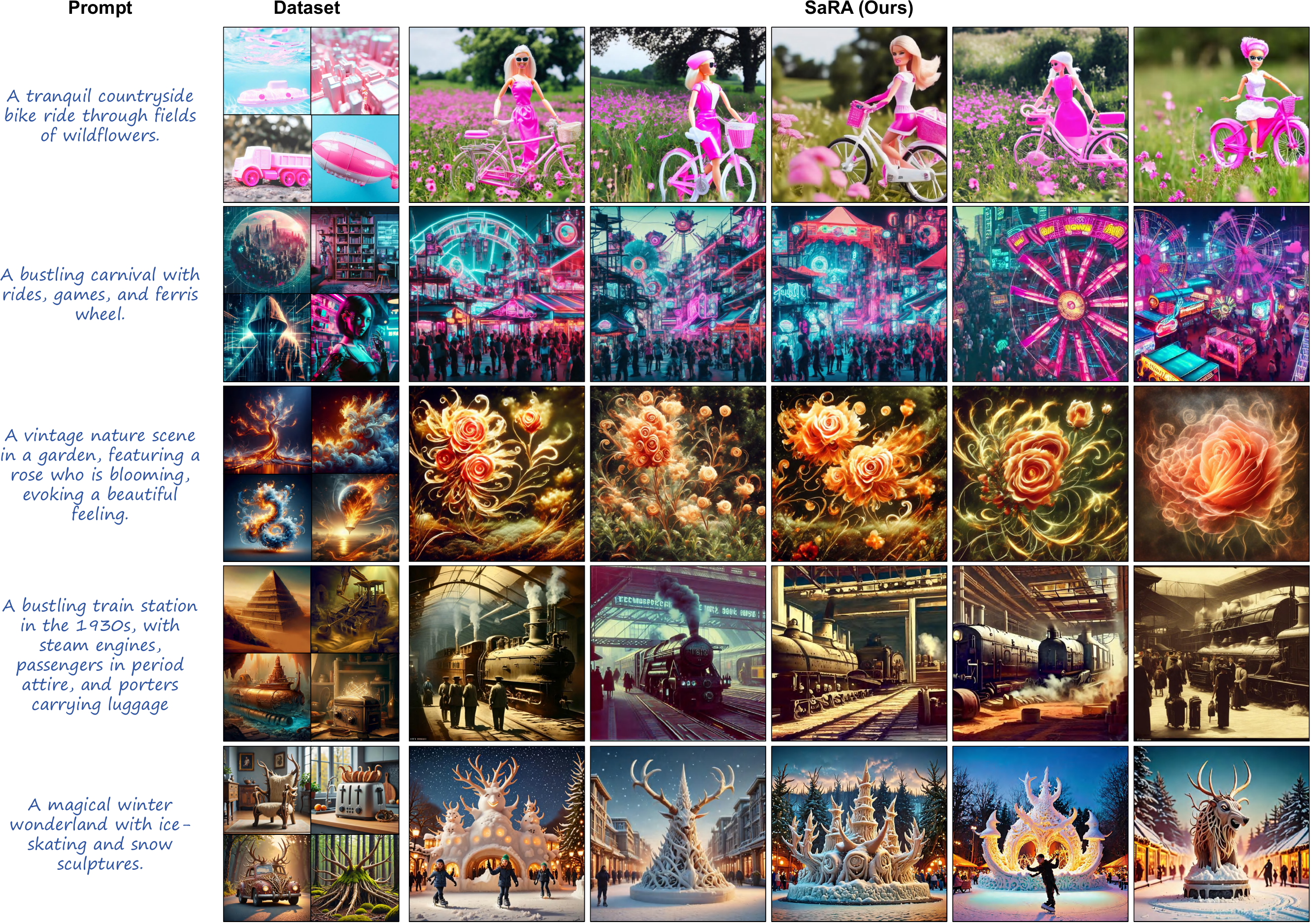}
\vspace{-0.2in}
\caption{\reb{More generation results by SaRA for different downstream datasets on SD 2.0.}}
\label{fig:more SD2.0-results}
\vspace{-0.1in}
\end{figure}

\begin{figure}[h]
\centering
\includegraphics[width=1.0\textwidth]{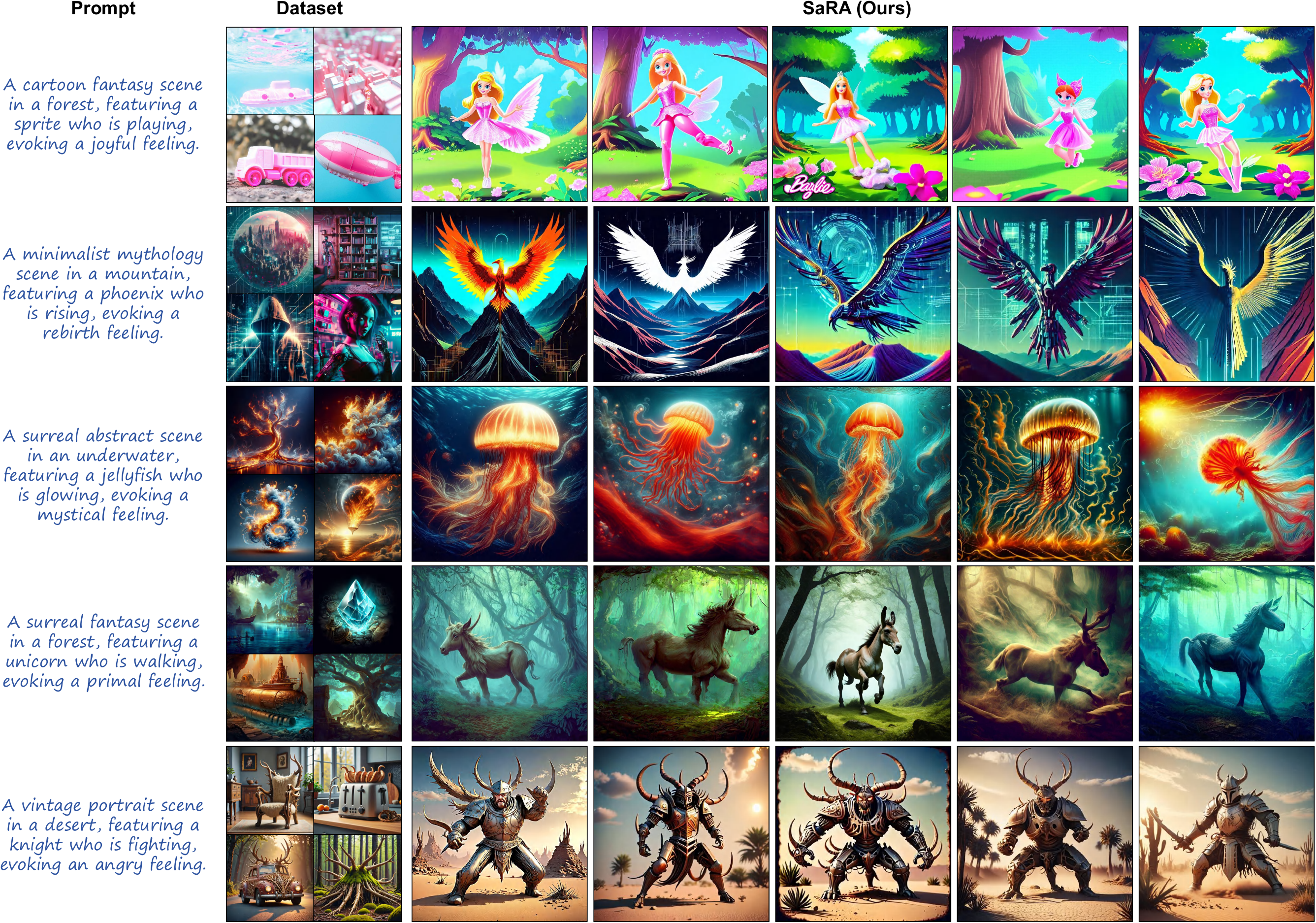}
\vspace{-0.2in}
\caption{\reb{More generation results by SaRA for different downstream datasets on SD 3.0.}}
\label{fig:more SD3.0-results}
\vspace{-0.1in}
\end{figure}

\reb{\textbf{Visualization results.}} We compare our model with LoRA~\citep{hu2021lora}, Adaptformer~\citep{chen2022adaptformer}, LT-SFT~\citep{ansell2021Lt-sft} and full-parameter finetuning method. We train all methods for 5,000 iterations and use the trained models to generate 500 images based on 500 text descriptions (generated by GPT-4). The quantitative results are presented in the main paper. In this section, we show more qualitative results on Stable Diffusion 1.5, 2.0, and 3.0 with resolutions 512, 768, and 1,024. The results from Stable Diffusion 1.5, 2.0, and 3.0 are shown in Figs.~\ref{fig:SD1.5-results}-~\ref{fig:SD3.0-results} respectively. 
It can be seen that our model generates images \yr{that} contain most of the features \yr{in} the target domain and \yr{are well} consistent with the given prompts under different datasets.
\reb{Moreover, to show the generation diversity of Our SaRA, we further generate more images by the trained SaRA weights on Stable Diffusion 1.5, 2.0, and 3.0, where for each SaRA weight, we generate 5 images with the same prompt and different random seeds. The generated results are shown in Fig.~\ref{fig:more SD1.5-results}-~\ref{fig:more SD3.0-results}. It can be seen that SaRA can generate the target-domain images with high diversity, while keeping the semantics consistent with the given prompts, demonstrating a good preservation of the model prior.}

\reb{
\textbf{More compared methods.} In this section, we compare our model with additional state-of-the-art parameter fine-tuning methods on Stable Diffusion 1.5, including DoRA~\citep{liu2024dora} and DiffPruning~\citep{guo2020diffpruninig}, which are the representative reparameterized PEFT and selective PEFT approaches, respectively. The comparison results are presented in Tab.~\ref{tab:compare with dora}. The results show that DoRA performs comparably to LoRA, while DiffPruning cannot learn enough tas-specified knowledge, which results in an extremely high FID. In contrast, our model achieves the best performance as evaluated by VLHI, attaining the lowest FID and competitive CLIP score. Moreover, to demonstrate the effectiveness of our SaRA method among various selective PEFT approaches, we compare SaRA with LT-SFT~\citep{ansell2021Lt-sft}, FishMask~\citep{sung2021fishmask}, DiffPruning~\citep{guo2020diffpruninig}, and an ablated method that fine-tunes the largest parameters on Stable Diffusion 1.5 with 50M trainable parameters. The qualitative comparison results are shown in Fig.~\ref{fig:comparison with more sft}. It can be observed that LT-SFT does not learn the target style well in the ElementFire and Horinfy datasets. FishMask tends to generate artifacts as it tunes some effective parameters in the pretrained weights, disrupting part of the model priors. DiffPruning fails to capture task-specific information, resulting in outputs that differ significantly from the target style (despite the fact that we have tried different hyperparameters). Additionally, the ablated model that fine-tunes the largest parameters tends to overfit, similar to the full-parameter fine-tuning model. Since the most important parameters are all fine-tuned, it is prone to overfitting to the target domain, leading to generated images that do not align well with the given prompts. In contrast, our SaRA fits the five datasets well while preserving the model priors, indicating superior performance among the different selective PEFT methods.}

\reb{\textbf{More experiments on Stable Diffusion XL.} In this section, we present additional comparison experiments on one of the most widely used stable diffusion models, Stable Diffusion XL 1.0~\citep{podell2023sdxl}, capable of generating images at a resolution of $1024\times 1024$. The results, summarized in Tab.~\ref{tab:quantitative comparison on SD xl.}, demonstrate that our SaRA consistently achieves the best performance on Stable Diffusion XL 1.0, further validating the effectiveness and robustness of our approach.}


\reb{\textbf{More evaluation metrics.} While the CLIP score~\citep{radford2021clip} measures overall similarity between images and text, it may overlook finer details during evaluation. To address this, we incorporate the attribute evaluation metric (denoted as B-VQA) from T2I-CompBench++~\citep{huang2023t2ibench}, which assesses the alignment between generated images and input text prompts at a more granular level. The comparison results are presented in Tab.~\ref{tab:T2Ibench on sd1.5}, showing that our model achieves the best or second-best B-VQA score in most cases, demonstrating its ability to preserve fine-grained details described in the input text prompts.}

\begin{table}[t]
\renewcommand{\arraystretch}{1.3}
\resizebox{1.0\linewidth}{!}{
\begin{tabular}{c|c|c|ccc|ccc|ccc|ccc|ccc|ccc}
\toprule
\multirow{2}{*}{Backbone} & \multirow{2}{*}{Params} & \multirow{2}{*}{Model} & \multicolumn{3}{c}{BarbieCore} & \multicolumn{3}{c}{Cyberpunk} & \multicolumn{3}{c}{ElementFire} & \multicolumn{3}{c}{Expedition} & \multicolumn{3}{c}{Hornify} & \multicolumn{3}{c}{Mean}     \\ 
                          &                        &                         & FID $\downarrow$    & CLIP $\uparrow$   & VLHI $\uparrow$ & FID $\downarrow$ & CLIP $\uparrow$    & VLHI $\uparrow$  & FID $\downarrow$  & CLIP $\uparrow$     & VLHI $\uparrow$  & FID $\downarrow$  & CLIP $\uparrow$    & VLHI $\uparrow$  & FID $\downarrow$ & CLIP $\uparrow$   & VLHI $\uparrow$ & FID $\downarrow$   & CLIP $\uparrow$     & VLHI $\uparrow$ \\ \midrule
\multirow{16}{*}{SD 1.5}  & \multirow{5}{*}{50M}              & DoRA & 158.40 & \underline{29.48}  & 1.43 & \underline{119.06} & \underline{28.16} & \underline{1.49} & 171.96 & \textbf{27.67} & \underline{1.41}  & \textbf{131.33}   & 26.94  & 1.24                               & 150.33  & \textbf{26.83}  & \textbf{1.44}  & \underline{146.22}  & \underline{27.82}  & \underline{1.53} \\&      & LoRA                       & 161.88     & \textbf{29.93}                                                 & \underline{1.34} & \textbf{117.49}    & \textbf{28.22}      & \textbf{1.62} & 181.66      & \underline{27.47}      & 1.20 &                                        136.31     & \underline{27.39}      & 1.32 & 156.36  & \underline{26.80}      &\underline{1.28} & 150.74 & \textbf{27.96}      & 
                          1.45 \\
                          &                                   & Adaptformer                                 & 166.09     & 29.00      & 1.00 & 126.21    & 27.13      & 0.64 & \underline{151.27}      & 26.57      & 1.29 & 138.01     & 26.41      & 0.63 & \underline{151.53}  & 26.20      & 1.18 & 146.62 & 27.06      & 1.18 \\
                          &                                   & LT-SFT                                      & \underline{157.80}     & 23.80      & 0.54 & 123.59    & 25.71     & 0.37 & 171.67      & 25.11      & 0.44 & 139.29     & \textbf{27.81}      & \underline{1.46} & 158.52  & 26.35      & 1.06 & 150.18 & 25.76      & 0.49\\
                          &                                   & \textbf{SaRA (Ours)}                                        & \textbf{148.54}     & 28.60      & \textbf{1.75} & 121.67    & 27.30      & 1.02 & \textbf{132.67}      & 26.77      & \textbf{1.63} & 131.56     & 27.34      & \textbf{1.48} & \textbf{140.36}  & 25.40      & 1.15 & \textbf{134.96} & 27.08      & \textbf{1.55} \\
                          \cline{2-21}
                          & \multirow{5}{*}{20M}    & DoRA & 158.85 & 29.22 & 1.37 & \textbf{116.23} & \underline{28.42}  & \textbf{1.78} & \underline{169.91}  & \underline{27.33}    & 1.31  & \underline{133.80}    & 26.86    & 1.09  & \underline{148.97}   & 26.82    & \textbf{1.47} & \underline{145.55} & \underline{27.73}  & 1.51 \\&           & LoRA                                        & 159.64     & \textbf{29.65}      & \underline{1.40} & 117.21    & \textbf{28.43}     & \underline{1.71} & 174.79      & \textbf{27.61}      & \underline{1.35} & 136.38     & 27.00      & 1.07 & 155.85  & \textbf{27.16}      & \underline{1.43} & 148.77 & \textbf{27.97}      & \underline{1.52} \\
                          &                                   & Adaptformer                                 & 159.02     & 29.08      & 1.34 & 123.88    & 28.07      & 1.11 & 174.17      & 26.53      & 0.95 & 137.03     & 26.67      & 0.83 & 157.09  & 26.63      & 1.20 & 150.24 & 27.39      & 1.21 \\
                          &                                   & LT-SFT                                     & \underline{156.60}     & 23.76      & 0.59 & 119.75    & 25.33      & 0.53 & 191.01      & 25.96      & 0.49 & 144.57     & \textbf{28.01}      & \textbf{1.37} & 165.47  & \underline{26.89}      & 1.10 & 155.48 & 25.99      & 0.42 \\
                          &                                   & \textbf{SaRA (Ours)}                                        & \textbf{153.68}     & \underline{29.33}      & \textbf{1.63} & \underline{116.69}    & 28.24      & 1.69 & \textbf{138.64}      & 26.63     &\textbf{1.50} & \textbf{129.98}     & \underline{27.04}      & \underline{1.36} & \textbf{145.62}  & 26.40      & 1.39 & \textbf{136.92} & 27.53      & \textbf{1.69} \\
                          \cline{2-21}
                          & \multirow{5}{*}{5M}       & DoRA & \underline{156.61} & 29.07   & \underline{1.45}  & \textbf{113.26} & 27.62    & \textbf{1.74}  & \underline{178.70} & \underline{27.57}    & \textbf{1.28}  & \underline{135.59} & 26.88    & 1.02  & 161.21 & \textbf{27.34}    & \underline{1.37}  & \underline{149.07} & 27.70    & 1.38 \\&          & LoRA                                        & 163.80     & \textbf{29.93}      & 1.25 & \underline{117.58}    & \textbf{28.32}      & \underline{1.65} & 184.99      & \textbf{27.74}      & \underline{1.25} & 137.96     & 27.10      & \underline{1.07} & \textbf{153.57}  & 26.93      & \textbf{1.40} & 151.58 & \textbf{28.00}      & \textbf{1.44} \\
                          &                                   & Adaptformer                                 & 164.22     & 29.37      & 1.14 & 120.98    & 28.11      & 1.33 & 184.84      & 26.66      & 0.84 & 143.01     & \underline{27.35}      & 1.01 & 171.34  & 26.85      & 0.94 & 156.88 & 27.67      & 1.13 \\
                          &                                   & LT-SFT                                      & 169.24     & 24.23      & 0.08 & 127.01    & 25.43      & 0.03 & 202.47      & 26.90      & 0.68 & 153.49    & \textbf{27.96}      & 0.97 & 176.41  & \textbf{27.34}      & 1.00 & 165.72 & 26.37      & 0.27 \\
                          &                                   & \textbf{SaRA (Ours)}                                        & \textbf{153.69}     & \underline{29.39}      & \textbf{1.64} & 118.74    & \underline{28.17}      & 1.52 & \textbf{174.86}      & 27.04      & 1.13 & \textbf{134.45}     & 27.06      & \textbf{1.18} & \underline{157.24}  & 26.97      & 1.33 & \textbf{147.80} & \underline{27.73}      & \textbf{1.44} \\
                          \cline{2-21}
                          & 10M     & DiffPruning     & 217.43 & 31,41 & 1.00 & 180.25 & 28.43
                           & 1.00 & 241.72 & 27.49 & 0.91 & 184.56 & 28.67 & 1.00 & 206.73 & 28.30   & 1.00 & 206.14 & 28.86 & 1.00\\ \cline{2-21}
                           
                           & 860M     & Full-finetune                                   & 147.81     & 27.77      & 1.65 & 120.22    & 27.84      & 1.47 & 136.49      & 25.10      & 0.95 & 129.07     & 26.75      & 1.21 & 134.86  & 24.64      & 1.00 & 133.69 & 26.42      & 1.30 \\ 
                           
                           \bottomrule

\end{tabular}
}
\vspace{-0.1in}
\caption{\reb{Comparison with different parameter-efficient fine-tuning methods (including additional DoRA and DiffPrune) on Stable Diffusion 1.5. For most of the conditions, our model achieves the best FID and VLHI score, indicating that our model learns domain-specific knowledge successfully while keeping the prior information well.}}
\label{tab:compare with dora}
\end{table}

\begin{table}[t]
\renewcommand{\arraystretch}{1.3}
\resizebox{1.0\linewidth}{!}{
\begin{tabular}{c|c|c|ccc|ccc|ccc|ccc|ccc|ccc}
\toprule
\multirow{2}{*}{Backbone} & \multirow{2}{*}{Params} & \multirow{2}{*}{Model} & \multicolumn{3}{c}{BarbieCore} & \multicolumn{3}{c}{Cyberpunk} & \multicolumn{3}{c}{ElementFire} & \multicolumn{3}{c}{Expedition} & \multicolumn{3}{c}{Hornify} & \multicolumn{3}{c}{Mean}     \\ 
                          &                        &                         & FID $\downarrow$    & CLIP $\uparrow$   & VLHI $\uparrow$ & FID $\downarrow$ & CLIP $\uparrow$    & VLHI $\uparrow$  & FID $\downarrow$  & CLIP $\uparrow$     & VLHI $\uparrow$  & FID $\downarrow$  & CLIP $\uparrow$    & VLHI $\uparrow$  & FID $\downarrow$ & CLIP $\uparrow$   & VLHI $\uparrow$ & FID $\downarrow$   & CLIP $\uparrow$     & VLHI $\uparrow$ \\ \midrule
\multirow{16}{*}{SD XL} & \multirow{5}{*}{50M} & DoRA    & \underline{164.42}    & \textbf{31.76}  & \textbf{1.77} & \underline{126.45} & \textbf{29.20} & \underline{1.76} & 175.78 &\underline{28.23} & 0.74 & 139.84 & \textbf{27.60} & 1.12 & 164.53 & \underline{27.29} & 0.90 & 154.20 & \textbf{28.82} & 1.06 \\ &
                                               & Lora     & 168.59   & \underline{31.68} & 1.51 & 132.38 & 28.96  & 1.26 & \underline{134.27} & 27.65 & \textbf{1.25}  & 130.37 &\underline{27.30} & \underline{1.34} & 154.78 & \textbf{27.32} & \textbf{1.38}  & 144.08 & \underline{28.58}         & \underline{1.45}          \\
                        &                      & Adaptformer & 171.33 & 30.69 & 1.06 & 135.74 & 28.71 & 0.83 & 139.71 & 27.34 & 0.92 & 135.68 & 27.11 & 0.98 & 151.20 & 26.94 & 1.15 & 146.73   & 28.16     & 1.06          \\
                        &                      & LT-SFT    & 165.41  & 30.20 & 1.24  & 131.08 & 28.65 & 1.16 & 140.62 & 27.48 & 0.97 & \underline{126.10} & 26.97  & 1.30 & \underline{150.94} & 27.11  & \underline{1.34}  & \underline{142.83}  & 28.08          & 1.21          \\
                        &                      & \textbf{SaRA (Ours)}     & \textbf{162.53}   & 30.67  & \underline{1.54} & \textbf{126.04} & \underline{29.01} & \textbf{1.79} & \textbf{129.92} & \textbf{28.73}  & \textbf{2.00} & \textbf{124.48} & 27.18 & \textbf{1.51} & \textbf{144.28} & 26.66 & 1.18 & \textbf{137.45} & 28.45        & \textbf{1.71} \\ \cline{2-21}
                        & \multirow{5}{*}{20M} & DoRA    & 165.18    & 31.41 & 1.62 & \textbf{124.22} & 28.95  & \textbf{1.75} & 177.07 & \textbf{28.25} & 0.72  & 138.72 & 27.64 & 1.20 & 163.20 & 27.28  &
                        0.95 & 153.68 & \underline{28.71} & 1.03 \\ &
                                               & Lora      & \underline{163.46}  & \underline{31.58}  & \underline{1.77} & 132.38 & 28.96 & 1.26 & \textbf{139.89} & 28.02 & \underline{1.31} & \textbf{131.63} & 27.52 & \textbf{1.43} & 157.04 & 27.32 & \underline{1.27}  & \underline{144.88} & 28.68                  & \underline{1.46}          \\
                        &                      & Adaptformer & 168.54 & 31.25 & 1.38 & 137.61 & 28.99 & 0.87 & 155.14 & 28.12 & 0.94  & 137.73 & 27.56 & 1.19 & 159.13 & \underline{27.38}  & 1.24  & 151.63  & 28.66           & 1.10          \\
                        &                      & LT-SFT    & 178.51  & 31.44  & 0.88 & 131.72  & \underline{29.01}  & 1.34 & 149.82 & 28.01 & 1.03 & 140.51 & \textbf{27.91}  & 1.30 & \underline{154.82} & 27.16  & 1.21   & 151.08   & 28.71         & 1.16          \\
                        &                      & \textbf{SaRA (Ours)}       & \textbf{162.38} & \textbf{31.61}  & \textbf{1.84} & \underline{128.55} & \textbf{29.21} & \underline{1.72} & \underline{142.60} & \underline{28.22}  & \textbf{1.35} & \underline{135.44}& \underline{27.72} & \underline{1.39}  & \textbf{153.33} & \textbf{27.44} & \textbf{1.58} & \textbf{144.46} & \textbf{28.84} & \textbf{1.58} \\ \cline{2-21}
                        & \multirow{5}{*}{5M}  & DoRA       & \textbf{166.21} & 31.19  & \textbf{1.49} & \textbf{124.68} & 29.09  & \underline{1.81} & 174.47  & 28.05 & 0.66      & 139.24 & 27.37 & 1.00  & 165.32 & \underline{27.26} &
                        0.83 & 153.98 & \underline{28.59} & 0.94 \\ &
                                               & Lora      & \underline{169.38}  & 30.97  & \underline{1.25} & \underline{126.76} & 29.01 & 1.73 & \underline{151.41}     & 27.80  & \underline{0.86} & \underline{138.03} & 27.41  & \underline{1.07}  & \textbf{157.21} & 27.01  & \underline{0.94} & \underline{148.56}  & 28.44               & \underline{1.13}          \\
                        &                      & Adaptformer & 178.61 & 30.88 & 0.71 & 138.76 & \underline{29.21} & 0.92 & 160.38 & 27.99 & 0.72 & 144.51  & 27.63  & 0.94 & 161.77 & 26.96  & 0.68  & 156.81  &28.53                & 0.76          \\
                        &                      & LT-SFT     & 174.77  & \underline{31.65} & 1.16 & 129.10 & 29.15 & 1.64 & 165.69 & \underline{28.08} & 0.62 & 147.41 & \underline{27.58}   & 0.78 & 165.84 & 27.11  & 0.65   & 156.56   & 28.71             & 0.88          \\
                        &                      & \textbf{SaRA (Ours)}     & 174.95   & \textbf{31.84} & 1.20 & 127.01 & \textbf{29.33} & \textbf{1.92} & \textbf{144.27}           & \textbf{28.40}       & \textbf{1.41} & \textbf{137.07} & \textbf{27.66}  & \textbf{1.28} & \underline{158.02} & \textbf{27.45} & \textbf{1.36} &  \textbf{148.26} & \textbf{28.94}  & \textbf{1.44} \\ \cline{2-21}
                        & Full-finetune        & 2085M      & 160.72 & 28.55  & 1.00 & 128.94  & 27.81 & 0.77 & 144.56 & 27.01 & 0.59 & 124.59 & 26.41 & 1.00 & 146.60 & 26.48     
                        & 0.89  & 141.08    & 27.25           & 0.81   \\
                        
                                \bottomrule
\end{tabular}}
\vspace{-0.1in}
\caption{\reb{Comparison with different parameter-efficient fine-tuning methods on Stable Diffusion XL. For most of the conditions, our model achieves the best FID and VLHI score, indicating that our model learns domain-specific knowledge successfully while keeping the prior information well.}}

\label{tab:quantitative comparison on SD xl.}
\end{table}

\begin{table}[t]
\renewcommand{\arraystretch}{1.3}
\resizebox{1.0\linewidth}{!}{
\begin{tabular}{c|c|c|ccc|ccc|ccc|ccc|ccc|ccc}
\toprule
\multirow{2}{*}{Backbone} & \multirow{2}{*}{Params} & \multirow{2}{*}{Model} & \multicolumn{3}{c}{BarbieCore} & \multicolumn{3}{c}{Cyberpunk} & \multicolumn{3}{c}{ElementFire} & \multicolumn{3}{c}{Expedition} & \multicolumn{3}{c}{Hornify} & \multicolumn{3}{c}{Mean} \\
                          &                        &                         & FID $\downarrow$         & B-VQA $\uparrow$     & VLHI $\uparrow$     & FID $\downarrow$         & B-VQA $\uparrow$    & VLHI $\uparrow$   & FID $\downarrow$         & B-VQA $\uparrow$   & VLHI $\uparrow$    & FID $\downarrow$         & B-VQA $\uparrow$  & VLHI $\uparrow$    & FID $\downarrow$         & B-VQA $\uparrow$   & VLHI $\uparrow$   & FID $\downarrow$         & B-VQA $\uparrow$  & VLHI $\uparrow$  \\ \midrule
\multirow{16}{*}{SD 1.5}  & \multirow{5}{*}{50M}   & DoRA                    & 158.40     & 0.37    & 1.16    & \underline{119.06}       & 0.42   & 0.58  & 171.96        & 0.48   & 0.96   & \textbf{131.33}       & 0.52   & \underline{1.39}   & \underline{150.33}      & 0.49  & 1.31  & \underline{146.22}     & 0.46 & 1.20 \\
                          &                        & Lora                    & 161.88     & \textbf{0.40}    & 1.26    & \textbf{117.49}       & \textbf{0.47}   & \textbf{1.35}  & 181.66        & \underline{0.51}   & 0.89   & 136.31       & 0.52   & 1.23   & 156.36      & 0.49  & 1.15  & 150.74     & 0.48 & 1.22 \\
                          &                        & Adaptformer             & 166.09     & 0.38    & 0.82    & 126.21       & 0.46   & 0.56  & \underline{151.27}        & \textbf{0.68}   & \textbf{1.73}   & 138.01       & 0.53   & 1.22   & 151.53      & 0.50  & \underline{1.37}  & 146.62     & \textbf{0.51} & \underline{1.60} \\
                          &                        & LT-SFT                  & \underline{157.80}     & 0.38    & \underline{1.27}    & 123.59       & 0.46   & 0.81  & 171.67        & 0.46   & 0.93   & 139.29       & \textbf{0.54}   & 1.35   & 158.52      & \textbf{0.51}  & 1.25  & 150.18     & 0.47 & 1.19 \\
                          &                        & \textbf{SaRA (Ours)}                    & \textbf{148.54}     & \textbf{0.40}    & \textbf{1.84}    & 121.67       & \textbf{0.47}   & \underline{1.05}  & \textbf{132.67}        & 0.50   & \underline{1.58}   & \underline{132.54}       & \underline{0.53}   & \textbf{1.48}   & \textbf{140.36}      & \textbf{0.51}  & \textbf{1.73}  & \textbf{135.15}     & \underline{0.48} & \textbf{1.75} \\ \cline{2-21}
                          & \multirow{5}{*}{20M}   & DoRA                    & 158.85     & 0.37    & 1.14    & \textbf{116.23}       & 0.44   & 0.98  & \underline{169.91}        & 0.48   & \underline{0.99}   & \underline{133.80}       & 0.52   & \underline{1.29}   & \underline{148.97}      & 0.49  & \underline{1.36}  & \underline{145.55}     & 0.46 & 1.25 \\
                          &                        & Lora                    & 159.64     & \textbf{0.40}    & \textbf{1.36}    & 117.21       & 0.47   & \underline{1.35}  & 174.79        & \textbf{0.50}   & 0.97   & 136.38       & 0.52   & 1.22   & 155.85      & 0.49  & 1.23  & 148.77     & 0.48 & \underline{1.29} \\
                          &                        & Adaptformer             & 159.02     & 0.38    & 1.22    & 123.88       & 0.46   & 0.72  & 174.17        & 0.49   & 0.96   & 137.03       & 0.51   & 0.98   & 157.09      & 0.48  & 1.09  & 150.24     & 0.46 & 1.13 \\
                          &                        & LT-SFT                  & \underline{156.60}     & 0.38    & 1.30    & 119.75       & \textbf{0.48}   & 1.40  & 191.01        & 0.50   & 0.73   & 144.57       & \textbf{0.54}   & 1.19   & 165.47      & \textbf{0.51}  & 1.08  & 155.48     & 0.48 & 1.12 \\
                          &                        & \textbf{SaRA (Ours)}                    & \textbf{153.68}     & \textbf{0.40}    & \textbf{1.64}    & \underline{116.69}       & \underline{0.47}   & \textbf{1.50}  & \textbf{138.64}        & \textbf{0.50}   & \textbf{1.49}   & \textbf{129.98}       & \textbf{0.54}   & \textbf{1.75}   & \textbf{145.62}      & \underline{0.50}  & \textbf{1.53}  & \textbf{136.92}     & \textbf{0.48} & \textbf{1.72} \\ \cline{2-21}
                          & \multirow{5}{*}{5M}    & DoRA                    & \underline{156.21}     & 0.37    & 1.21    & \textbf{113.26}       & 0.44   & \underline{1.29}  & \underline{178.70}        & 0.47   & 0.84   & 135.59       & 0.52   & \underline{1.24}   & 161.21      & 0.48  & 1.01  & \underline{148.99}     & 0.46 & 1.12 \\
                          &                        & Lora                    & 163.80     & \textbf{0.41}    & \underline{1.25}    & \underline{117.58}       & 0.46   & 1.27  & 184.99        & 0.50   & 0.83   & 137.96       & 0.52   & 1.17   & \textbf{153.57}      & 0.49  & \textbf{1.22}  & 151.58     & 0.48 & \underline{1.20} \\
                          &                        & Adaptformer             & 164.22     & 0.34    & 0.64    & 120.98       & 0.46   & 1.03  & 184.84        & 0.49   & 0.82   & 143.01       & \underline{0.53}   & 1.11   & 171.34      & 0.49  & 0.79  & 156.88     & 0.46 & 0.94 \\
                          &                        & LT-SFT                  & 169.24     & 0.40    & 0.91    & 127.01       & \textbf{0.49}   & 1.00  & 202.47        & \textbf{0.52}   & 0.61   & 153.49       & \textbf{0.56}   & 1.00   & 176.41      & \textbf{0.53}  & 1.00  & 165.72     & \textbf{0.50} & 0.94 \\
                          &                        & \textbf{SaRA (Ours)}                    & \textbf{153.69}     & \textbf{0.41}    & \textbf{1.70}    & 118.74       & \underline{0.47}   & \textbf{1.34}  & \textbf{174.86}        & \underline{0.51}   & 1.00   & 134.45       & 0.52   & \textbf{1.30}   & \underline{157.24}      & \underline{0.50}  & \underline{1.21}  & \textbf{147.80}     & \underline{0.48} & \textbf{1.36} \\ \cline{2-21}
                          & Full-finetune          & 860M                    & 147.81     & 0.30    & 1.00    & 120.22       & 0.47   & 1.15  & 136.49        & 0.26   & 0.95   & 129.07       & 0.48   & 1.00   & 134.86      & 0.40  & 1.00  & 133.69     & 0.38 & 1.00 \\
                           \bottomrule

\end{tabular}
}
\vspace{-0.1in}
\caption{\reb{Comparison on FID and B-VQA from T2i-compbench++~\citep{huang2023t2ibench} with different parameter-efficient fine-tuning methods on Stable Diffusion 1.5. }}
\label{tab:T2Ibench on sd1.5}
\end{table}



\begin{table}[t]
\renewcommand{\arraystretch}{1.1}
\resizebox{1.0\linewidth}{!}{
\begin{tabular}{l|ccc|ccc|ccc|ccc}
\toprule
\multirow{2}{*}{ \pzo \pzo \pzo \pzo\pzo\pzo\pzo Methods} & \multicolumn{3}{c|}{Dog} & \multicolumn{3}{c|}{Clock} & \multicolumn{3}{c|}{Backpack} & \multicolumn{3}{c}{Mean} \\
 & CLIP-I \yr{$\uparrow$} & CLIP-T\yr{$\uparrow$} & VLHI \yr{$\uparrow$} & CLIP-I \yr{$\uparrow$} & CLIP-T \yr{$\uparrow$}& VLHI \yr{$\uparrow$} & CLIP-I \yr{$\uparrow$} & CLIP-T \yr{$\uparrow$}& VLHI \yr{$\uparrow$} & CLIP-I \yr{$\uparrow$} & CLIP-T \yr{$\uparrow$}& VLHI \yr{$\uparrow$} \\ \midrule
 Textual Inversion & 0.788 & 23.94 & 0.36 & 0.789 & \textbf{24.15} & 1.00 & 0.654 & 24.09 & 0.00 & 0.744 & 24.06 & 0.39 \\
Dreambooth + Full Fine-tune & 0.776 & \underline{25.85} & \underline{1.13} & \underline{0.894} & 22.39 & 1.13 & 0.856 & \textbf{25.44} & \underline{1.76} & 0.842 & 24.56 & 1.36 \\
Dreambooth + LoRA & \textbf{0.895} & 23.64 & 1.00 & \textbf{0.913} & 21.71 & 1.00 & \textbf{0.917} & 25.23 & \textbf{1.84} & \textbf{0.908} & 23.53 & 1.00 \\
Dreambooth + Adaptformer & 0.772 & 25.42 & 0.91 & 0.885 & 23.18 & \underline{1.38} & 0.873 & 25.25 & 1.69 & 0.843 & \underline{24.62} & \underline{1.41} \\
Dreambooth + LT-DFT & 0.757 & 23.94 & 0.13 & 0.893 & 22.45 & 1.14 & 0.869 & 25.00 & 1.49 & 0.840 & 23.80 & 0.79 \\
Dreambooth + SaRA (Ours)  &   \underline{0.790} & \textbf{25.87} & \textbf{1.24} & 0.887 & \underline{23.51} & \textbf{1.53} & \underline{0.886} & \underline{25.27} & \underline{1.76} & \underline{0.854} & \textbf{24.88} & \textbf{1.67} \\
\bottomrule
\end{tabular}}
\vspace{-0.15in}
\caption{Quantitative comparisons between different PEFT methods on image customization.}
\label{tab:comparison on dreambooth}
\vspace{-0.15in}
\end{table}

\section{More comparison results on image customization}
\label{sec: more comparison results on image customization}
\begin{figure}[t]
\centering
\includegraphics[width=1.0\textwidth]{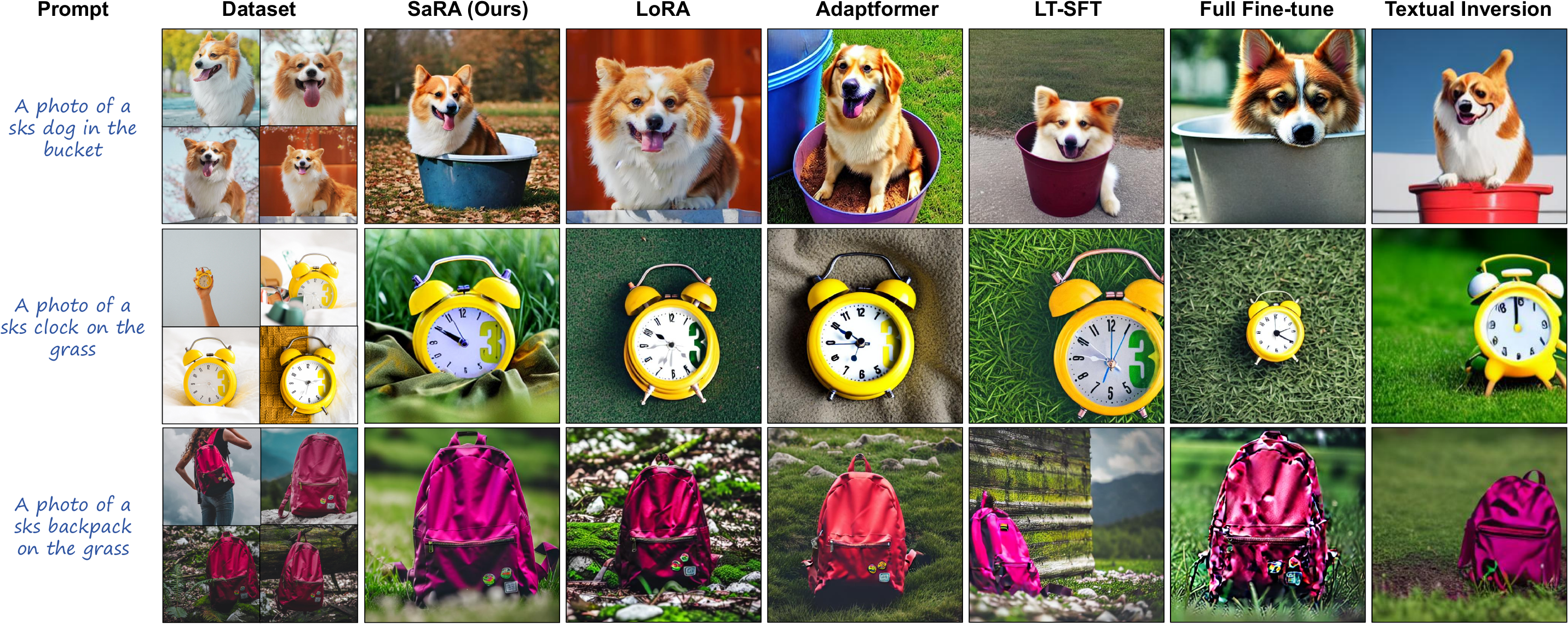}
\caption{Qualitative comparisons among different PEFT methods on image customization by fine-tuning the UNet model in Dreambooth~\citep{dreambooth}. Our model can accurately capture the target feature while preventing the model from overfitting, outperforming Dreambooth with other PEFT methods and Textual inversion~\citep{textualinversion}.
}
\label{fig:comparison on dreambooth}

\end{figure}

{\bf Image Customization.} Image customization aims to learn a common subject from a few images and then apply it to new images.
Dreambooth~\citep{dreambooth} trains the UNet of a diffusion model to bind the target subject to a rare token and then generates images with the specified content based on the rare token. 
Since Dreambooth requires fine-tuning the UNet network, we compare the performance of full-finetune (original Dreambooth), LoRA, Adaptformer, LT-SDT, and our method in image customization. 
We compute the CLIP-Text score and CLIP-IMG Score for the generated data, along with VLHI balancing both the two metrics.
As shown in Tab.~\ref{tab:comparison on dreambooth}, \hut{ LoRA achieves a high CLIP-IMG score but the lowest CLIP-Text score, indicating a severe overfitting problem. Other PEFT methods, including full-parameter fine-tuning, achieve relatively low CLIP-IMG and CLIP-Text scores. In contrast,} our method achieves the best CLIP-Text score, a competitive CLIP-IMG score (only lower than the overfitted LoRA), and the best average VLHI score across three datasets, demonstrating its effectiveness in image customization tasks. We further conduct the qualitative comparison on fine-tuning Dreambooth. As shown in Fig.~\ref{fig:comparison on dreambooth}, our method can learn the subject content well while preserving the prior information of the diffusion model, thereby improving the consistency between the generated images and the given texts, which demonstrates the effectiveness of SaRA in image customization.

\begin{figure}[t]
\centering
\includegraphics[width=1.0\textwidth]{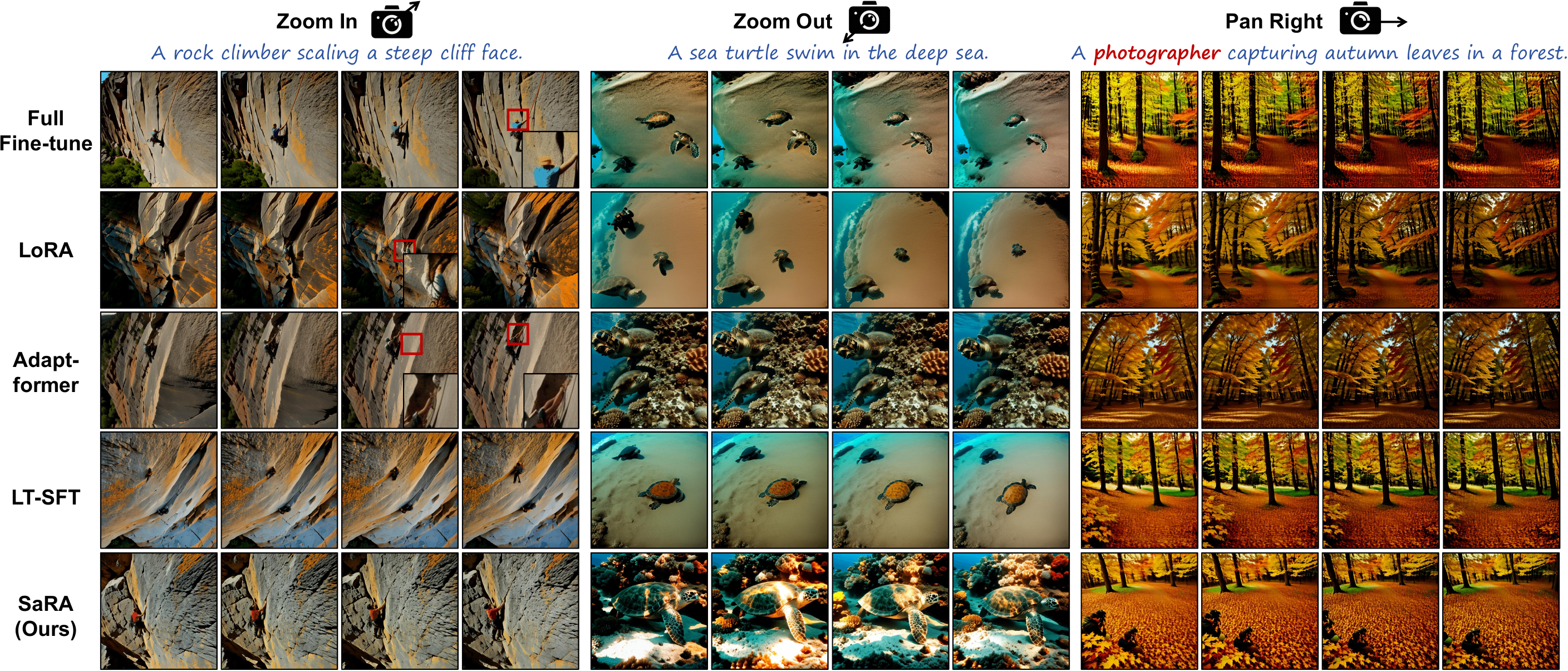}
\caption{\hut{The comparison results of the video generation model~\citep{animatediff}, fine-tuned using different PEFT methods on three video datasets featuring zoom-in, zoom-out, and pan-\yr{right} camera motions. 
The red box\yr{es} 
highlight artifacts generated by the compared methods, indicating that these methods \yr{have lost} some model prior\yr{s} of specific content during \yr{the} fine-tuning \yr{process}. 
In the sea turtle examples, full fine-tuning, LoRA, and LT-SFT exhibit noticeable content degradation. 
And for the pan-right examples, all the compared methods fail to capture the photographer, indicating significant model overfitting. 
\yr{In contrast, our method achieves excellent camera motion control while preserving video content well.}}
}
\label{fig:comparison_on_video_generation}
\end{figure}

\section{Controllable \yr{V}ideo \yr{G}eneration.} 
\label{sec: controllable video generation}

We further investigate the effectiveness of our method in fine-tuning video generation models. 
AnimateDiff~\citep{animatediff} \yr{is a representative video generation model} based on Stable Diffusion~\citep{stablediffusion}, \yr{which} inserts temporal attention modules between the original spatial attention modules to model temporal correlations, enabling a diverse text-to-video generation. 
To achieve more controllable generation, AnimateDiff fine-tunes the temporal attention module using different camera motion data, such as Pan Left, \yr{Pan} Right, Zoom In, \yr{and Zoom} Out, to control the camera movements precisely. 
We compare the effectiveness of various PEFT methods in fine-tuning AnimateDiff for \hut{three} types of camera movements\hut{, including Zoom In, Zoom Out, and Pan Right}. Specifically, we collected 1,000 video-text pairs with identical camera movements for each type of camera motion. The temporal attention modules are fine-tuned using full fine-tuning, LoRA, Adaptformer, LT-SDT, and our SaRA. 
\hut{As shown in Fig.\yr{~\ref{fig:comparison_on_video_generation}}, the compared methods usually suffer from generating artifacts in the results (shown in red boxes), indicating that these
methods have lost some model priors of specific content during the fine-tuning process. Moreover, for the sea turtle examples, full fine-tuning, LoRA, and LT-SFT exhibit noticeable content degradation. And for the pan-right examples, all the compared methods fail to capture the photographer, indicating a significant model overfitting problem. In contrast,
our method achieves excellent camera motion control while \yr{achieving good} consistency between the \yr{video} content and the text. }

\section{Scaling Weight for SaRA Parameters}
\label{sec: Weights for SaRA Parameters}

\reb{Our SaRA aims to learn a sparse low-rank parameter matrix $\Delta P$, which is added to the pre-trained weights $P_0$. Similar to LoRA~\citep{hu2021lora}, when applying the learned parameter $\Delta P$ to the pre-trained one, we can assign a scaling weight $\alpha$ for the $\Delta P$ to control the emphasis extent on the learned target-domain knowledge by:}

\reb{
\begin{equation}
    P=P+\alpha \Delta P.
\end{equation}
}

\begin{figure}[t]
\centering
\includegraphics[width=1.0\textwidth]{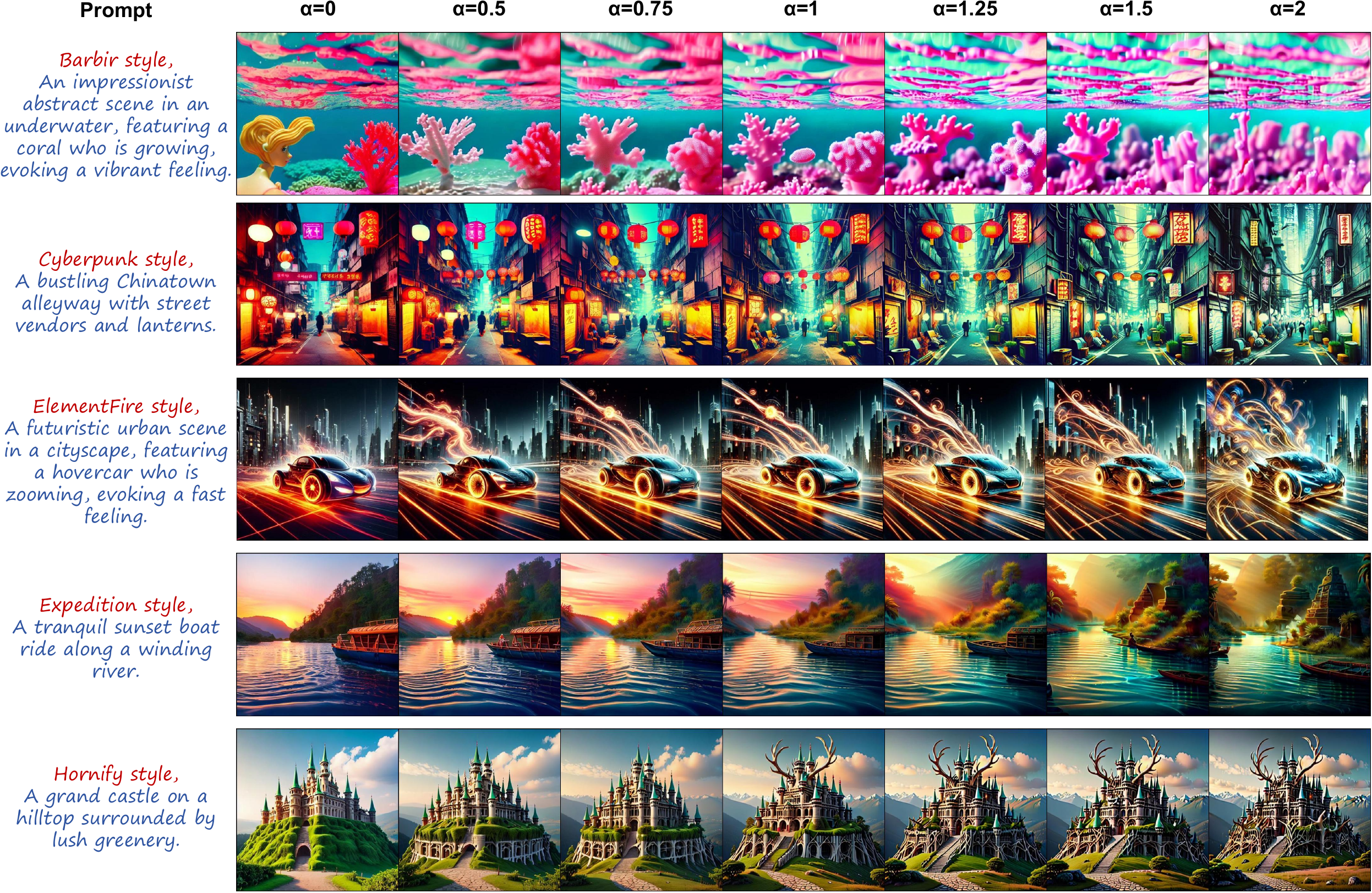}
\caption{\reb{The generated results from different weight $\alpha$ for the learned SaRA parameters by Stable Diffusion 1.5. As $\alpha$ increases, the generated image contains more tar-get domain features.}}
\label{fig:sara weight}
\end{figure}

\reb{We show the results on different $\alpha$ ranging from $0$ to $2$ on five datasets in Fig.~\ref{fig:sara weight}. It can be seen that as the scaling weight $\alpha$ increases, the model tends to generate images with more target-domain features, but may lose part of the information specified by the given texts.}

\section{Merging Different SaRA Parameters}
\label{sec:merge sara params}
\reb{For two SaRA parameters $\Delta P_1$ and $\Delta P_2$ learned from two different datasets, we aim to find whether they can be combined to form new parameters that contain the knowledge from both the two datasets. We combine the two parameters by:}

\reb{
\begin{equation}
    \Delta P=\alpha_1 \Delta P_1+\alpha_2\Delta P_2.
\end{equation}
}

\begin{figure}[t]
\centering
\includegraphics[width=1.0\textwidth]{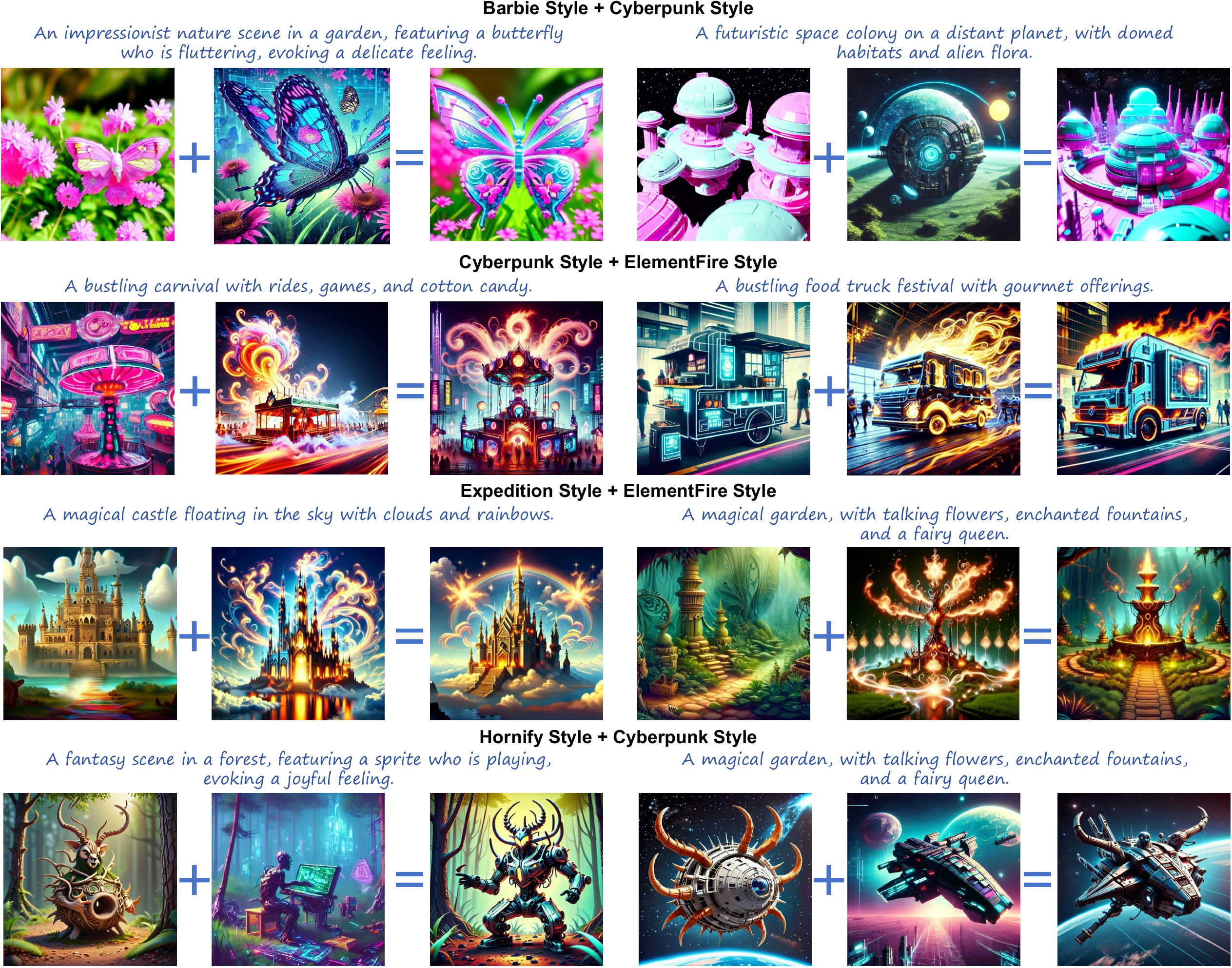}
\caption{\reb{Combining the SaRA parameters learned from two different datasets, the model can generate images with features from both datasets. We show the combination results for 'Barbie Style' +'Cyberpunk style', 'Cyberpunk Style' +'ElementFire style', 'ElementFire Style' +'Expedition style', and 'Hornify Style' +'Cyberpunk style' in this figure.}}
\label{fig:sara combination}
\end{figure}

\begin{figure}[h]
\centering
\includegraphics[width=1.0\textwidth]{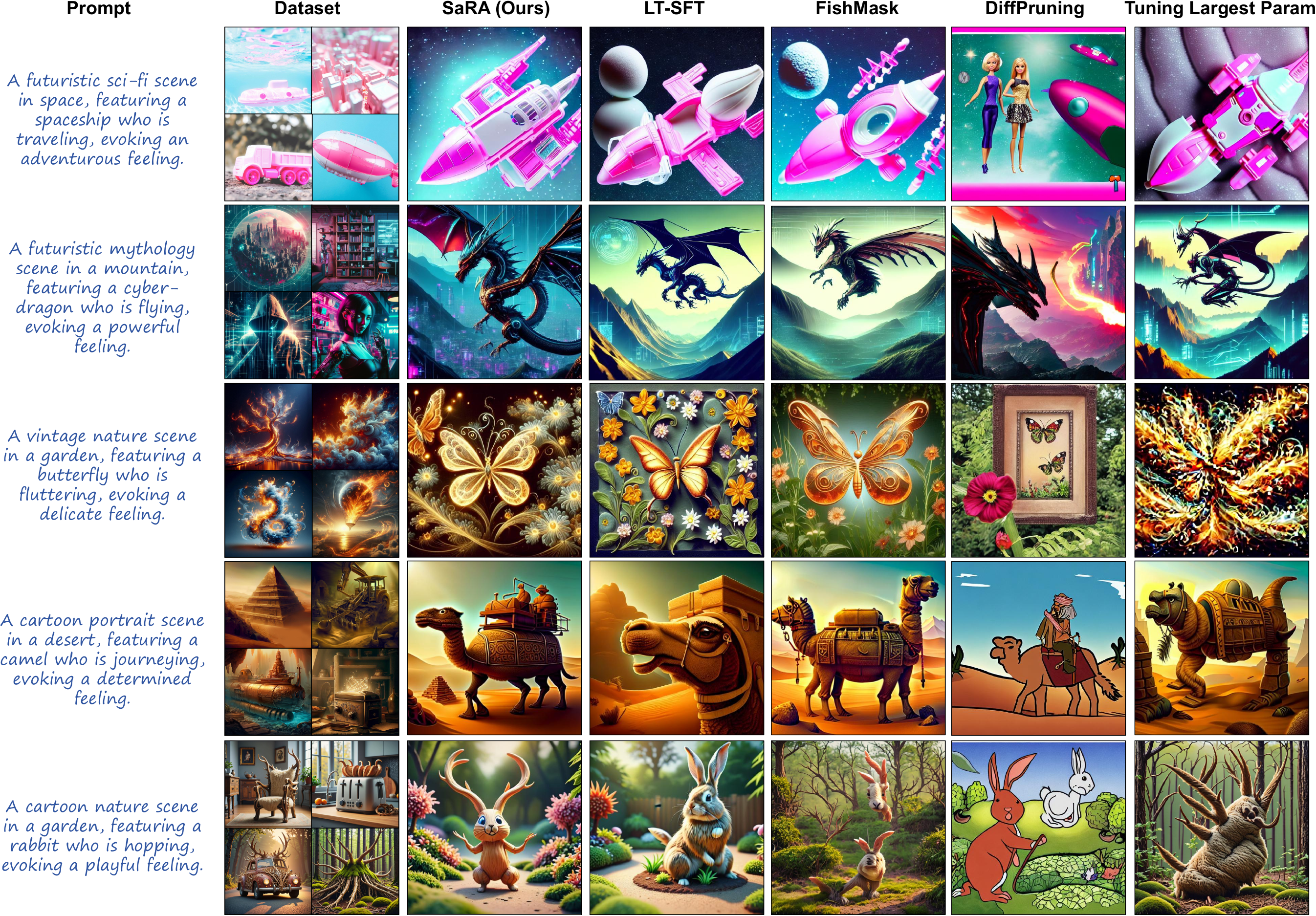}
\vspace{-0.2in}
\caption{\reb{More qualitative comparison with the existing Selective PEFT methods (LT-SFT~\citep{ansell2021Lt-sft}, FishMask~\citep{sung2021fishmask}, DiffPruning~\citep{guo2020diffpruninig}), and the ablated model that fine-tunes the largest parameters) on SD 1.5.}}
\label{fig:comparison with more sft}
\vspace{-0.1in}
\end{figure}

\reb{Then, we employ the combined SaRA parameter $\Delta P$ to generate images. We choose four combinations: 'Barbie Style' +'Cyberpunk style', 'Cyberpunk Style' +'ElementFire style', 'ElementFire Style' +'Expedition style', and 'Hornify Style' +'Cyberpunk style', where we simply assign $\alpha_1=\alpha_2=0.6$. The generated images are shown in Fig.~\ref{fig:sara combination}. It can be seen that, after combining the SaRA parameters learned from two different datasets, the output images contain the features from both two datasets, which indicates that we can merge different SaRA parameters together, enabling more flexible and abundant generation results.}

\section{\reb{More ablation studies.}}
\label{sec: more ablation studies}

\reb{{\bf Ablation study on threshold.}
In SaRA, the threshold is an important hyperparameters that influence the size of the parameter space directly. We conduct experiments on different thresholds ranging from 2e-4 to 1e-1 on the Expedition dataset and Stable Diffusion 1.5, and evaluate the generated images by FID and CLIP score. The results are shown in Tab.~\ref{tab: ablation on threshold}. It can be seen that when the threshold is too small (e.g., 2e-4), the FID becomes much higher, indicating learning less target domain knowledge. And when the threshold is large (i.e., threshold$\le$2e-3), the model performs quite stably. Since we have a low-rank loss, the model with a high threshold also keeps the CLIP score well. In summary, our SaRA performs well in different thresholds, demonstrating the robustness of our model.
}

\begin{table}[h]
\centering
\begin{tabular}{c|ccccccc}
\toprule
Threshold & 2e-4   & 8e-4   & 2e-3   & 5e-3   & 1e-2   & 5e-2   & 1e-1   \\
\midrule
FID $\downarrow$      & 134.45 & 129.98 & 132.54 & 131.05 & 130.42 & 130.71 & 129.88 \\
CLIP $\uparrow$     & 27.06  & 27.04  & 27.38  & 27.21  & 27.15  & 27.04  & 27.02 \\
\bottomrule
\end{tabular}
\caption{\reb{Ablation study on the threshold.}}
\label{tab: ablation on threshold}
\end{table}

\reb{{\bf More ablation study on the low-rank loss.} In this section, we conduct more ablation studies on the low-rank loss $\mathcal{L}_{rank}$, where we choose Stable Diffusion 1.5 and two additional datasets (Cyberpunk and ElementFire) for the experiment. The results are shown in Tab.~\ref{tab:more ablation on low-rank loss}, where we can see that the mode without $\mathcal{L}_{rank}$ always tends to get a worse CLIP score, indicating a significant performance drop. Therefore, the low-rank loss is quite necessary in our model to keep the model prior.
}

\begin{table}[h]
\centering
\begin{tabular}{c|cc|cc}
\toprule
\multirow{2}{*}{method}         & \multicolumn{2}{c}{Cyberpunk} & \multicolumn{2}{c}{ElementFire} \\ 
         & FID       & CLIP        & FID    & CLIP  \\ \midrule
SaRA                & 121.67    & 27.30       & 132.67 & 26.77 \\
SaRA w/o. $\mathcal{L}_{rank}$ & 120.33    & 26.52       & 131.56 & 25.88\\
\bottomrule
\end{tabular}
\caption{\reb{More ablation studies on the low-rank loss $\mathcal{L}_{rank}$.}}
\label{tab:more ablation on low-rank loss}
\end{table}

\section{Analysis on Training efficiency}
\label{sec: Analysis on Training efficiency}

In Sec.4.4 of the main paper, we propose unstructural backpropagation, which allows selective PEFT to store and update only the gradients of trainable parameters, significantly reducing memory usage during training. We conducted experiments on the Stable Diffusion 2.0 model using an 80G NVIDIA A100 GPU, comparing the memory usage and training time of LT-SFT (Selective PEFT method), LoRA, and our method across different batch sizes. The results, shown in Fig. 5 of the main paper, demonstrate that our method achieves the lowest memory consumption and training time under all batch sizes. Compared to LT-SFT, we reduce memory usage by a fixed 9.2G (equivalent to the total gradient size of fixed parameters) and achieve over 45\% memory reduction for smaller batch sizes. Furthermore, compared to LoRA, our method saves over 52\% memory and 49\% training time for larger batch sizes, showcasing the efficiency of our SaRA in model fine-tuning.

\section{\yr{Further Analysis to} Understand what SaRA \yr{have} 
 learn\yr{ed}}
\label{sec: further analysis to understand what we have learned}

\textbf{The Correlation between $\Delta P$ and $P$.}
We further investigate what exactly is learned by the sparse parameter matrix $\Delta P_{Ours}$ obtained through our method. 
First\yr{ly}, we examine the relationship between $\Delta P$ and the pre-trained parameter matrix $P$. 
We \yr{want to know} whether $\Delta P$ \yr{has} learn\yr{ed} new knowledge \yr{that is} not present in $P$, or \yr{it amplifies some existing but} previously 
\yr{not emphasized}
knowledge in $P$. 
To answer this question, we study the subspaces of $\Delta P$ and $P$. 
We \yr{first conduct SVD decomposition on $\Delta P\yr{_{Ours}}$, and obtain the left and right singular-vector matrices $U_{\Delta P_{Ours}}$ and $V_{\Delta P_{Ours}}^T$. 
We then}
project $P$ into the first $r$-dimensional subspace of $\Delta P$ using 
\yr{$U_{\Delta P_{Ours}}PV_{\Delta P_{Ours}}^T$}.
We quantify the correlation between $P$ and the first $r$-dimensional subspace of $\Delta P$ by calculating the Frobenius norm of 
\yr{this projection $\|U_{\Delta P_{Ours}}PV_{\Delta P_{Ours}}^T\|_F$,
where a smaller norm indicates lower correlation between the subspace of $\Delta P$ and $P$.}

For a valid reference, we further decompose the parameter matrix $\Delta P_{LoRA}$ learned by LoRA using SVD to obtain \yr{the} respective $U\yr{_{\Delta P_{LoRA}}}$ and $V\yr{_{\Delta P_{LoRA}}}^T$ matrices, 
and \yr{project} the pre-trained parameter matrix $P$ \yr{into the first $r$-dimensional subspace of $\Delta P_{LoRA}$ using $U_{\Delta P_{LoRA}}P{V_{\Delta P_{LoRA}}}^T$.} 

\yr{In addition}, we calculate \yr{an amplification factor}
to determine how much the parameter matrix $\Delta P$ amplifies the directions \yr{that are} not emphasized by $P$. 
The amplification factor is computed as 
\yr{$f_{a} = \frac{\|\Delta P\|_F}{\|UPV^T\|_F}$}.
The higher the amplification factor is, the more task-specific knowledge is learned.

We investigate the relationship between the first $r=4, 16, 64$ dimensional subspaces of $\Delta P$ and $P$. The results are shown in Tab.~\ref{tab:relationship between P and delta P}\footnote{$\|\Delta P_{ours}\|_F=4.40$ and $\|\Delta P_{LoRA}\|_F=4.62$.}, \yr{from which} we \yr{can} draw the following conclusions:

\textit{1. The learned sparse matrix $\Delta P_{Ours}$ from our model \yr{has} a significant amplification factor, such as \textbf{25.72} times for $r=4$\yr{, which indicates the correlation between the first 4-dimensional subspace of $\Delta P_{Ours}$ and $P$ is low, and $\Delta P_{Ours}$} primarily amplifies the directions \yr{that are} not emphasized in $P$.}

\textit{2. Compared to the low-rank parameter matrix $\Delta P_{LoRA}$ learned by LoRA, our model achieves a higher amplification factor across different values of $r$, indicating that our method can learn more knowledge \yr{that is} not emphasized in $P$ than LoRA.}

\textit{3. As $r$ increases, the amplification factor gradually decreases, suggesting that the knowledge learned by $\Delta P$ is mostly contained within $P$, and the primary role of $\Delta P$ is to amplify some of the existing \yr{but previously not emphasized} knowledge in $P$ 
\yr{(if new knowledge that is not present in $P$ has been learned by $\Delta P$, the correlation should remain low as $r$ increases, and the amplification factor should remain high, which is not the case presented in our experiments)}.}



\begin{table}[t]
\renewcommand{\arraystretch}{1.2}
\resizebox{1.0\linewidth}{!}{
\begin{tabular}{c|ccc|ccc|ccc}
\toprule
Rank          & \multicolumn{3}{c|}{r=4} & \multicolumn{3}{c|}{r=16} & \multicolumn{3}{c}{r=64} \\
Matrices              & $\Delta P_{Ours}$   & $\Delta P_{LoRA}$   & $P$    & $\Delta P_{Ours}$  & $\Delta P_{LoRA}$   & $P$      & $\Delta P_{Ours}$  & $\Delta P_{LoRA}$    & $P$     \\ \midrule
$\|U P V^T\|_F$              & 0.17  & 0.34 & 9.36 & 0.48 & 1.14 & 14.82 & 2.68  & 3.90  & 23.91 \\
Amplification & \textbf{25.72}   & 13.45  & -    & \textbf{6.50}    & 4.05   & -      & \textbf{1.64}   & 1.18    & -    \\ \bottomrule
\end{tabular}}
\caption{The correlation between the learned parameter matrices $\Delta P$ and the pretrained weights $P$. Our learned parameter matrix $\Delta P_{Ours}$ \yr{amplifies} the directions that are not emphasized in the pretrained weights $P$, and \yr{has} a larger amplification factor than LoRA, indicating our model learns more task-specific knowledge than LoRA.}
\label{tab:relationship between P and delta P}
\end{table}

\begin{figure}[t]
\centering
\includegraphics[width=1.0\textwidth]{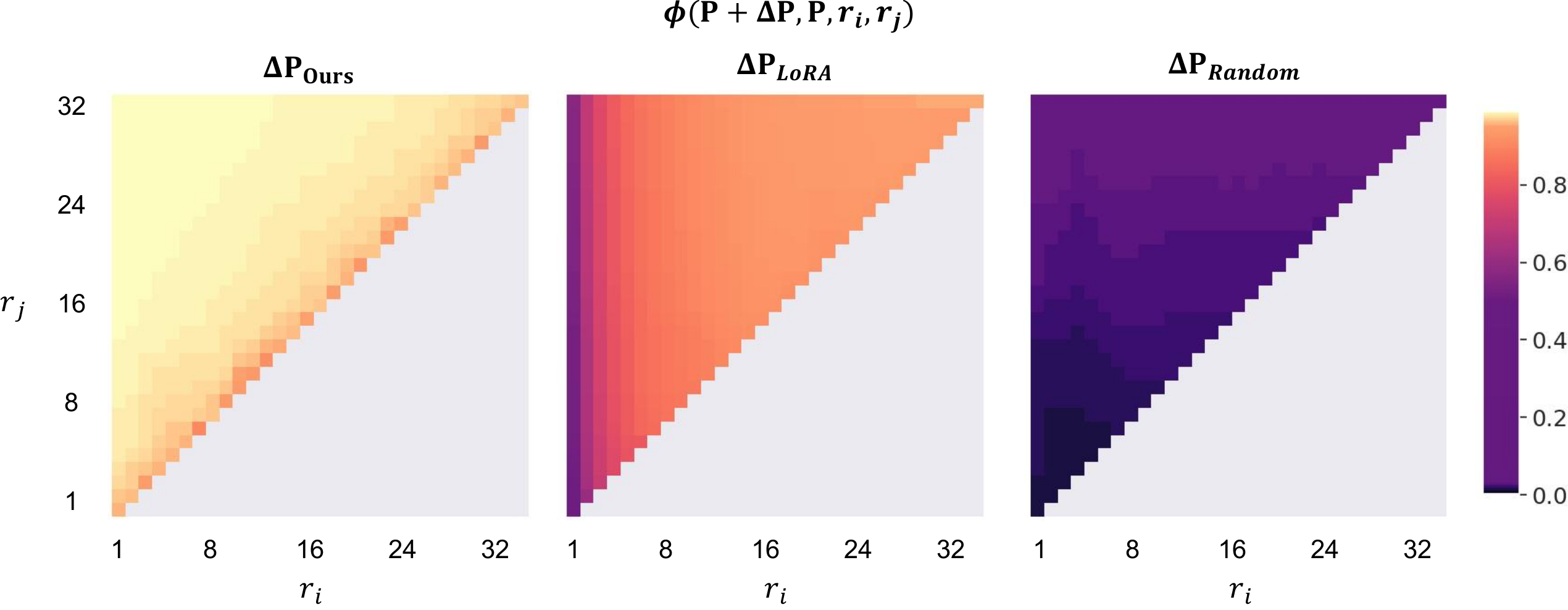}
\vspace{-0.15in}
\caption{Subspace similarity between $P+\Delta P$ and $P$. 
\yr{The similarity $\phi(P+\Delta P_{Ours},P,r_i,r_j)$} between our updated parameter matrix and the pretrained parameter matrix achieves a similarity larger than \textbf{96\%} \yr{across} different dimensions of the subspace \yr{$(r_i,r_j)$}.
\yr{The results indicare} the learned sparse parameter matrix $\Delta P$ of our model keeps the prior information in the pretrained parameters $P$ well.}
\label{fig:subspace sim between delta P+P and P}

\end{figure}

\begin{figure}[t]
\centering
\includegraphics[width=1.0\textwidth]{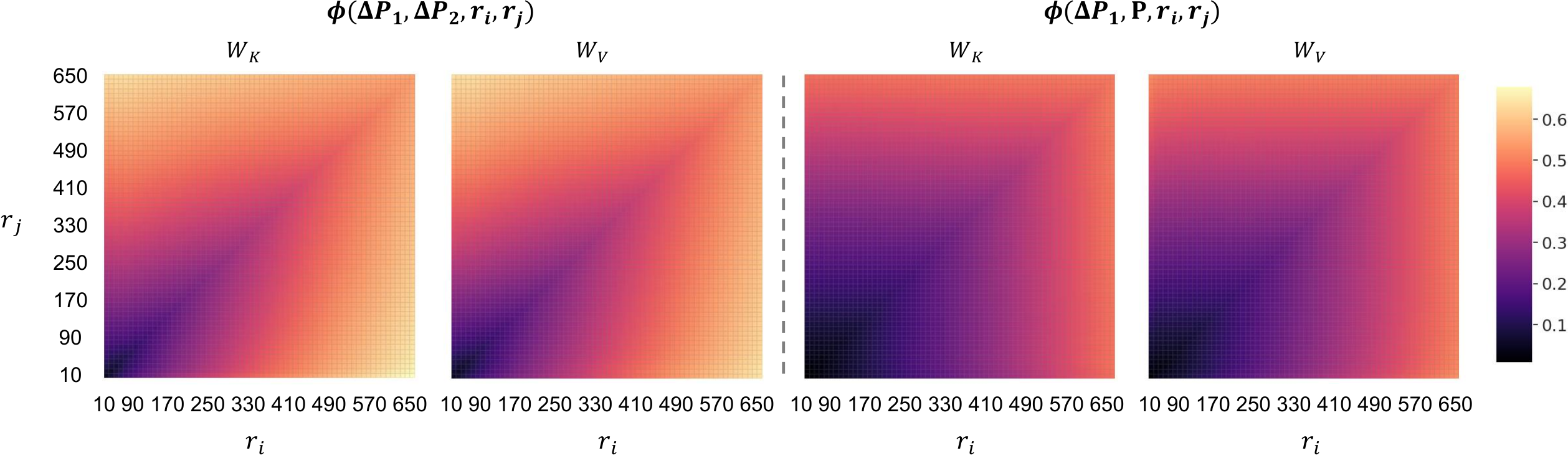}
\vspace{-0.15in}
\caption{Subspace similarity between $\Delta P_1$ and $\Delta P_2$ \yr{under} thresholds $2e-3$ and $8e-4$. The total similarity between their learned matrices exceeds $60\%$ (when $r$ is around 650), compared to only $40\%$ similarity between $\Delta P_1$ and the pre-trained weights $P$, demonstrating the learned matrices from different threshold\yr{s} learn similar task-\yr{specific} knowledge, while emphasizing different directions (relative smaller similarity when $r$ is small).}
\label{fig:subspace sim between delta p}
\end{figure}

\textbf{The Correlation between $P+\Delta P$ and $P$.}
We aim to further understand whether our learned parameter matrix $\Delta P_{Ours}$ disrupts the information in the original parameter space \yr{(spanned} by the pretrained weights $P$\yr{)}, \yr{which may lead to} overfitting and loss of prior information. 
\yr{To analyze the preservation of prior information}, we \yr{calculate} the correlation between the final updated parameter matrix $(P + \Delta P)$ and the pretrained weights $P$. 
\yr{Specifically, w}e calculate the similarity between the subspaces of $(P + \Delta P)$ and $P$. 
We decompose $(P + \Delta P)$ and $P$ using Singular Value Decomposition (SVD) to obtain the left-singular unitary matrices $U$, and examine the similarity between the subspaces spanned by the first $r_i$ singular vectors of $U_{P + \Delta P}$ and the first $r_j$ singular vectors of $U_{P}$. 
We quantify the subspace similarity using the normalized subspace similarity based on the Grassmann distance~\citep{hu2021lora}:
\begin{equation}
    \begin{aligned}
        \phi(P_1,P_2,r_i,r_j)=\frac{\| U_1^{r_iT} U_2^{r_j} \|_F^2}{min(r_i,r_j)} \in [0,1],\\ \text{where } U_k\Sigma_kV_K^T=SVD(P_k), k=\{1,2\}.
    \end{aligned}
    \label{eq:subspace similarity}
\end{equation}
We calculate the similarity between the pre-trained parameter matrix $P$ and the updated parameter matrices obtained from \yr{three approaches: 
1)} our model \yr{$(P+\Delta P_{Ours})$}, 
\yr{2)} LoRA \yr{$(P+\Delta P_{LoRA})$}, 
and \yr{3)} a random parameter matrix added to the pre-trained parameters \yr{$(P+\Delta P_{Random})$}. 
The results are shown in Fig.~\ref{fig:subspace sim between delta P+P and P}. 
As a reference, the subspace similarity $\phi(\yr{P+}\Delta P_{Random},P,r_i,r_j)$ of randomly updated parameters approaches zero across different dimensions $r_j$ and $r_j$, indicating random weights will destroy the prior information in the pre-trained weights absolutely. 
In contrast, the similarity $\phi(\yr{P+}\Delta P_{Ours},P,r_i,r_j)$ between our learned parameter matrix added to the pre-trained parameters and the original parameter matrix exceeds $96\%$ across different subspace dimensions $(r_i,r_j)$, indicating that our learned parameter matrix effectively preserves the information in the original parameter matrix. 
\yr{In addition}, compared to the parameter matrix $\yr{(P+}\Delta P_{Lora}\yr{)}$ \yr{updated} by LoRA, our \yr{updated} parameter matrix \yr{$(P+\Delta P_{Ours})$} shows greater subspace similarity with the pre-trained parameters \yr{$P$}, demonstrating that our model's learned sparse parameter matrix better \yr{preserves} the prior information of the pre-trained parameters, effectively avoiding model overfitting. 
Combining this with the conclusions from the previous section, we can further conclude that \textbf{\textit{Our model can learn more task-specific knowledge, while more effectively preserving the prior information of the pre-trained parameter matrix than LoRA.}}

\textbf{The Correlation between $\Delta P$ \yr{under} Different Threshold\yr{s}.}
We further investigate the relationship between \yr{the} learned parameter matrices $\Delta P$ under different thresholds. 
Our experiments focus on the matrices for Key $W_K$ and Value $W_V$ from the medium block's attention modules in SD1.5. 
In this \yr{experiments}, we select two thresholds, $2e-3$ and $8e-4$ (correspond\yr{ing} to $\Delta P_1$ and $\Delta P_2$), and compute the similarity of their subspaces using Eq. (\ref{eq:subspace similarity}). 
For comparison, we also calculate the similarity between the parameter matrix $\Delta P_1$ learned with a threshold of $2e-3$ and the pre-trained parameter matrix $P$. 
The results are shown in Fig.~\ref{fig:subspace sim between delta p}. 
It can be observed that the overall similarity between the parameter matrices learned under the two thresholds exceeds $60\%$ (peaking at around $r = 650$), indicating that the knowledge learned under different thresholds is \yr{roughly} similar, \yr{but} with different emphases (lower similarity at smaller $r$).
In contrast, the similarity between $\Delta P_1$ and the pre-trained parameters $P$ is consistently below $40\%$.
\yr{The higher similarity between $\Delta P_1$ and $\Delta P_2$}
suggest\yr{s} that the learned parameter matrices from different thresholds indeed capture similar task-specific knowledge, which supports the feasibility of fine-tuning the model with fewer parameters.

\section{Hyperparameter Analysis}
\label{sec: hyperparameter analysis}

In this section, we conduct experiments on different hyperparameters in our model: learning rate, progressive iteration \yr{(the iteration for progressive parameter adjustment)}, and \yr{the} weight for rank loss $\lambda_{rank}$. We \yr{chose} Stable Diffusion 1.5 and Expedition dataset for the following experiments (if not specified, the threshold $\theta_t=2e-3$) and evaluated the results by FID, CLIP Score, and VLHI.



\begin{figure}[t]
\centering
\includegraphics[width=1.0\textwidth]{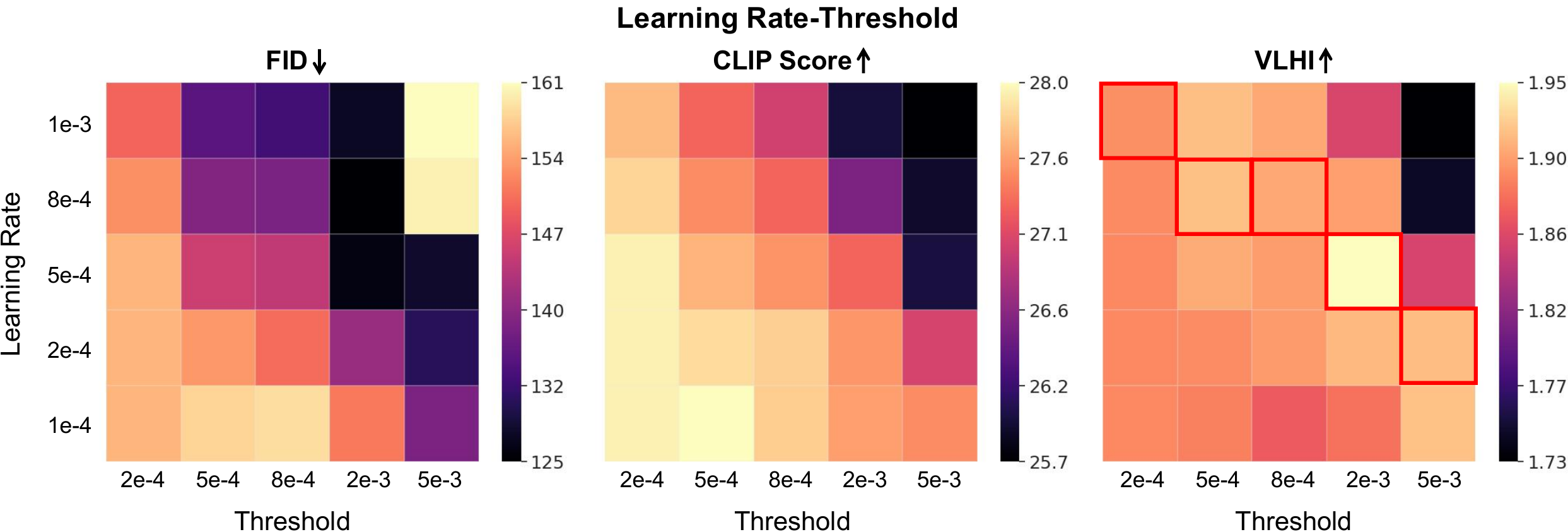}
\vspace{-0.15in}
\caption{The comparison results on different learning rates and thresholds. The model with a larger threshold should employ a larger learning rate to learn the target-domain information well (\textcolor{red}{red} boxes indicate the best results).}
\label{fig:lr-threshold}
\vspace{-0.15in}
\end{figure}

\textbf{Learning Rates and Thresholds.}
\label{sec:lr and th}
We first investigate the two most critical hyperparameters: learning rate and threshold. We selected learning rates $\{1e-4, 2e-4, 5e-4, 8e-4, 1e-3\}$ and thresholds $\{2e-4, 5e-4, 8e-4, 2e-3, 5e-3\}$ for our experiments, resulting in a total of 25 models. The quantitative results are shown in Fig.~\ref{fig:lr-threshold}. It can be observed that, for the same learning rate, as the threshold increases, the model's FID gradually decreases while the CLIP Score gradually increases. This indicates that a larger \yr{learnable} parameter set can learn more task-specific information\yr{,} but is also more likely to lose pre-trained prior knowledge. For the same threshold, increasing the learning rate yields similar results. However, for relatively large thresholds (e.g., $5e-3$), a high learning rate (e.g., $8e-4$ and $1e-3$ in the figure) may cause the model training to collapse. Therefore, selecting an appropriate learning rate is crucial for achieving \yr{good} results.

\begin{wrapfigure}{r}{0.5\textwidth}
\centering
\includegraphics[width=0.5\textwidth]{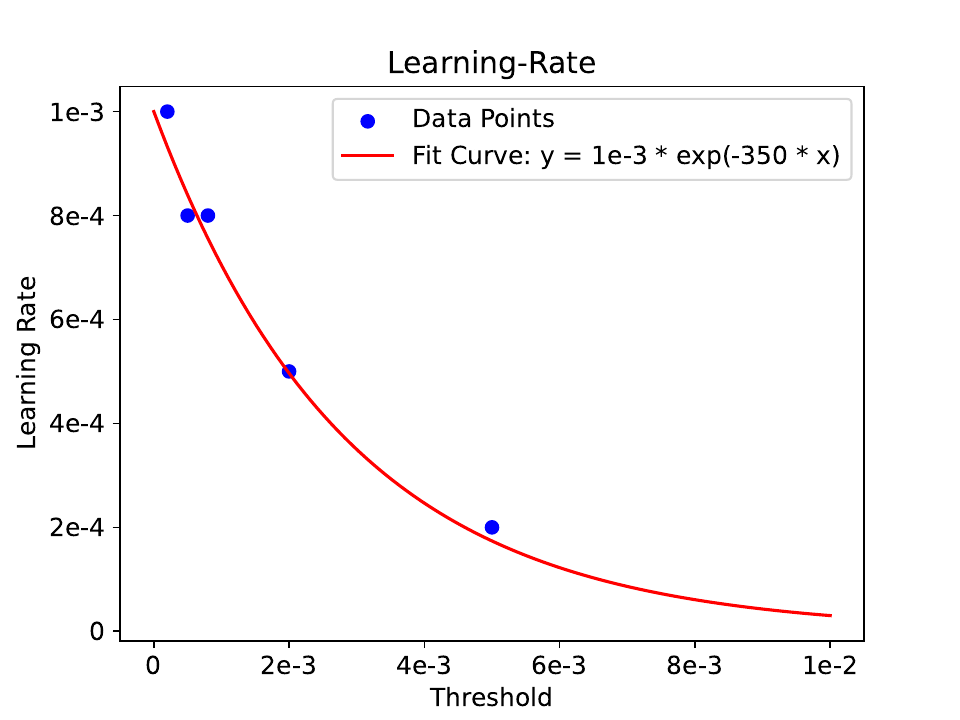}
\vspace{-0.15in}
\caption{The fit curve of the best pairs of learning rate and threshold.}
\label{fig:fit curve}
\end{wrapfigure}

We \yr{further use the VLHI metric to} analyze the performance of \yr{the} models trained with different learning rates under various thresholds \yr{by balancing} FID and CLIP \yr{scores}, \yr{as shown in} the third column in Fig.~\ref{fig:lr-threshold}.
The optimal learning rate for each threshold is marked with a red box. It can be seen that as the threshold increases, a gradually decreasing learning rate should be used to prevent severe overfitting. Conversely, as the threshold decreases, a larger learning rate should be employed to enhance the model's ability to learn task-specific knowledge. In summary, there is a negative correlation between the learning rate and the threshold. To adaptively select an optimal learning rate, we fit an exponential function $f(x) = a \times e^b$ using the five data points shown in the figure. The resulting function for adaptively computing the learning rate for different thresholds is shown in Fig.~\ref{fig:fit curve}. The curve fits the five data points well, and when the threshold approaches $0$, the learning rate is approximately $1e-3$, which does not result in an excessively high learning rate. Similarly, for larger thresholds (e.g., $1e-2$), the learning rate is around $3e-5$, comparable to the learning rate used in full fine-tuning, avoiding an excessively low learning rate. We do not consider even larger thresholds, as these parameters are highly effective in the model, and fine-tuning them would contradict the purpose of our method. Therefore, we derive a function \yr{to adaptively} comput\yr{e a good} learning rate \yr{$Lr$} based on the threshold \yr{$\theta_t$:}
\begin{equation}
Lr = 10^{-3} \times e^{-350 \theta_t}.
\end{equation}

\begin{figure}[t]
\centering
\includegraphics[width=1.0\textwidth]{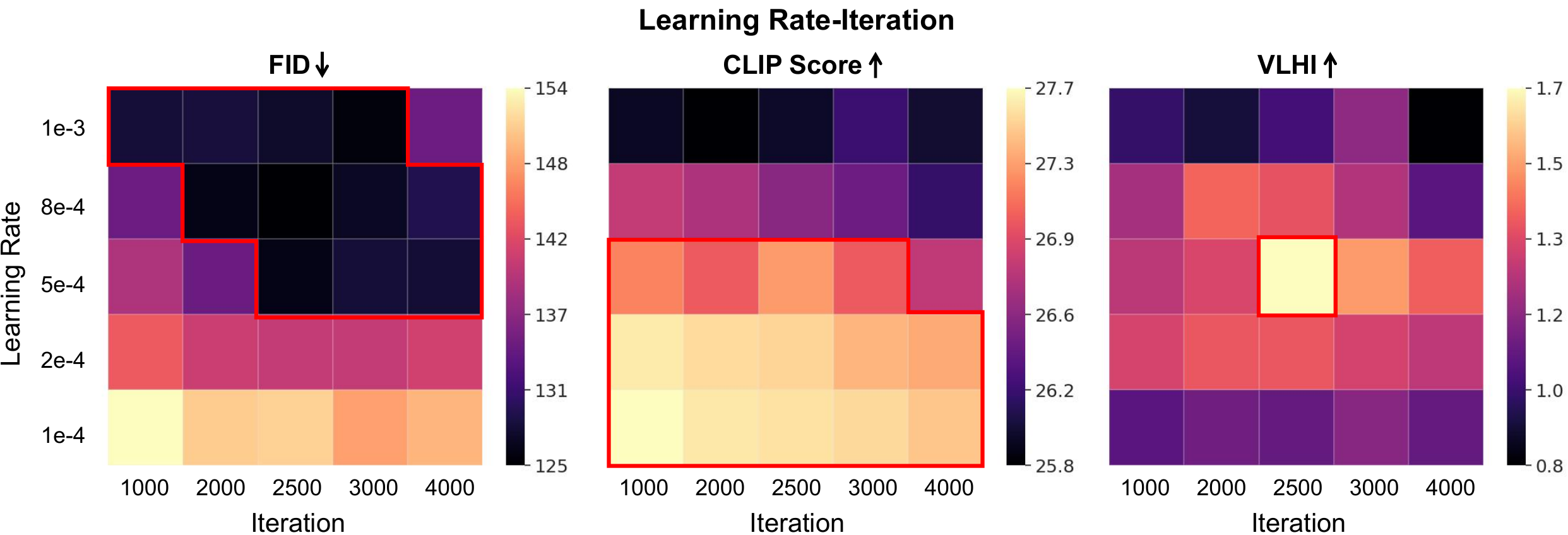}
\vspace{-0.2in}
\caption{The comparison results on different learning rates and progressive iterations. A larger learning rate or progressive iteration improves the FID while sacrificing the CLIP Score.}
\label{fig:lr-iter}
\vspace{-0.1in}
\end{figure}

\textbf{Learning Rates and Progressive Iteration.} 
We then study the effects of learning rate and progressive iteration \yr{(the iteration for progressive parameter adjustment)} together. 
We train the models with learning rates $\{1e-4,2e-4,5e-4,8e-4,1e-3\}$ and progressive iteration $\{1000,2000,2500,3000,4000\}$, which \yr{forms} 25 models in total. The quantitative results are shown in Fig.~\ref{fig:lr-iter}. For all the metrics (FID, CLIP Score, and VLHI), the brighter the color is, the better the model performs. It can be seen that\yr{,} as the learning rate or progressive iteration grows, the model learns more task-\yr{specific} knowledge (a better FID), while the CLIP score becomes worse. Therefore, we should balance both the learning rate and progressive iterations, where the model with learning rate $5e-4$ and progressive iteration $2000$ achieves the best VLHI, reaching both a good FID and CLIP Score.

\begin{figure}[t]
\centering
\includegraphics[width=1.0\textwidth]{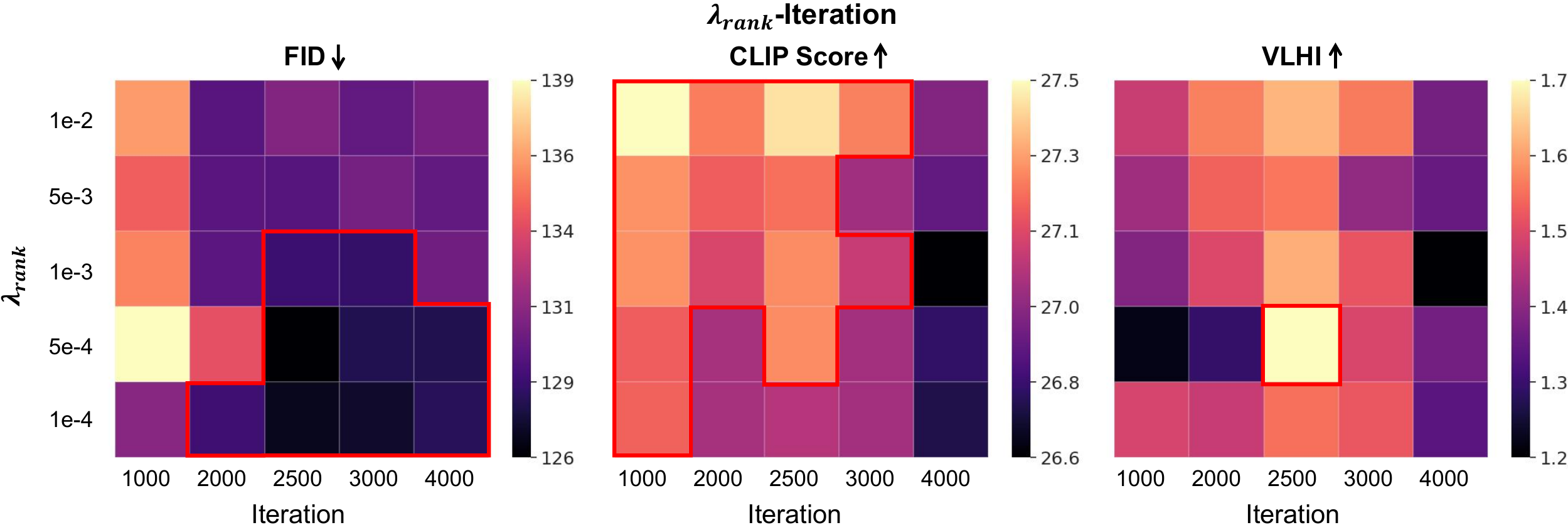}
\vspace{-0.2in}
\caption{The comparison results on different weights $\lambda_{rank}$ for rank loss and progressive iterations. A larger $\lambda_{rank}$ improves the CLIP Score while sacrificing the FID.}
\label{fig:rank-iter}
\vspace{-0.15in}
\end{figure}

\textbf{$\mathbf{\lambda_{rank}}$ and Progressive Iteration.}
We then analyze the influence of the weight for rank loss $\lambda_{rank}$ and progressive iteration at the same time.
We train the models with $\lambda_{rank}$ $\{1e-4,5e-4,1e-3,5e-3,1e-2\}$ and progressive iteration $\{1000,2000,2500,3000,4000\}$, which \yr{constitutes} 25 models in total. The quantitative results are shown in Fig.~\ref{fig:rank-iter}. It can be seen that as $\lambda_{rank}$ increases, the FID becomes worse while the CLIP Score performs better, demonstrating that $\lambda_{rank}$ helps the model keep the prior information in the pre-trained weights, but with a less effect in fitting the target domain. Therefore, to \yr{simultaneously} reach a relatively good FID and CLIP Score, we choose $\lambda_{rank}=5e-3$ with progressive iteration $2500$, which results in the best VLHI.

\begin{figure}[t]
\centering
\includegraphics[width=1.0\textwidth]{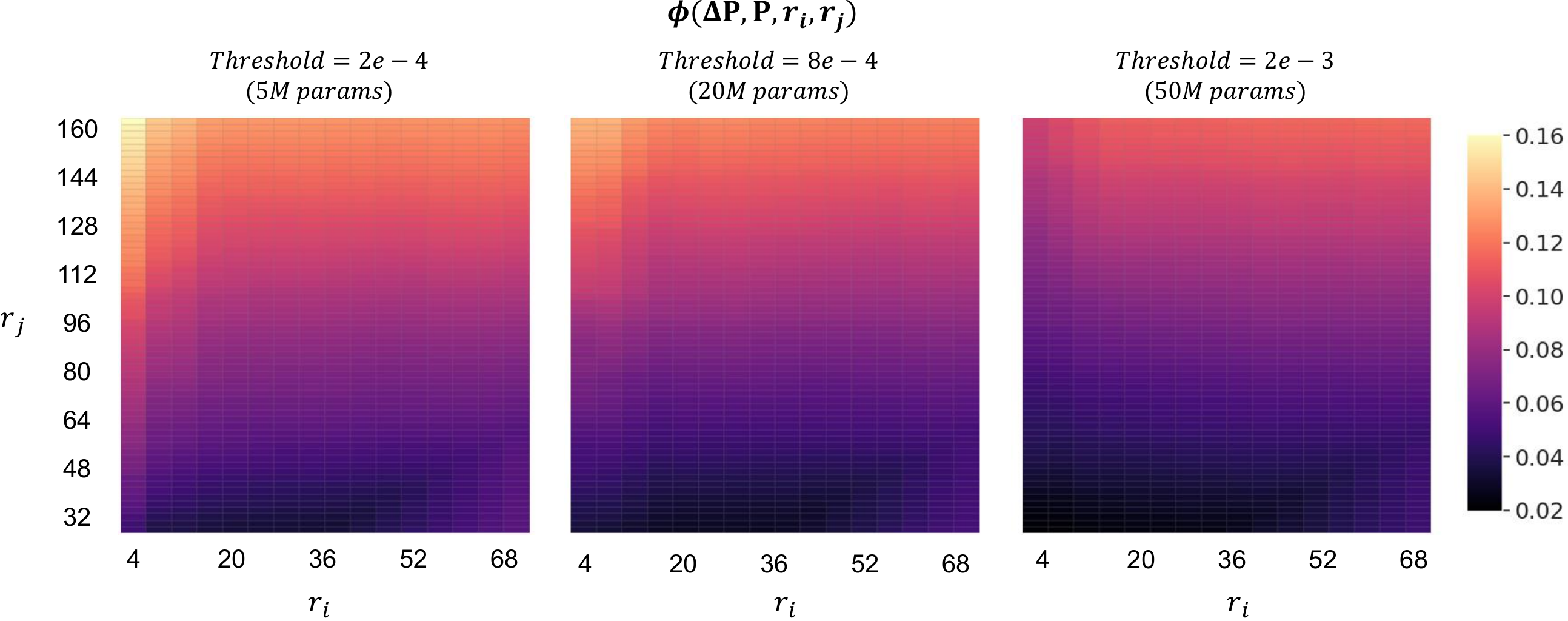}
\vspace{-0.2in}
\caption{Subspace similarity between $\Delta P$ and $P$ \yr{under different thresholds}.}
\vspace{-0.05in}
\label{fig:subspace sim between delta P and P}
\end{figure}

\begin{figure}[t]
\centering
\includegraphics[width=1.0\textwidth]{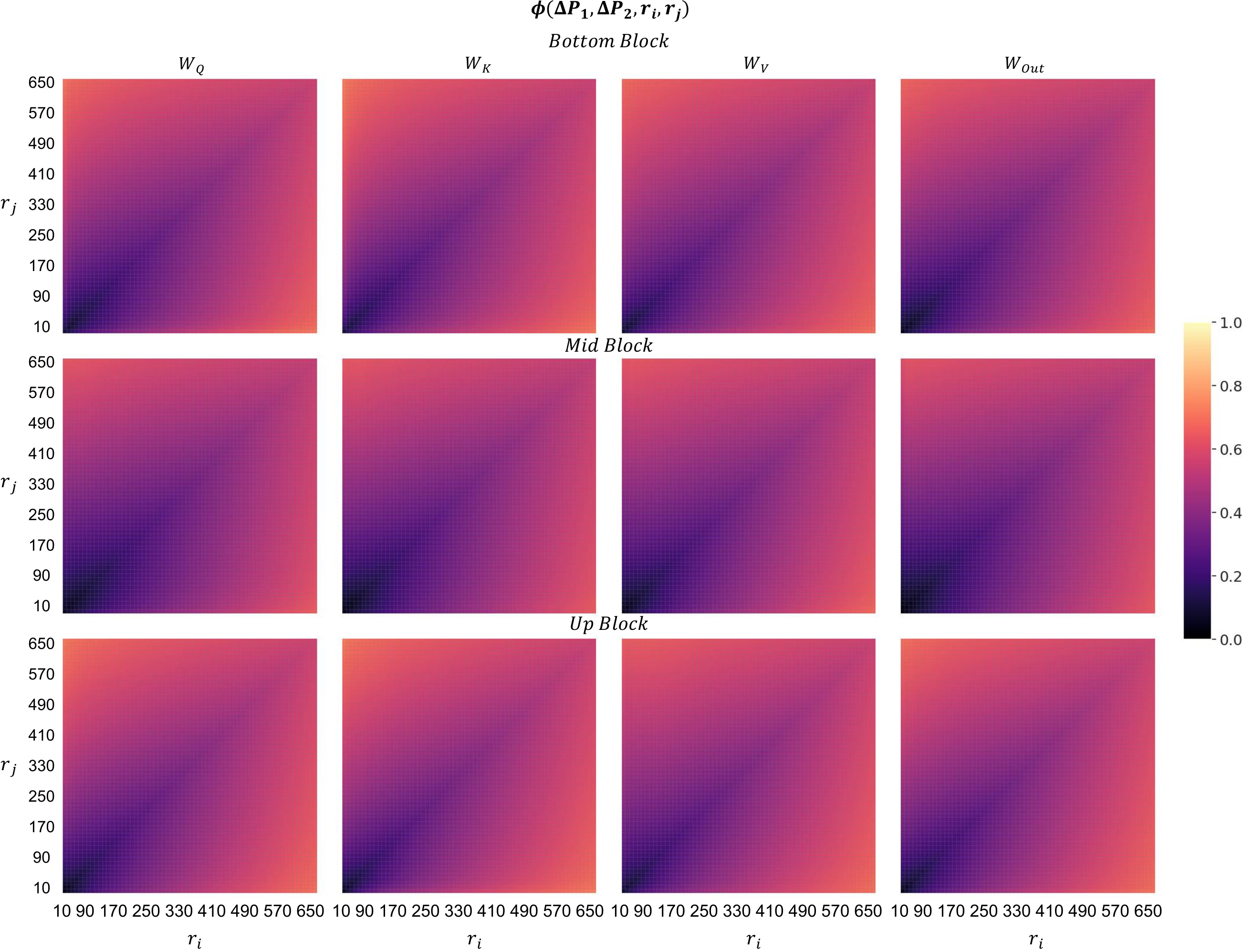}
\vspace{-0.2in}
\caption{Subspace similarity between $\Delta P$ from threshold $\theta_t=2e-3$ (50M parameters) and $\theta_t=8e-4$ (20M parameters), 
in different attention layers\hut{, where the x-axis and y-axis represent $\theta_t=2e-3$ and $\theta_t=8e-4$ respectively}. The subspace similarity across different layers exhibits consistent behavior, demonstrating that the knowledge learned by $\Delta P$ remains invariant across layers and modules, indicating strong robustness.}
\vspace{-0.15in}
\label{fig:heatmap-different_layer}
\end{figure}

\section{More Analysis on the learned weight matrix $\Delta P$}

\label{sec: more analysis on the learned weight matrix}
\textbf{The Correlation between $\Delta P$ and $P$ \yr{under} Different Threshold\yr{s}.} 
We compute the subspace similarity between the learned matrices $Delta_P$ under different thresholds and the pre-trained weights $P$ by Eq. (6) of the main paper. The results are shown in Fig.~\ref{fig:subspace sim between delta P and P}. It demonstrates that $\Delta P$ does not contain the top singular directions of W, since the overall similarity between the singular directions in the learned matrices $\Delta P$ and the top 32 directions of $P$ is barely around $4\%$. And it further validates that the matrices $\Delta P$ contain more task-\yr{specific} information rather than repeating the directions that are already emphasized in the pre-trained weights. Moreover, by comparing the $\Delta P$ from different thresholds, we can find that as the thresholds grow, the subspace similarity between $\Delta P$ and $P$ becomes smaller, indicating that a larger threshold can learn more task-\yr{specific} information, therefore a large threshold can contribute to a better FID as shown in Tab. 1 of the main paper.

\begin{figure}[t]
\centering
\includegraphics[width=1.0\textwidth]{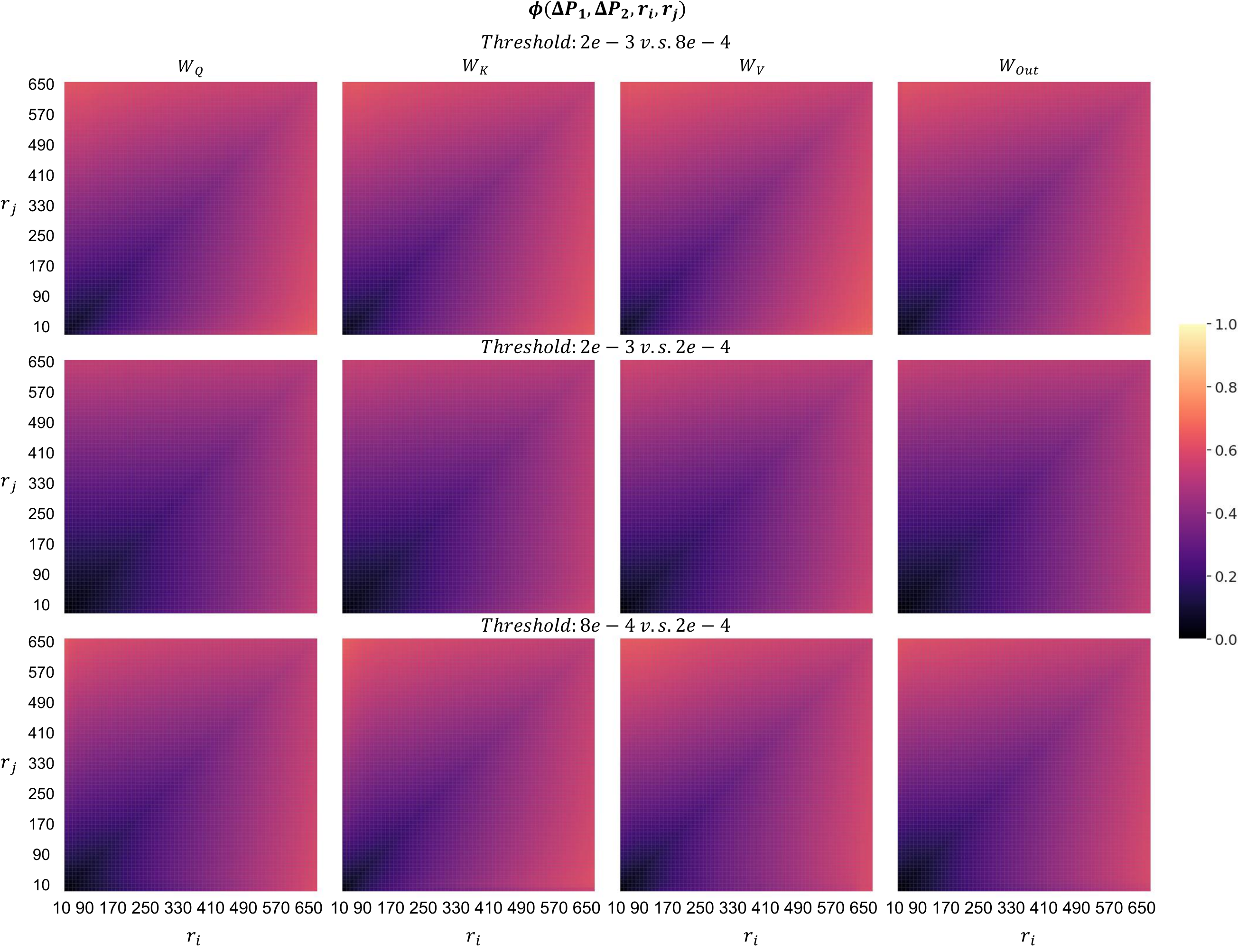}
\vspace{-0.2in}
\caption{Subspace similarity between $\Delta P$ 
\yr{across different threshold pairs: $(2e-3, 8e-4)$, $(2e-3, 2e-4)$, and $(8e-4, 2e-4)$ (from top to bottom).}
}
\label{fig:heatmap-different_threshold}
\vspace{-0.15in}
\end{figure}

\textbf{More Analysis on $\Delta P$ from Different Layers.} 
In the main paper, we have analyzed the subspace similarity between the learned matrices $\Delta P$ from different thresholds, and \yr{concluded} that the matrices from different thresholds learn similar task-\yr{specific} knowledge, but emphasize different directions. To further validate this conclusion, we conduct more quantitative analysis between different thresholds \hut{($\theta_t=2e-3$ and $8e-4$)}
from the attention \yr{layers} in the bottom, medium, and up blocks. Moreover, we take all the learnable matrices in the attention module into consideration, including the Query, Key, Value, and FFN matrices (corresponds to $W_Q$, $W_K$, $W_V$, and $W_{Out}$ respectively). The results can be referred to in Fig.~\ref{fig:heatmap-different_layer} \hut{the x-axis and y-axis represent $\theta_t=2e-3$ and $\theta_t=8e-4$ respectively.}, where the heatmaps show almost the same color and distributions, indicating that our conclusion is consistent for the learned matrices from different modules and different \yr{attention} layers.

\textbf{Further Analysis of $\Delta P$ \yr{a}cross Different Threshold Pairs.} 
In the main paper, we analyzed the subspace similarity between the learned matrices $\Delta P$ from thresholds of $2e-3$ and $8e-4$, concluding that the learned matrices from different thresholds capture similar task-specific knowledge while emphasizing different directions. In this section, we extend our quantitative analysis to additional threshold pairs: $(2e-3, 8e-4)$, $(2e-3, 2e-4)$, and $(8e-4, 2e-4)$. We consider all learnable matrices in the attention module, including the Query, Key, Value, and FFN matrices (corresponding to $W_Q$, $W_K$, $W_V$, and $W_{Out}$, respectively). The results are presented in Fig.~\ref{fig:heatmap-different_threshold}. The subspace similarity between the thresholds $(2e-3, 2e-4)$ is lower than that of $(2e-3, 8e-4)$ and $(8e-4, 2e-4)$, suggesting that matrices learned from a closer threshold pair exhibits greater subspace similarity and acquire more similar knowledge.

\section{\reb{Limitations}}
\label{sec:limitation}
\reb{Our SaRA focuses on fine-tuning the ineffective parameters of a pre-trained model. However, if the model size is relatively small (e.g., not as large as diffusion models, which typically exceed 100M parameters), the number of ineffective parameters may be insufficient to effectively adapt the model to the downstream dataset. As a result, SaRA is better suited for fine-tuning large models rather than smaller ones. Additionally, since there is no rigorous proof that parameters with the smallest absolute values are always ineffective, caution is warranted to account for potential exceptions, which could lead to reduced performance of SaRA in certain scenarios.}

\end{document}